\documentclass[journal]{IEEEtran}
\ifCLASSINFOpdf
\else
\fi
\usepackage{amsmath}
\usepackage{array}

\usepackage{graphicx}
\usepackage{multirow}
\usepackage{booktabs}
\usepackage{array}
\usepackage{soul,color}
\usepackage{tablefootnote}
\usepackage{esvect}
\usepackage{comment}
\usepackage{pgfplots,capt-of}
\usepackage{booktabs}
\usepackage{adjustbox}
\usepackage{caption}
\usepackage{subcaption}
\usepackage{adjustbox}
\usepackage{xcolor}
\usepackage{cite}
\usepackage{longtable}
\usepackage{supertabular}
\usepackage{amssymb}
\usepackage{nicefrac}
\usepackage{amsmath,scalerel}

\RequirePackage[normalem]{ulem} 
\RequirePackage{color}\definecolor{RED}{rgb}{1,0,0}\definecolor{BLUE}{rgb}{0,0,1} 
\begin{document}

\title{Graph-Based Deep Learning for Medical \\ Diagnosis and Analysis: Past, Present and Future}

\author{
David Ahmedt-Aristizabal,
Mohammad Ali Armin, 
Simon Denman,
Clinton Fookes,
Lars Petersson
\thanks{D. Ahmedt-Aristizabal, A. Armin and L. Petersson are with the Imaging and Computer Vision group, CSIRO Data61, Canberra, Australia. 
({Corresponding author: \tt\footnotesize david.ahmedtaristizabal@data61.csiro.au})}
\thanks{D. Ahmedt-Aristizabal, S. Denman and C. Fookes are with SAIVT, Queensland University of Technology, Brisbane, Australia. 
}
}

%



\maketitle

\begin{abstract}
With the advances of data-driven machine learning research, a wide variety of prediction problems have been tackled. It has become critical to explore how machine learning and specifically deep learning methods can be exploited to analyse healthcare data.
A major limitation of existing methods has been the focus on grid-like data; however, the structure of physiological recordings are often irregular and unordered which makes it difficult to conceptualise them as a matrix. As such, graph neural networks have attracted significant attention by exploiting implicit information that resides in a biological system, with interactive nodes connected by edges whose weights can be either temporal associations or anatomical junctions. 
In this survey, we thoroughly review the different types of graph architectures and their applications in healthcare. We provide an overview of these methods in a systematic manner, organized by their domain of application including functional connectivity, anatomical structure and electrical-based analysis. We also outline the limitations of existing techniques and discuss potential directions for future research.
\end{abstract}


\IEEEpeerreviewmaketitle

\section{Introduction}

\IEEEPARstart{M}{edical}
diagnosis refers to the process by which one can determine which disease or condition explains a patient's symptoms. The required information for a diseases diagnosis is obtained from a patient's medical history and various medical tests that capture the patient's functional and anatomical structures through diagnostic imaging data such as functional magnetic resonance imaging (\textit{f}MRI), magnetic resonance imaging (MRI), computed tomography (CT), ultrasound (US) and X-ray; and other diagnostic tools include electroenchephalogram (EEG). However, given the often time-consuming diagnosis process which is prone to subjective interpretation and inter-observer variability, clinical experts have begun to benefit from computer-assisted interventions.
%
Automation is also of benefit in situations where there is limited access to healthcare services and physicians. Automation is being pursued to increase the quality and decrease the cost of healthcare systems~\cite{sutton2020overview}. Deep learning offers an exciting avenue to address these demands by incorporating the task of feature engineering within the learning task~\cite{lecun2015deep}.
%
There are several review papers available that analyse the benefits of traditional machine learning and deep learning methods for the detection and segmentation of medical anomalies and anatomical structures, analysis of motor disorders and sequential data, computer-aided detection and computer-aided diagnosis~\cite{hosseini2020review,ahmed2018neuroimaging,ahmedt2017automated,shen2017deep}.

Graph networks belong to an emerging area that has also made a tremendous impact across many technological domains. 
Much of the information coming from disciplines such as chemistry, biology, genetics, and healthcare, is not well suited to vector-based representations, and instead requires complex data structures. 
Graphs inherently capture relationships between entities, and are thus potentially very useful in many of these applications to encode relational information between variables. For example, in healthcare, it is possible to construct a knowledge graph by relating subjects with diseases or symptoms during the Physician's decision process~\cite{choi2019graph}, or to model RNA-sequences for breast cancer analysis~\cite{rhee2018hybrid}. 
Hence, special attention has been devoted to the generalization of graph neural networks (GNN) into non-structural (unordered) and structural (ordered) scenarios. However while the use of graph-based representations is becoming more common in the medical domain, such approaches are still scarce compared to conventional deep learning methods, and their potential to address many challenging medical problems is yet to be fully realised.

\begin{figure}[!t]
\centering
\includegraphics[width=1\linewidth]{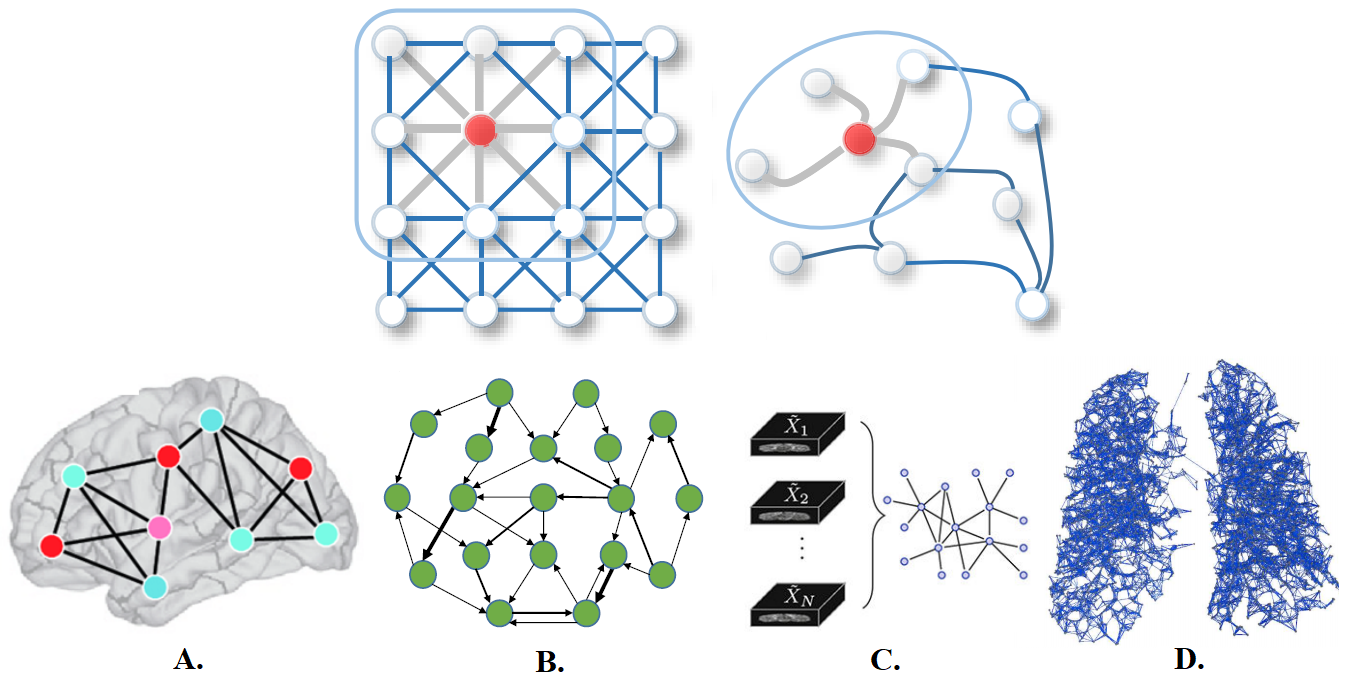}
\vspace{-15pt}
\caption{
\textbf{Top.} Traditional 2D grid representation and graph-based representation (the neighbors of a node are unordered and variable in size).
\textbf{A.} and \textbf{B.} Brain graph of \textit{f}MRI and EEG data for brain responses and emotion analysis, respectively.
\textbf{C.} DMRI sampling represented by a graph (DMRI brain reconstruction).
\textbf{D.} Graph-like representation for organ segmentation (CT -pulmonary airway).
Image adapted from~\cite{wu2020comprehensive,zhang2019functional,song2018eeg,hong2019multifold,selvan2020graph}.
}
\label{fig:Fig3}
\vspace{-10pt}
\end{figure}

The popularity of the rapidly growing field of deep learning on GNNs is also reflected by the numerous recent surveys on graph representations and their applications.
Existing reviews provide a comprehensive overview on deep learning for non-Euclidean data, graph deep learning frameworks and a taxonomy of existing techniques~\cite{wu2020comprehensive,bronstein2017geometric}; 
or introduce general applications which cover biology and signal processing domains~\cite{georgousis2021graph,zhang2020deep,abadal2020computing,zhou2018graph}. 
Although some papers have surveyed medical image analysis using deep learning techniques and have introduced the concept of GNNs for the assessment of neurological disorders~\cite{zhang2020survey}, to the best of our knowledge, no systematic review exists that introduces and discusses the current applications of GNNs to unstructured medical data. 

In this paper, we endeavour to provide a thorough methodological review of multiple graph neural networks (GNN) models proposed for use in medical diagnosis and analysis. We seek to explain the fundamental reasons why GNNs are worth investigating in this domain, and highlight the emerging medical analytics challenges that GNNs are well placed to address.

\vspace{-6pt}
\subsection{Why graph-based deep learning for medical diagnosis and analysis?}

The success of deep learning in many fields is due in part to the availability of rapidly increasing computing resources and large experimental datasets, and in part to the ability of deep learning to extract representations from data structured as regular grids (\textit{i.e.} images) through stacked convolutional operations.
Recent progress in deep learning has increased the potential of medical image analysis by enabling the discovery of morphological, textural and temporal representations from images and signals solely from the data. 

Although CNNs have shown impressive performance in the medical field for imaging (MRI, CT) and non-imaging applications (\textit{f}MRI, EEG), their conventional formulation is limited to data structured in an ordered grid-like fashion. Several physical human processes generate data that is naturally embedded in a graph structure. Traditional CNNs do not capture complex neighborhood information as they analyse local areas based on fixed connectivity (determined by the convolutional kernel), leading to limited performance and interpretability of the analysis of functional and anatomical structures. Therefore, machine-learning models that can exploit graph structures are at an advantage as they enable an effective representation of complex physical entities and processes, and irregular relationships. 

Graph neural networks (GNNs) are a deep learning-based method that operate over graphs, and have been adopted in diverse fields including social network analysis and drug discovery using computational chemistry~\cite{wu2020comprehensive}.
Graph models are becoming increasingly powerful, allowing their application to challenging open problems in the medical field. For example, the relationship between channels and frequencies for brain signals is rather arbitrary and complicated. Compared with CNNs, graph neural networks represent signals from brain regions as nodes in a topological graph and represent the relationships between them using the graph edges. This structure can preserve rich connection information compared to what is possible with the 2D and 3D matrices used by regular CNNs. 

Graph convolutional networks (GCNs) have extended the theory of signal processing on graphs~\cite{shuman2013emerging} to enable the representation learning power of CNNs to be applied to irregular graph data. GCNs generalize the convolution operation to non-Euclidean graph data. The graph convolutional operation aims to generate representations for vertices by aggregating its own feature and the features of its neighboring vertices. The relationship-aware representations generated by GCNs tremendously enhance the discriminative ability of CNN features, and the improved model interpretability can help clinicians to determine, for example, the parts of the brain that are most involved in one particular task. 
GNNs have seen a surge in popularity due to their successes in modeling unstructured and structured relational data including brain signals (\textit{f}MRI and EEG), and in the detection and segmentation of organs (MRI, CT) as represented in Fig.~\ref{fig:Fig3}.

Below, we outline several application domains which are well suited to graph networks, and outline the reasons why graph neural networks are becoming more widely used within these domains.

\subsubsection{Brain activity analysis}
Brain signals are an example of a graph signal, and the graph representation can encode the complex structure of the brain to represent either physical or functional connectivity across different brain regions. At the structural level, the network is defined by the anatomical connections between regions of brain tissue. At the functional level, the graph nodes represent brain regions of interest (ROI), while edges capture the correlation between their activities computed via an \textit{f}MRI correlation matrix~\cite{parisot2018disease}.

The structure of EEG channels captured during examination are an example of an irregular layout, and they cannot be simply modelled using the physical position of electrodes alone. GCNs offer advantages when dealing with discriminative feature extraction from signals in the discrete spatial domain, and for applications such as EEG analysis can capture hidden relationships among EEG signals from different channels. GCNs provide an effective way to discover and model this intrinsic relationship between different nodes of the graph or contacts~\cite{song2018eeg}. 

GNN models also offer advantages when considering the need to develop deep-learning scoring models which allow a direct interpretation of non-Euclidean spaces. This explanation can help to identify and localize regions relevant to a model’s decisions for a particular task. An example is how certain brain regions are related to a specific neurological disorder, which are defined as biomarkers~\cite{li2019graph,li2020pooling}. 

\subsubsection{Brain surface representation}
The structures in medical images have a spherical topology (\textit{i.e.} brain cortical or subcortical surfaces) and these are at-times represented by triangular meshes with large inter- and intra-subject variations in vertex numbers and local connectivity. Due to the absence of a consistent and regular neighborhood definition, conventional CNNs cannot be directly applied to these surfaces~\cite{zhao2019spherical}. GCNs, however, can be applied to graphs with varying numbers of nodes and connectivity~\cite{gopinath2020learnable}.
Spherical CNN architectures can render valid parametrizations in the spherical space without introducing spatial distortions on the sphere (spherical mapping)~\cite{wu2019intrinsic}, and geometric features can be augmented by utilizing surface registration methods~\cite{hao2020automatic}.
GCNs can also offer more flexibility to parcellate the cerebral cortex (surface segmentation) by providing better generalization on target-domain datasets where surface data is aligned differently, without the need for manual annotations or explicit alignment of these surfaces~\cite{gopinath2020graph}.

\subsubsection{Segmentation and labeling of anatomical structures}
Segmentation of vessels and organs is a critical but challenging stage in the medical image processing pipeline due to anatomical complexity. 
Traditional deep learning segmentation approaches classify each pixel of an image into a class by extracting high-level semantic features. CNNs fail because regions in images are rarely grid-like and require non-local information.
Compared with these pixel-wise methods, a graph-based method learns and regresses the location of the vessels and organs directly and allows the model to learn local spatial structures~\cite{noh2020combining,tian2020graph}. GCNs can also propagate and exchange local information across the whole image to learn the semantic relationships between objects.

\subsubsection{Multi-modal medical data analysis}
Multi-modal neuroimage analysis is increasing in prevalence due to the limitations of single modalities, which is resulting in larger and increasingly complex data sets. 
It can be difficult to combine imaging and non-imaging data from populations into a unified model. For disease classification, traditional multi-modal learning-based approaches usually summarize features of all modalities with a CNN, which ignores the interactions and associations between subjects in a population. 
The association among instances (subjects) is important, and neighboring patients in the graph should be considered when, for example, learning embeddings for brain functional networks. Recently, researchers have utilized advances in graph convolutional networks to address these concerns.
Graphs provide a natural way to represent the population data and model complex interactions by combining features of different modalities for disease analysis~\cite{huang2020edge}. Each subject is modeled as a node (patients or healthy controls) along with a set of features, and the graph edges are defined based on the similarity between the features of the subjects~\cite{rakhimberdina2020population}.

\vspace{-6pt}
\subsection{Scope of review}

The application of graph neural networks to medical signal processing and analysis is still in its nascent stages. In this paper, we present a survey that captures the current efforts to apply graph neural networks to medical diagnostic tasks, and present the current state of the art methods and trends in the area.

\begin{figure}[t!]
  \centering
  \begin{tabular}{cc}
  \subfloat[Modalities]
  {
    \resizebox{0.15\textwidth}{!}{%
    \begin{tabular}{
    >{\raggedright\arraybackslash}p{1.5cm} 
    c
    }
    \toprule
    \textbf{Modality} &
    \textbf{\#Papers} \\
    \midrule
    MRI                                 & 31 \\    
    rs-\textit{f}MRI                    & 20 \\
    EEG                                 & 19 \\
    %
    \bottomrule
    \end{tabular}}
  }
   &  
  \subfloat[Literature review chronology]
  {
    \begin{tikzpicture}[scale=0.53, transform shape,baseline={(0,0.6)}]
        \begin{axis}[
                ybar,
                symbolic x coords={2017,2018,2019,2020,2021},
                xtick=data,
                nodes near coords,
            ]
            \addplot table[x=interval,y=carT]{\mydata};
        \end{axis}
    \end{tikzpicture}
  }
  \\
  \end{tabular}
  \caption{
  a) Summary of the most representative modalities analysed from the review papers and their number of applications.
  b) Chronology of published manuscripts considered in this review paper.
  }%
  \label{fig:literature-review}%
\vspace{-8pt}
\end{figure}
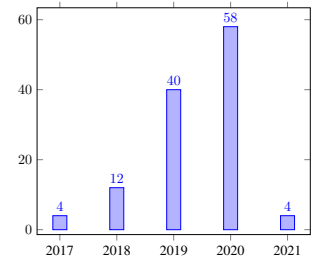

The survey encompasses research papers on various applications of GNNs in medical data understanding and diagnosis. Papers included in the survey are obtained from various journals, conference proceedings and open-access repositories (Arxiv, bioRxiv). 
Unranked conferences and journals and manuscripts that do not provide information on the clinical application, models and experimental setup are excluded from the review. The total number of applications considered in our survey are summarised in Fig.~\ref{fig:literature-review}. 
We found that MRI and rs-\textit{f}MRI constitute the major data modality used for applications in healthcare followed by EEG. The area of digital pathology (WSI) is omitted from this review due to the diverse applications of GCNs to this domain, which we feel merit their own separate review paper.
%

\pgfplotstableread[row sep=\\,col sep=&]{
    interval & carT \\
    2017      & 4 \\
    2018      & 12 \\
    2019      & 40 \\
    2020      & 58  \\
    2021      & 4  \\    
    }\mydata

\vspace{-6pt}
\subsection{Contribution and organisation}

Compared to other recent reviews that cover the theoretical aspects of graph networks in multiple domains, our manuscript has novel contributions which are summarized as follows:

\begin{enumerate}
\item We identify a number of challenges facing traditional deep learning when applied to medical signal analysis, and highlight the contributions of graph neural networks to overcome these.
\item We introduce and discuss diverse graph frameworks proposed for medical diagnosis and their specific applications. We cover work for biomedical imaging applications using graph networks combined with deep learning techniques.
\item We summarise the current challenges faced by graph-based deep learning, and propose future directions in healthcare based on the currently observed trends and limitations.
\end{enumerate}


Based on the previous summary of surveyed papers analysed in this manuscript and their specific applications, in Section~\ref{sec:sec2} we briefly describe the most common graph-based deep learning models used in this domain including GCNs and its variants, with temporal dependencies and attention structures.

In Section~\ref{sec:sec3} we explain all the use cases identified in the literature review. We organise publications according to the input data (functional connectivity, electrical-based, and anatomical structure) and cluster approaches based on specific applications (\textit{e.g.} Alzheimer's disease, breast cancer detection, organ segmentation, or brain data regression).

Finally, Section~\ref{sec:sec4} highlights the limitation of current GNNs adopted for medical diagnosis and introduces graph-based deep learning techniques that can be utilised in this domain. We also provide some research directions and future possibilities for the use of GNNs in healthcare that have not been covered in the literature, such as for behavioural analysis. 

%
%
\section{Graph Neural Networks Background}
\label{sec:sec2}%

In this section we introduce several graph-based deep learning models including GCNs and their variants with temporal dependencies, and attention structures, which have been used as the foundation for the medical applications covered in this manuscript. We aim to provide technical insights regarding the architectures. A deep analysis of each architecture can be found in multiple survey papers in this domain~\cite{wu2020comprehensive,zhang2020deep,zhou2018graph}.

\vspace{-6pt}
\subsection{Overview}
Graph neural networks~\cite{scarselli2008graph} aim to extend existing neural networks through graph theory, enabling them to operate over data in a graph structure. Gori et al.~\cite{gori2005new} introduced the notion of graphs to estimate the learning of graph-structured data through propagation of information to neighboring nodes. 

Following the success of convolutional neural networks, Bruna et al.~\cite{bruna2013spectral} was one of the pioneers to apply  convolution operations to a graph neural network by employing a spectrum of graph Laplacian operations, that translate convolutional properties into the Fourier domain emerging in a more straighforward representation of graph data. However, this is computationally expensive and ignores local features.
Defferrard et al.~\cite{defferrard2016convolutional} proposed the ChebyNet, which approximates the spectral filters by truncated Chebyshev polynomials, avoiding the computation of the Fourier basis.
Kipf and Welling~\cite{kipf2017semi} presented the GCN using a localized first-order approximation of spectral convolutions on the graph. It uses a simple layer-wise propagation rule to encode the relationships of nodes from the graph structure into node features, and that helps to generate more informative feature representations. 
Thanks to its simplicity and scalability, the GCN has been successfully applied to computer vision applications including image classification, visual reasoning, semantic segmentation, object tracking, action recognition and others~\cite{wu2020comprehensive,zhou2018graph}.
Some variants have been proposed by, for example, combining ChebyNet with Recurrent Neural Networks (RNN) for structured sequence modeling~\cite{seo2018structured}.

Due to the use of Laplacian matrix computations, spectral approaches can only take homogeneous graph datasets as inputs, where the adjacency matrix is fixed across the data. This is a limitation for multiple domains such as problems that utilise brain cortex data. Several spatial approaches on the other hand can take heterogeneous graphs as inputs, where each graph can have a different number of vertices and a different adjacency matrix~\cite{wu2020comprehensive}. 

\vspace{-6pt}
\subsection{Graph construction and traditional framework}
\begin{figure}[!t]
\centering
\includegraphics[width=0.6\linewidth]{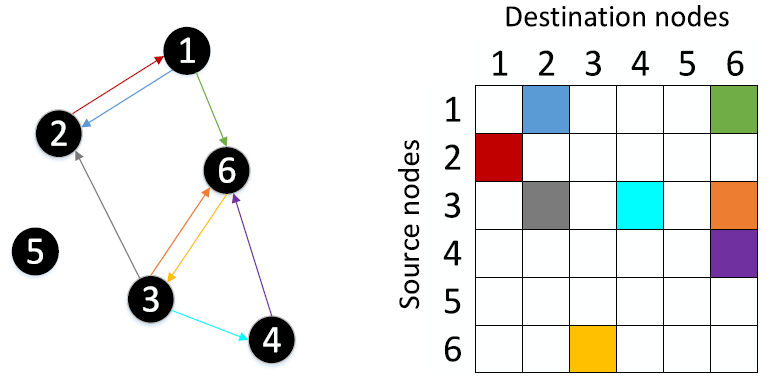}
\caption{
Example of a directed graph (left) and the corresponding adjacency matrix (right). Image adapted from~\cite{song2018eeg}.
}
\label{fig:Fig10}
\vspace{-6pt}
\end{figure}

A graph can be represented as $G=(\mathcal{V},\mathcal{E},W)$ where $V$ represents the set of $N$ nodes, $|\mathcal{V}|=N$; $\mathcal{E}$ denotes the set of edges connecting these nodes and $W$ is the adjacency matrix. The adjacency matrix describes the connections between any two nodes in $\mathcal{V}$, in which the importance of the connection between the \textit{i}-th and the \textit{j}-th nodes is measured by the entry of $W$ in the \textit{i}-th row and \textit{j}-th column, and denoted by $w_{ij}$. Fig.~\ref{fig:Fig10} demonstrates an example of a graph containing six vertices and the edges connecting the nodes of the graph, along with the graph adjacency matrix.

Commonly used methods to determine the entries, $w_{ij}$, of $W$ include the Pearson correlation-based graph, the K-nearest neighbor (KNN) rule method, and the distance-based graph~\cite{shuman2013emerging}. For example, a typical distance function is computed using a thresholded Gaussian kernel which can be expressed as,
\begin{equation}
w_{ij} = 
\begin{cases}
    \text{exp}(-\frac{[\text{dist}_{(i,j)}]}{2\theta^2}),   & \text{if} \; \text{dist}_{(i,j)} \leq \tau \\
    0,                                                      & \text{otherwise}
\end{cases}
\end{equation}
where $\tau$ and $\theta$ are two parameters to be fixed based, for example, on the physical distance between electrode pairs and $\text{dist}_{(i,j)}$ is the distance between the \textit{i}-th and \textit{j}-th node.

\begin{figure}[!t]
\centering
\includegraphics[width=1\linewidth]{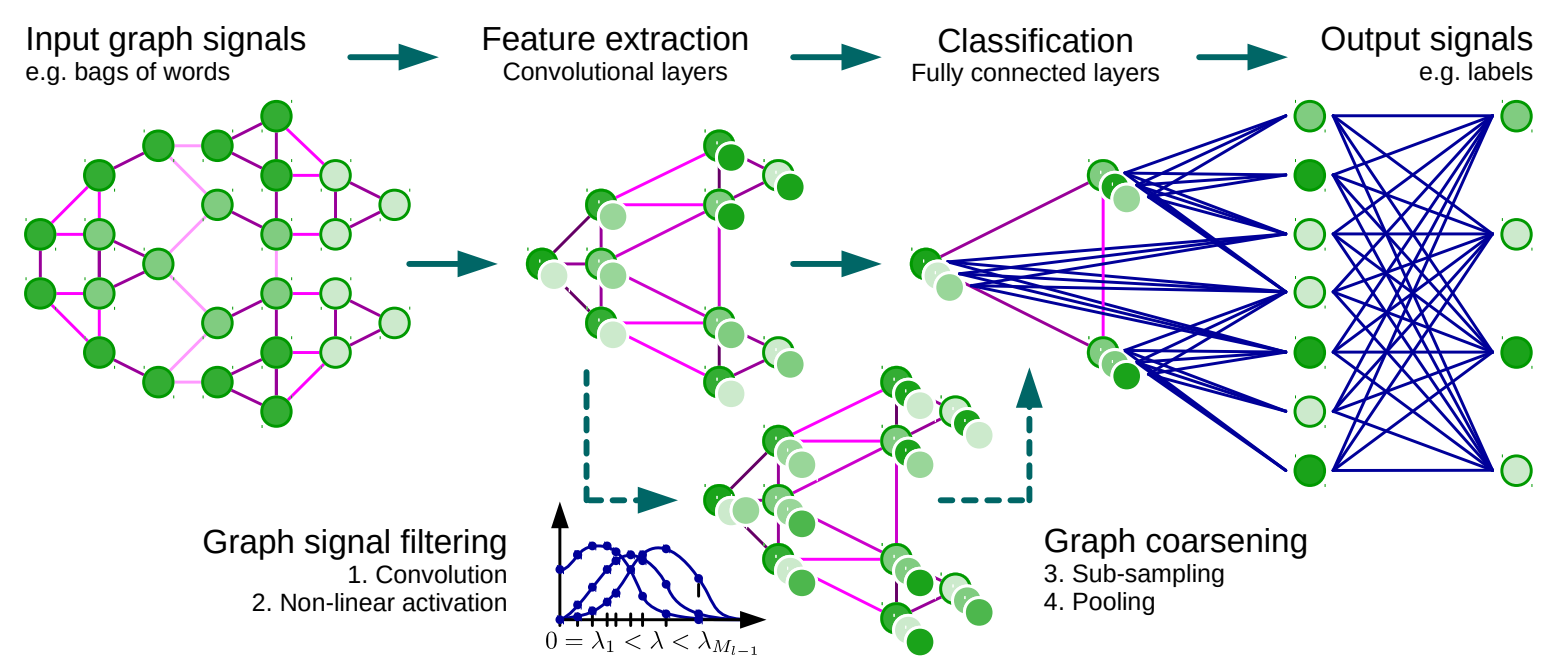}
\caption{
Architecture of a CNN applied to graphs and the four ingredients of a graph convolutional layer. Image adapted from~\cite{defferrard2016convolutional}.
}
\label{fig:Fig7}
\vspace{-6pt}
\end{figure}

The first step in a graph classification task is to transform the raw data into a graph representation. Then, the GCN describes the intrinsic relationships between different nodes of the graph. 
A graph pooling layer in the GCN pools information from multiple vertices to one vertex, to reduce the graph size and expand the receptive field of the graph signal filters. The feature vectors from the last graph convolutional layer are concatenated into a single feature vector, which is fed to a fully connected layer to obtain classification results. This framework is depicted in Fig.~\ref{fig:Fig7}.

GCNs can be categorised as: spectral-based~\cite{defferrard2016convolutional,kipf2017semi} and spatial-based~\cite{niepert2016learning,hamilton2017inductive}.
Spectral-based GCNs rely on the concept of spectral convolutional neural networks, that build upon the graph Fourier transform and the normalized Laplacian matrix of the graph. Spatial-based GCNs define a graph convolution operation based on the spatial relationships that exist among the graph nodes. 

Based on the original graph neural networks proposed in~\cite{scarselli2008graph}, we introduce the most representative GNN variants that have been proposed for several clinical applications.

\vspace{-6pt}
\subsection{Spectral-GCNs}
The convolution operation is defined in the Fourier domain by computing the eigendecomposition of the graph Laplacian~\cite{bruna2013spectral}. The normalized graph Laplacian is defined as $L=I_N-D^{-1/2}AD^{-1/2}=U \Lambda U^T$ ($D$ is the degree matrix and $A$ is the adjacency matrix of the graph), where the columns of $U$ is the matrix of eigenvectors and $\Lambda$ is a diagonal matrix of its eigenvalues. The operation can be defined as the multiplication of a signal $x \in \mathbb{R}^N$ (a scalar for each node) with a filter $g_{\theta}=\text{diag}(\theta)$, parameterized by $\theta \in \mathbb{R}^N$,
\begin{equation}
g_{\theta} \star x =  U g_\theta (\Lambda) U^Tx
\end{equation}

\subsubsection{ChebNet}
In GCN a Chebyshev polynomial $T_m(x)$ of order $m$ evaluated at $\tilde{L}$ is used~\cite{defferrard2016convolutional} and the operation is defined as,
\begin{equation}
g_{\theta} \star x \approx \sum_{m=0}^{M-1} \theta_m T_m (\tilde{L})x ,
\label{eq:eq2}
\end{equation}
where $\tilde{L}$ is a diagonal matrix of scaled eigenvalues defined as $\tilde{L}=\nicefrac{2L}{\lambda_{\text{max}}}-I_N$. $\lambda_{\text{max}}$ denotes the largest eigenvalue of $L$. The Chebyshev polynomials are defined as $T_m(x)=2xT_{k-1}(x)-T_{k-2}(x)$ with $T_0(x)=1$ and $T_1(x)=x$. By introducing Chebyshev polynomials, ChebNet is not required to calculate the eigenvectors of the Laplacian matrix, and that reduces the computational cost.
Such an architecture has been proposed in the medical domain for the analysis of emotions~\cite{jang2018eeg}.

\subsubsection{GCN}
By reducing the size of the convolution filter $K=1$ to alleviate the problem of overfitting to the local neighborhood structure of graphs with a very wide node degree distribution~\cite{kipf2017semi}, and a further approximation $\lambda \approx 2$, Equation~\ref{eq:eq2} can be simplified to,
\begin{equation}
g_{\theta} \star x \approx \theta_0^{'}x + \theta_1^{'}x (L-I_N)x = \theta_0^{'}x + \theta_1^{'} D^{-1/2}AD^{-1/2}x
\end{equation}
Here, $\theta_0^{'}, \theta_1^{'}$ are two unconstrained variables. After adding constraints such that $\theta_0^{'} = -\theta_1^{'} =  \theta $ is obtained, 
\begin{equation}
g_{\theta} \star x \approx \theta_0 (I_N + D^{-1/2} A D^{-1/2} ) x
\end{equation}

Stacking this operation will cause numerical instabilities and the explosion or disappearance of gradients. 
Thus, Kipf and Welling~\cite{kipf2017semi} generalize the definition to a signal $X \in \mathbb{R}^{NXC} $ with $C$ input channels and $F$ filters for feature maps as follows,
\begin{equation}
Z = \tilde{D}^{-1/2} \tilde{A}\tilde{D}^{-1/2} X \Theta ,
\label{eq:eq3}
\end{equation}
where $\Theta \in \mathbb{R}^{CXF}$ is the matrix formed by the filter bank parameters, and $Z \in \mathbb{R}^{NXF}$ is the signal matrix obtained by convolution.

Other GNN variants introduced or adopted by methods analysed in this review are: 
\begin{itemize}
    \item {GCN with dynamic weights}~\cite{song2018eeg}.
    \item {Dynamic GCN with broad learning systems}~\cite{wang2018eeg,zhang2019gcb}.
    \item {Edge weights}~\cite{azevedo2020deep2}.
    \item {Adaptive graph convolutional network}~\cite{gopinath2019adaptive}.
    \item {Graph domain adaptation}~\cite{gopinath2020graph}.
    \item {Isomorphism graph-based model}~\cite{kim2020understanding,li2019graph}.
    \item {Synergic GCN}~~\cite{yang2019classification,zhang2017classification}.
    \item {Simple graph convolution network}~\cite{wu2019simplifying,rakhimberdina2019linear,zhong2020eeg}.
    \item {Graph-based segmentation models} (\textit{e.g.} 3D Unet-graph ~\cite{noh2020combining,juarez2019joint}, Spherical Unet~\cite{hao2020automatic,zhao2019spherical}).
\end{itemize}

\vspace{-6pt}
\subsection{Graph networks with temporal dependency}

GNNs have primarily been developed for static graphs that do not change over time. However, several real-world graphs are dynamic and evolve over time; for example, brain activity recorded using \textit{f}MRI. This variant of GNNs known as dynamic graphs aim to learn hidden patterns from the spatial and temporal dependencies of a graph. 
These models can be divided into two main types:
\begin{itemize}
    \item RNN-based approaches: These methods capture spatio-temporal dependencies by using graph convolutions to filtering inputs and hidden states passed to a recurrent unit.
    \item CNN-based approaches: These approaches tackle spatial–temporal graphs in a non-recursive manner. They use temporal connections to extend static graph structures so that they can apply traditional GNNs on the extended graphs.
\end{itemize}

\subsubsection{RNN-based approaches} 
The aim of these models is to learn node representations with recurrent neural architectures (RNNs). They assume a node in a graph constantly exchanges information/messages with its neighbors until a stable equilibrium is reached.
In a deep learning model, RNNs introduce the notion of time by including recurrent edges that span adjacent time steps~\cite{lipton2015critical}. RNNs perform the same task for every element of a sequence, with the output being dependant on the previous computations and is therefore termed recurrent. LSTMs~\cite{greff2016lstm} were proposed to increase the flexibility of RNNs by employing an internal memory, termed the cell state, to address the vanishing gradient problem. Three logic gates are also introduced to adjust the cell state and produce the LSTM output. GRUs~\cite{cho2014learning} are a variant of LSTMs which combine the forget and input gates, simplifying the model. 

\textit{DCRNN model}:
Diffusion convolutional recurrent neural networks (DCRNN)~\cite{li2018diffusion} introduce the diffusion graph convolutional layer to capture spatial dependencies, and uses a sequence-to-sequence architecture with GRUs to capture temporal dependencies. 
A DCRNN uses a graph diffusion convolution layer to process the inputs of a GRU such that the recurrent unit receives historic information from the last time step as well as neighbourhood information from the graph convolution.
The advantage of a DCRNN is its ability to handle long-term dependencies because of the recurrent network architectures.

Given a graph $G=(\mathcal{V},\mathcal{E},W)$, the diffusion convolution operation that models the spatial dependencies over a graph signal $X \in \mathbb{R}^{N \times P}$ with $N$ nodes and $P$ input features and a convolution filter $f_{\theta}$ is defined as, 
\vspace{-3pt}
\begin{equation}
\resizebox{0.43\textwidth}{!}{$X_{:,p*G}f_\theta = \sum_{k=0}^{K-1} (\theta_{k,1}(D_O^{-1}W)^k + \theta_{k,2}(D_I^{-1}W^T)^k) X_{:,p} \text{for} p \in \{1,...,P\}$} ,
\label{eq:eq1}
\end{equation}
where $D_O^{-1}W$ and $D_I^{-1}W^T$ are the state transition matrices of the outward and inward diffusion processes respectively, and $K$ is the number of maximum diffusion steps.

To model the temporary dependency, the matrix multiplications in the GRU are replaced with a diffusion convolution, which leads to the diffusion convolutional gate recurrent unit (DCGRU) represented as, 
\vspace{-4pt}
\begin{equation}
\begin{split}
r_t = \sigma (\Theta_{r*G}[X_t,H_{t-1}]+b_r) \\
u_t = \sigma (\Theta_{u*G}[X_t,H_{t-1}]+b_u) \\
C_{t} = \text{tanh} (\Theta_{C*G} [X_t,(r_t \odot H_{t-1})]+b_C) \\
H_{t} = u_{t} \odot H_{t-1} + (1-u_t) \odot C_{t} ,
\end{split}
\end{equation}
where $r_T$ and $u_t$ represent the gating functions: reset and update, respectively; $*G$ denotes the diffusion convolution defined in Equation~\ref{eq:eq1}; $\Theta_r$, $\Theta_u$, $\Theta_C$ are the parameters for the corresponding convolutional filters, and $X_t$, $H_t$ corresponds to the input and output of DCGRU at time $t$, respectively. Finally, the DCGRU can be used to build recurrent neural network layers and be trained using backpropagation through time.
Such RNN-based approached coupled with GNNs have been implemented for emotions analysis~\cite{liu2019sparse}.

\textit{GCRN model}:
The graph convolutional recurrent network (GCRN)~\cite{seo2018structured} combines an LSTM network with ChebNet. A dynamic graph consists of time-varying connectivity among ROIs, and temporal information are handled by using LSTM units. To this end,  matrix multiplication operators in the traditional LSTM replaced with the graph convolution which is presented in Equation~\ref{eq:eq3}, the gates (G) of the $t-th$ hidden cell of the graph convolution LSTM follow these formulas,

\vspace{-7pt}
\begin{footnotesize}
\begin{equation}
\begin{split}
f_t = \sigma (w_{xf} * x_t + w_{hf} * H_{t-1} + w_{Cf} \odot C_{t-1} + b_f) \\
i_t = \sigma (w_{xi} * x_t + w_{hi} * H_{t-1} + w_{Ci} \odot C_{t-1} + b_i) \\
C_{t} = f_t  \odot C_{t} + i_t \odot \text{tanh} (w_{xc} * x_t + w_{hc} * H_{t-1} +b_C ) \\
o_t = \sigma (w_{xo} * x_t + w_{ho} * H_{t-1} + w_{Co} \odot C_{t-1} + b_o) \\
H_{t} = o_{t} \odot \text{tanh} (C_{t}) , 
\end{split}
\end{equation}
\end{footnotesize}
where $f_t$, $i_t$, $C_{t}$ and $o_t$ correspond to the forget gate, input gate, memory cell, and output gate, respectively. $*$ denotes the graph convolution operator, $x_t$ the  $t-th$ input of the time series,  $\sigma$ the activation function, and $w-s$ and $b-$ are the graph convolutional kernel weights and biases.
Such a framework has been used in~\cite{xing2019dynamic} and ~\cite{yin2020eeg} for Alzheimer's disease and emotion classification, respectively.

\subsubsection{CNN-based approaches} 

Although RNN-based models are widely used for time series analysis, they still suffer from time-consuming iterations, complex gate mechanisms, and slow response to dynamic changes. CNN-based approaches operate with  fast training, stable gradients and low memory requirements~\cite{gehring2017convolutional}. These approaches interleave 1D-CNN layers with graph convolutional layers to learn temporal and spatial dependencies, respectively.

\begin{figure}[t]
\centering
\includegraphics[width=0.8\linewidth]{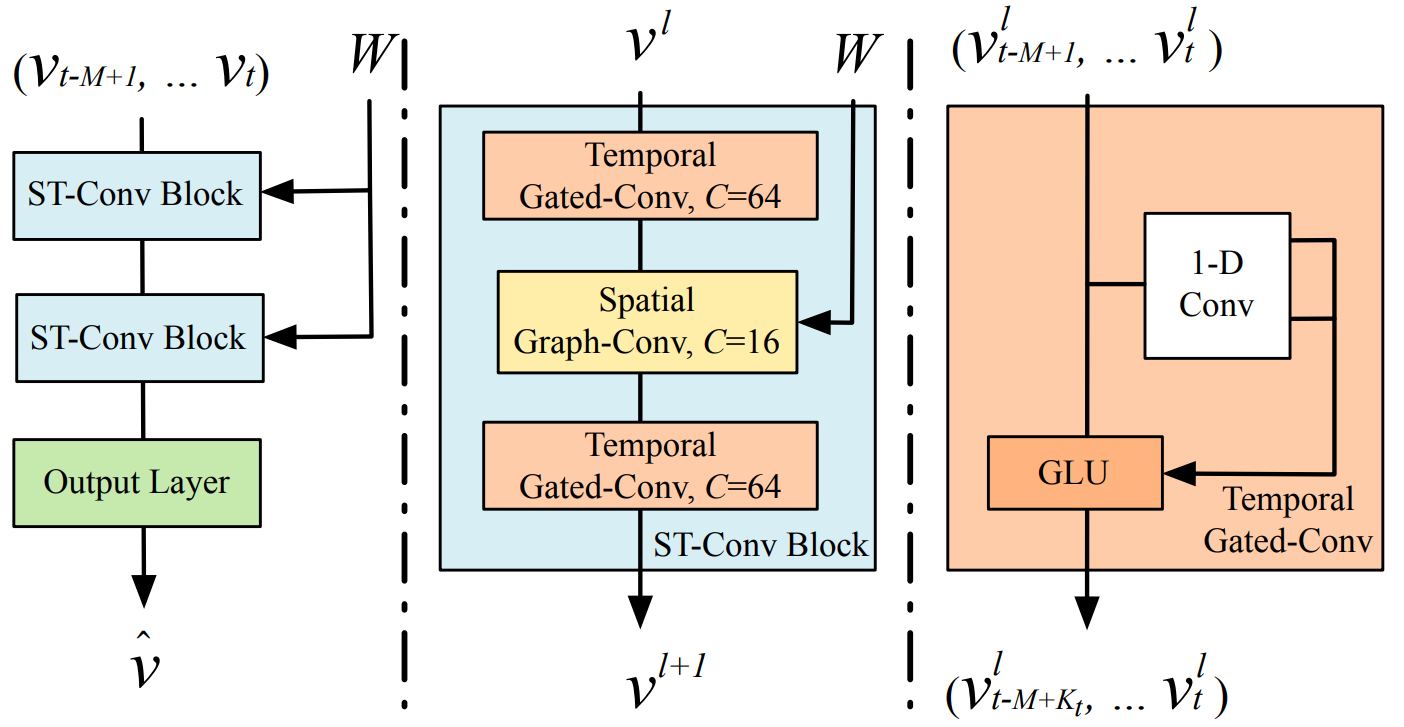}
\vspace{-2pt}
\caption{
STGCN contains multiple spatio-temporal convolutional blocks, each convolutional block uses two temporal gated convolutional layers with a spatial graph convolutional layer sandwiched between them. Image adapted from~\cite{yu2017spatio}.
}
\label{fig:Fig34}
\vspace{-6pt}
\end{figure}

\textit{STGCN model}:
The spatio-temporal graph convolutional network proposed by Yu et al.~\cite{yu2017spatio} employed convolutional structures on the time axis to capture dynamic temporal behaviors. This model integrates a 1-D convolutional layer with ChebNet or GCN layers. 
Fig.~\ref{fig:Fig34} illustrates the STGCN framework that consists of two spatio-temporal convolutional blocks and a fully connected output layer. Each spatio-temporal convolutional block stacks a gated 1-D convolutional layer, a graph convolutional layer, and another gated 1-D convolutional layer sequentially.

\begin{figure}[t]
\centering
\includegraphics[width=0.6\linewidth]{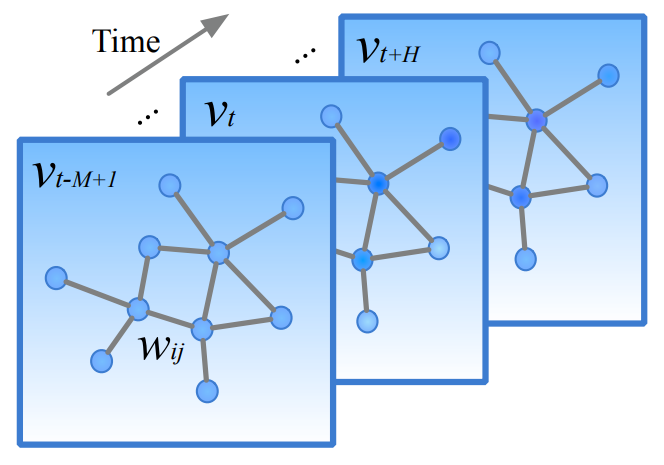}
\vspace{-2pt}
\caption{
An example of spatial temporal graph structure. Each $v_t$ indicates a frame of the current graph state at time $t$. Image adapted from~\cite{yu2017spatio}.
}
\label{fig:Fig12}
\vspace{-6pt}
\end{figure}

As illustrated in Fig.~\ref{fig:Fig12}, the observation $v_t$ is independent but linked by a pairwise connection in the graph. Therefore, the data point $v_t$ can be regarded as a graph signal that is defined on an undirected graph (or a directed graph) $G$ with weights $w_{ij}$.

The temporal convolution layer contains 1-D causal convolutions with a width-$K_t$ kernel, followed by gated linear units as a non-linearity as illustrated in Fig.~\ref{fig:Fig34} (right). The convolution kernel $\Gamma \in \mathbb{R}^{K_t \times C_i \times 2C_o}$ is designed to map the input $Y$ to a single output element $[PQ] \in \mathbb{R}^{M-K_t+1} \times (2C_o$. Thus, the temporal gated convolution can be defined as, 
\vspace{-2pt}
\begin{equation}
\Gamma * \tau Y = P \odot \sigma (Q) \in \mathbb{R}^{M-K_t+1}  ,
\end{equation}
where $P,Q$ are inputs of the gates in the gated linear units respectively, and $\odot$ indicates the element-wise Hadamard product. The sigmoid gate $\sigma (Q)$ controls which inputs $P$ of the current states are relevant for discovering compositional structure and dynamic variances in the time series.

The spatio-temporal convolutional block, which fuses features from both the spatial and temporal domains, is constructed to jointly process graph-structured time series data as depicted in Fig.~\ref{fig:Fig34} (mid). The input and output of the spatio-temporal convolutional block are all 3-D tensors. For the input $v^{l+1} \in \mathbb{R}^{M \times n \times C^l}$ of block $l$, the output $v^{l+1} \in \mathbb{R}^{(M-2(K_t-1)) \times n \times C^{l+1}}$ is computed by,
\begin{equation}
v^{l+1} = \Gamma_1^{l} * \tau Y ReLU ( \Theta^{l} * G (\Gamma_0^{l} * \tau Y v^l)) ,
\end{equation}
where $\Gamma_1^{l},\Gamma_0^{l}$ are the upper and lower temporal kernel within block $l$, respectively; $\Theta^{l}$ is the spectral kernel of graph convolution. 
Such adoption of CNNs to perform a convolution operation in the temporal dimension has been used for sleep state classification~\cite{jia2020graphsleepnet}.

\textit{ST-GCN model}:
ST-GCN are popular for solving problems that base predictions on graph-structured time series~\cite{yan2018spatial}. The main benefits of temporal GCN are that it uses a feature extraction operation that is shared over time and space.

The input to the ST-GCN is the joint coordinate vectors on the graph nodes. 
Multiple layers of spatio-temporal graph convolution operations process the input data and higher-level feature maps on the graph. The resultant classification is performed using a conventional dense layer and activation.

To represent the functional networks, let $G=(\mathcal{V},\mathcal{E})$ be an undirected spatio-temporal graph with $N$ ROIs, $T$ time points and, $\mathcal{E}$ temporal and spatial connections between a set of nodes $\mathcal{V} = \{ v_{t,i} | t=1,...,T; i=1,...,N  \}$. 
Thus, the temporal aspect of the graph is constructed by connecting the same ROI at the preceding time point. All nodes of the same time point are connected through the edges of the spatial graph, where the weight of an edge is determined by the functional affinity between the corresponding regions. The affinity between two regions $d(v_{tj},v_{ti})$ is defined as the magnitude of correlation between their concatenated series. 
Then, given $f_{in}(v_{ti})$ as the input feature at node $v_{ti}$, the spatio-temporal neighborhood $B(v_{ti})$ is defined as, 
\begin{equation}
B(v_{ti}) = \{ v_{qj} | d(v_{tj},v_{ti}) \leq K, |q-t| \leq [\Gamma/2]  \}  ,
\end{equation}
where the parameter $\Gamma$ controls the temporal range to be included in the neighbor graph (\textit{i.e.} temporal kernel size), and $K$ is the size of the spatial neighborhood (\textit{i.e.} the spatial kernel size).

At point $t$, the edge connection is defined by the adjacency matrix $A$ and an identity matrix $I$ representing self-connections, and the spatial graph convolution is defined with respect to the diagonal matrix $\Lambda$,
\begin{equation}
f_{t}^{'} = \Lambda^{1/2} (A + I) \Lambda^{1/2} f_{t} W_SG  ,
\end{equation}
where $\Lambda^{ii}= \sum_{j} A^{ij} + 1 $ and $W_{SG} \in \mathbb{R}^{C \times M}$  represents the spatial graph convolutional kernel. Then, the temporal convolution is performed on the resulting features. 
Given $f_{i}^{'} \in \mathbb{R}^{M \times T}$ (the features of node $v_i$ defined on the temporal graph of length $T$), and $W_{TG} \in \mathbb{R}^{M \times t}$ (a temporal convolutional kernel), a standard 1D convolution $f_{i}^{'} \otimes W_{TG} \in \mathbb{R}^{M \times t}$ is performed as the final output for $v_i$.
The work from~\cite{gadgil2020spatio2} is an example of applications of this model for gender classification.

\textit{TGCN model}:
Traditional temporal convolutional neural networks (TCNN) show that variations of convolutional neural networks can achieve impressive results for sequential data~\cite{bai2018empirical}. TCNNs use dilated causal convolutional layers where an output at time $t$ is convolved only with elements from time $t$ or earlier in the previous layer, \textit{i.e.}~inputs have no influence on output steps that precede them in time. In a dilated convolutional layer, a filter is sequentially applied to inputs by skipping input values with a pre-defined step (dilatation rate).

Wu et al.~\cite{wu2019graph} proposed a method for multi-resolution modeling of temporal dependencies, their temporal model is based on dilated convolutions. This approach is based on the fact that subsequent layers have dilated receptive fields.

Temporal graph convolutional networks (TGCN) takes structural times series data as input and apply feature extraction operations that are shared over both time and space.
A structural time series is represented as $(X,A)$ where $X \in \mathbb{R}^{TXpXc}$ is a multivariate time series where $T$ is the number of time steps, $p$ is the number of sequences, $c$ is the number of channels, and $A$ is the adjacency matrix. 
At layer $l$, TGCN computes a hidden representation $h^l \in \mathbb{R}^{T^lXpXc^l}$ in a hierarchical manner via the composition of multiple spatio-temporal convolutional layers.
TGCNs show promise in applications such as EEG electrode distributions, where several datasets of similar but not identical configurations need to be analyzed.
Methods including~\cite{covert2019temporal} and ~\cite{azevedo2020deep2} are examples of this approach for epilepsy and gender classification, respectively.

Other dynamic GNN variants adopted and introduced by research analysed in this review include: 
\begin{itemize}
    \item {Traditional fusion of GCN-LSTM}~\cite{yin2020eeg}.
    \item {Sequential GCN based on complex networks}~\cite{wang2020sequential}.
    \item{Approaches based on geometric deep learning} ~\cite{azevedo2020towards,azevedo2020deep}.
    \item {Temporal-adaptive GCN}~\cite{yao2020temporal,li2019classify}.
    \item {GCN with phase-locking value} ~\cite{wang2019phase,wang2020functional}.
\end{itemize}

\vspace{-6pt}
\subsection{Graph networks with attention mechanisms}

In real-world applications, graph-structured data can be both massive and noisy, and not all portions of the signal are equally important. As such, attention mechanisms can direct a network to focus on the most relevant parts of the input, suppressing uninformative features, reducing computational cost and enhancing accuracy. Attention mechanisms are beneficial as they allow for dealing with variable-sized inputs. Furthermore, attention provides a tool for interpreting the results given by the network and discovering the underlying dependencies that have been learnt.
Attention mechanisms are established in neuroscience and can be divided into two main types: soft-attention and self-attention mechanisms.

\subsubsection{Soft-attention mechanisms}

Soft-attention mechanisms allows the model to learn the most relevant parts of the input sequence during training and are often placed between encoders and decoders. Soft-attention mechanisms are end-to-end approaches that can be learned by gradient-based methods~\cite{yang2016hierarchical} 
A full-attention architecture can preserve the details from raw signals, and select the most crucial information. Each layer of the graph is connected to an attention layer, and all attention layers are jointly trained with the network, as per the approach introduced for predicting human motor intentions~\cite{jia2020attention}.
The attention mechanism can be formulated as follows,
\vspace{-4pt}
\begin{equation}
\begin{split}
u_t = \tanh ( W h_t + b), \\
\alpha_t = \dfrac{\exp(u_t^Tu_w)}{\sum_{j=1}^{n} \exp(u_t^Tu_w)} , \\
s_t = \sum_{t} \alpha_t h_t ,
\end{split}
\end{equation}
where $h_t$ is the output of each layer; $W$, $u_w$ and $b$ are trainable weights and bias. The importance of each element in $h_t$ is measured by estimating the similarity between $u_t$ and $h_t$, which is randomly initialized. $\alpha_t$ is a softmax function. The scores are multiplied by the hidden states to calculate the weighted combination, $s_t$ (attention-based final output).

Graph attention structures can also consist of two branches: a trunk branch extracts global features and an attention branch selects useful input channels.
The attention branch uses one graph convolutional layer to generate an attention vector $T \in \mathbb{R}^{N x 1}$ ($N$ vertex), which is formulated as follows, 
\begin{equation}
T = \phi (X, A_2) = [\tau_1,...,\tau_n]^T  ,
\end{equation}
where $A_2$ denotes the adjacency matrix used in the attention branch, $\phi(\cdot)$ denotes the graph convolution procedure, and $\tau_i$ indicates the contribution of the $i-th$ node to the classification task. A softmax is adopted on $T$ to generate a normalized attention vector $\tilde{T}$.

The output of the graph attention structure can be obtained by weighting the graph convolution results of each node with the corresponding weight parameters in the attention vector. Thus, $\tilde{T}$ is expanded to a diagonal matrix $\text{diag}(\tilde{T}) \in \mathbb{R}^{N x N}$. Let $f_{GA}$ denote the output of the graph attention, then the weighted procedure can be formulated as follows, 
\begin{equation}
f_{GA} = \text{diag} \cdot  f_{GCN}(X, A_1) ,
\end{equation}
where $A_1$ denotes the adjacency matrix of the trunk branch.
Soft-attention mechanisms have been used for emotion~\cite{liu2019sparse} analysis.

\subsubsection{Self-attention mechanisms}

Recent research in self-attention mechanisms~\cite{vaswani2017attention} indicates that models that rely entirely on attention computations without using convolution or recurrent architectures can achieve similar performance.
Inspired by this mechanism, graph attention networks (GAT)~\cite{velivckovic2017graph} incorporates the attention mechanism into the propagation steps by modifying the convolution operation. In a traditional GCN the weights typically depend on the degree of the neighboring nodes, while in GATs the weights are computed by a self-attention mechanism based on node features.
Veli{\v{c}}kovi{\'c} et al.~\cite{velivckovic2017graph} constructed a graph attention network by stacking a single graph attention layer, $a$, which is a single-layer feedforward neural network, parametrized by a weight vector $\vec{a} \in \mathbb{R}^{2F^{i}}$. The layer computes the coefficients in the attention mechanisms of the node pair $(i,j)$ by,
\begin{equation}
\alpha_{i,j} = \frac{ \text{exp} (\text{LeakyReLu} ( \vec{a}^T [W\vec{h}_i \mathbin\Vert W\vec{h}_j] ) ) }
{ \sum_{k \in N_i \mathbb{N} }  \text{exp} (\text{LeakyReLu} ( \vec{a}^T [W\vec{h}_i \mathbin\Vert  W\vec{h}_k] ) ) } ,
\end{equation}
where $\mathbin\Vert$ represents the concatenation operation. The attention layer takes as input a set of node features $h=\{\vec{h_1},\vec{h_2},...,\vec{h_N}\}, \vec{h_i} \in R^F$, where $N$ is the number of nodes of the input graph and $F$ the number of features for each node, and produces a new set of node features $h^{'}=\{\vec{h_1}^{'},\vec{h_2}^{'},...,\vec{h_N}^{'}\}, \vec{h_i}^{'} \in R^F$ as its output.
To generate higher-level features, as an initial step a shared linear transformation, parametrized by a weight matrix $W \in R^{F'*F}$ is applied to every node and subsequently a masked attention mechanism can be applied to every node, resulting in the following scores,
\begin{equation}
e_{ij} = a ( W \vec{h_i}, W \vec{h_j} ),
\end{equation}
that indicates the importance of node $j^{'}s$ features to node $i$. The final output feature of each node can be obtained by applying a non-linearity, $\sigma$,
\begin{equation}
h_i^{'} = \sigma ( \sum_{j \in N_i} \alpha_{ij} Wh_j ),
\end{equation}

\begin{figure}[t!]
\centering
\includegraphics[width=0.8\linewidth]{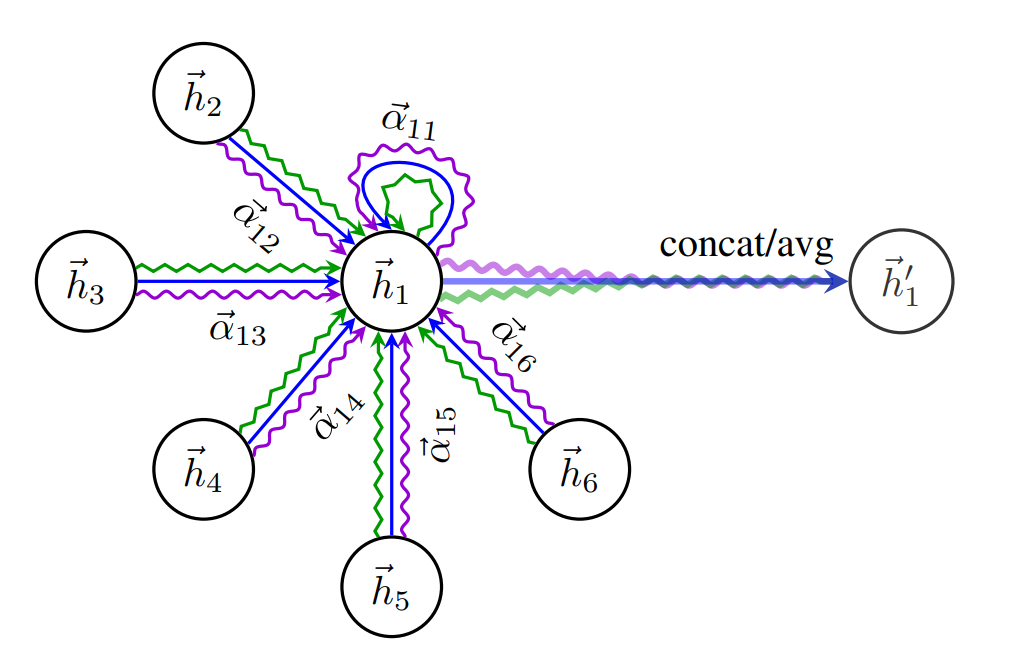}
\vspace{-6pt}
\caption{
An illustration of the process of generating output features through multiple attention heads. Each color denotes an independent attention vector. Image adapted from~\cite{velivckovic2017graph}.
}
\label{fig:Fig2}
\vspace{-6pt}
\end{figure}

The layer also uses multi-head attention to stabilise the learning process. $K$ different attention heads are applied to compute mutually independent features in parallel, and then concatenate their features, resulting in the following representations,
\begin{equation}
h_i^{'} = \Arrowvert_{K=1}^K  \sigma ( \sum_{j \in N_i} \alpha_{ij}^k W^k\vec{h_j} ),
\end{equation}
or by employing averaging and delay applying the final non-linearity (usually a softmax or logistic sigmoid for classification problems),
\begin{equation}
h_i^{'} = \sigma ( \frac{1}{K} \sum_{k=1}^K \sum_{j \in N_i} \alpha_{ij}^k W^k\vec{h_j} ),
\end{equation}
where $\alpha_{ij}^k$ is the normalized attention coefficient computed by the $k$-th attention mechanism.
The aggregation process is illustrated in Fig.~\ref{fig:Fig2}.

GAT based approaches have been used for ASD~\cite{li2020pooling}, gender classification~\cite{filip2020novel}, BD~\cite{yang2019interpretable}, PD~\cite{mcdaniel2019developing} and medical image enhancement~\cite{hu2020feedback}.

Other GNNs with attention mechanisms adopted and introduced by works discussed in this review are: 
\begin{itemize}
    \item {Attention mechanisms for feature representation}~\cite{lian2020learning}.
    \item {Attention mechanisms for multimodal fusion}~\cite{chen2020pathomic}.
    \item {Weighted GATs}~\cite{wang2020weighted}.
    \item {Edge-weighted GATs}~\cite{li2020pooling,yang2019interpretable}.
    \item {Attention based ST-GCN}~\cite{jia2020graphsleepnet,guo2019attention}.
    \item {Cross-modality with GAT-based embedding}~\cite{zhang2020deeprep}.
\end{itemize}

\vspace{4pt}
\section{Case studies of GNN for medical diagnostic analysis}
\label{sec:sec3}%

Graph convolutional networks have been utilized in multiple classification, prediction, segmentation and reconstruction tasks with non-structural (\textit{e.g.} \textit{f}MRI, EEG, iEEG) and structural data (\textit{e.g.} MRI, CT). There are several specificities in the usage of GNNs in each of the medical signals identified by our survey that we review in the following sections. 
%
These case studies for medical diagnosis are organised according to the input data and baseline graph framework adopted or proposed with its corresponding application and the dataset. 
Case studies have been divided into four main groups; functional connectivity analysis, electrical-based analysis, and anatomical structure analysis classification/regression and segmentation, which are detailed in Tables~\ref{table:function},~\ref{table:electrical},~\ref{table:anatomical1} and~\ref{table:anatomical2}, respectively. 
Rather than presenting an exhaustive literature review for each studied case, we discuss prominent highlights of how GNNs were used in each case.

\begin{table*}[t!]
\caption{Summary of GCN approaches adopted for functional connectivity and their applications.}
\vspace{-5pt}
\centering
\label{table:function}
\resizebox{1\textwidth}{!}{%
\begin{tabular}{l c l 
>{\raggedright\arraybackslash}p{5cm} 
>{\raggedright\arraybackslash}p{7.5cm}}
\toprule
\textbf{Authors} &
\textbf{Year} &
\textbf{Modality} & 
\textbf{Application} &  
\textbf{Dataset} \\
\midrule
Li et al.~\cite{li2020pooling} $\dagger$ & 2020 & t-\textit{f}MRI &
Classification: Autism disorder & ASD Biopoint Task (Yale Child Study Center~\cite{li2019graph}) (2 classes) \\ %
Li et al.~\cite{li2020braingnn} & 2020 & t-\textit{f}MRI &
Classification: Autism disorder & Biopoint~\cite{venkataraman2016bayesian} (2 classes) \\
Huang et al.~\cite{huang2020edge} & 2020 & rs-\textit{f}MRI &
Classification: Autism disorder & ABIDE~\cite{di2014autism} (2 classes) \\ 
Rakhimberdina et al.~\cite{rakhimberdina2020population} & 2020 & \textit{f}MRI &
Classification: Autism disorder & ABIDE~\cite{di2014autism} (2 classes) \\ 
Li et al.~\cite{li2020graph} & 2020 & t-\textit{f}MRI &
Classification: Autism disorder & Yale Child Study Center~\cite{li2019graph} (2 classes) \\ 
Jiang et al.~\cite{jiang2020hi} & 2020 & \textit{f}MRI &
Classification: Autism disorder & ABIDE~\cite{di2014autism} (2 classes) \\ 
Li et al.~\cite{li2019graph} & 2019 & t-\textit{f}MRI &
Classification: Autism disorder & Yale Child Study Center (private) (2 classes) \\ 
Kazi et al.~\cite{kazi2019inceptiongcn} & 2019 & rs-\textit{f}MRI &
Classification: Autism disorder & ABIDE~\cite{di2014autism} (2 classes) \\ 
Yao et al.~\cite{yao2019triplet} & 2019 & rs-\textit{f}MRI &
Classification: Autism disorder & ABIDE~\cite{di2014autism} (2 classes) \\ 
Anirudh et al.~\cite{anirudh2019bootstrapping} & 2019 & rs-\textit{f}MRI &
Classification: Autism disorder & ABIDE~\cite{di2014autism} (2 classes) \\ 
Rakhimberdina and Murata~\cite{rakhimberdina2019linear} & 2019 & \textit{f}MRI &
Classification: Autism disorder & ABIDE~\cite{di2014autism} (2 classes) \\ 
Ktena et al.~\cite{ktena2018metric} & 2018 & rs-\textit{f}MRI &
Classification: Autism disorder & ABIDE~\cite{di2014autism} (2 classes) \\ 
Parisot et al.~\cite{parisot2018disease} & 2018 & rs-\textit{f}MRI &
Classification: Autism disorder & ABIDE~\cite{di2014autism} (2 classes) \\ 
Ktena et al.~\cite{ktena2017distance} & 2017 & rs-\textit{f}MRI &
Classification: Autism disorder & ABIDE~\cite{di2014autism} (2 classes) \\ 
Parisot et al.~\cite{parisot2017spectral} & 2017 & rs-\textit{f}MRI &
Classification: Autism disorder & ABIDE~\cite{di2014autism} (2 classes) \\ 
%
%
Rakhimberdina and Murata~\cite{rakhimberdina2019linear} & 2019 & \textit{f}MRI &
Classification: Schizophrenia & COBRE~\cite{bullmore2009complex} (2 classes) \\ 
%
%
Rakhimberdina and Murata~\cite{rakhimberdina2019linear} & 2019 & rs-\textit{f}MRI &
Classification: Attention deficit disorder & ADHD-200~\cite{bullmore2012economy} (2 classes) \\ 
Yao et al.~\cite{yao2019triplet} & 2019 & rs-\textit{f}MRI &
Classification: Attention deficit disorder & ADHD-200~\cite{bullmore2012economy} (2 classes) \\ 
%
%
Yao et al.~\cite{yao2020temporal} $\star$ & 2020 & rs-\textit{f}MRI &
Classification: Major depressive disorder & MDD~\cite{yan2019reduced} (2 classes) \\ %
%
Yang et al.~\cite{yang2019interpretable} $\dagger$ & 2019 & \textit{f}MRI / sMRI &
Classification: Bipolar disorder & BD (private) \\
%
%
Zhang et al.~\cite{zhang2020deeprep} $\dagger$ & 2020 & \textit{f}MRI / MRI &
Classification: Gender & HCP S1200~\cite{van2013wu} (2 classes) \\
Kim et al.~\cite{kim2020understanding} & 2020 & rs-\textit{f}MRI  &
Classification: Gender & HCP S1200~\cite{van2013wu} (2 classes) \\
Filip et al.~\cite{filip2020novel} $\dagger$ & 2020 & \textit{f}MRI &
Classification: Gender & HCP S1200~\cite{van2013wu} (2 classes) \\
Gadgil et al.~\cite{gadgil2020spatio2} $\star$ & 2020 & rs-\textit{f}MRI  &
Classification: Gender & HCP S1200~\cite{van2013wu} (2 classes), NCANDA~\cite{brown2015national} (2 classes) \\
Azevedo et al.~\cite{azevedo2020towards} $\star$ & 2020 & rs-\textit{f}MRI  &
Classification: Gender & HCP S1200~\cite{van2013wu} (2 classes)\\
Azevedo et al.~\cite{azevedo2020deep} $\star$ & 2020 & rs-\textit{f}MRI  &
Classification: Gender & HCP S1200~\cite{van2013wu} (2 classes)\\
Azevedo et al.~\cite{azevedo2020deep2} $\star$ & 2020 & rs-\textit{f}MRI  &
Classification: Gender & UK Biobank~\cite{bycroft2018uk} (2 classes)\\
Arslan et al.~\cite{arslan2018graph} & 2018 & rs-\textit{f}MRI  &
Classification: Gender & UK Biobank~\cite{bycroft2018uk} (2 classes) \\
Ktena et al.~\cite{ktena2018metric} & 2018 & rs-\textit{f}MRI &
Classification: Gender & UK Biobank~\cite{sudlow2015uk} (2 classes) \\ 
%
%
Li et al.~\cite{li2020braingnn} & 2020 & rs-\textit{f}MRI &
Classification: Brain response stimuli & HCP 900~\cite{van2013wu} (7 classes) \\
Zhang et al.~\cite{zhang2019functional} & 2019 & \textit{f}MRI  &
Classification: Brain response stimuli & HCP S1200~\cite{van2013wu} (21 classes) \\
Guo et al.~\cite{guo2017deep} & 2017 & MEG  &
Classification: Brain response stimuli & Visual stimulus (private) (2 classes) \\
%
%
Isallari et al.~\cite{isallari2020gsr} & 2020 & \textit{f}MRI &
Regression: High-resolution connectome & SLIM~\cite{liu2017longitudinal} \\ %
\bottomrule
\multicolumn{5}{p{350pt}}
{ 
$\star$ GCN with temporal structures for medical diagnostic analysis. \newline
$\dagger$ GCN with attention structures for medical diagnostic analysis.
}
\end{tabular}}
\vspace{-6pt}
\end{table*}

\vspace{-6pt}
\subsection{Functional connectivity analysis}

This section mainly covers application of graph learning representation on functional brain connectivity, as with the best of our knowledge there are no applications that involved other body functions in the reviewed literature.

\subsubsection{Autism spectrum disorder}

Autism spectrum disorder (ASD) is a complex neurodevelopmental disorder characterized by recurring difficultines in social interaction, speech and nonverbal communication, and restricted/repetitive behaviours. The screening of ASD is challenging due to uncertainties associated with its symptoms~\cite{mastrovito2018differences}. Resting-state \textit{f}MRI (rs-\textit{f}MRI) and task \textit{f}MRI are the main modalities which are used to classify the population into ASD or health control (HC) groups. 

\begin{figure}[!t]
\centering
\includegraphics[width=1\linewidth]{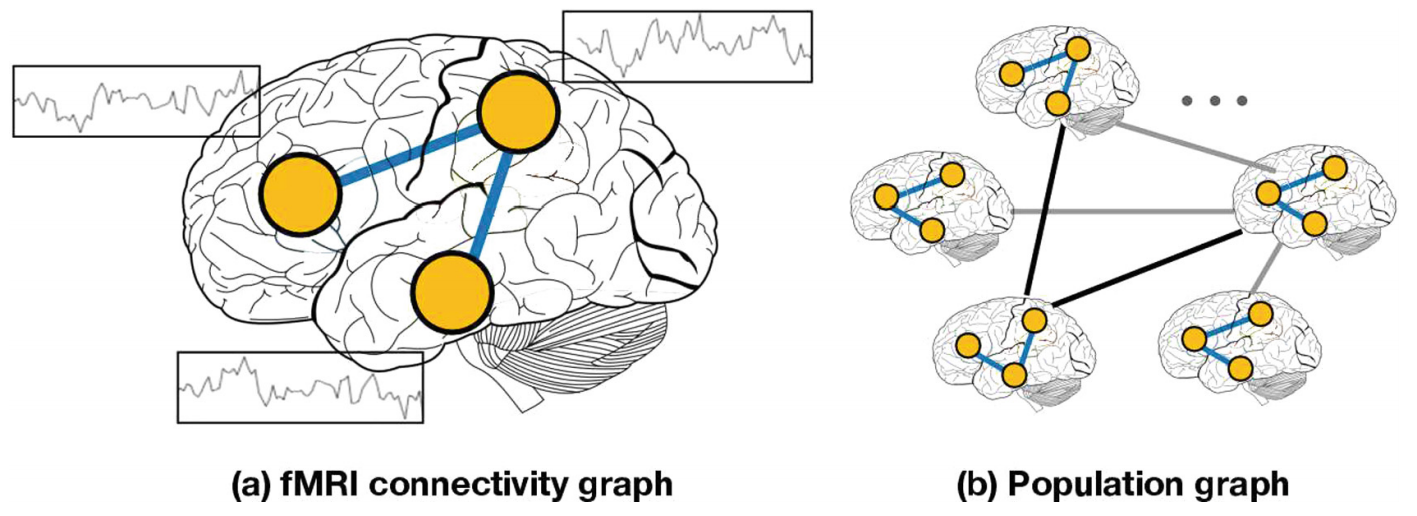}
\vspace{-10pt}
\caption{
Proposed graph-based approaches for modeling with rs-\textit{f}MRI data. Image taken from~\cite{rakhimberdina2019linear}.
}
\label{fig:Fig27}
\vspace{-6pt}
\end{figure}

The rapid development of GNNs has attracted interest in using these architectures to analyse \textit{f}MRI and non-imaging data for disease classification. Graph-based models can be classified into two groups based on the node definition as illustrated in Fig.~\ref{fig:Fig27}:
(a) Individual graph: nodes are brain regions and edges are functional correlations between time series observations from those regions. Therefore, each graph represents only one subject and graph comparison metrics are computed to analyse these graphs, which are represented in the left panel in Fig.~\ref{fig:Fig27}; 
(b) Population graph: in this approach each node represents a subject with corresponding brain-connectivity data, and edges are determined as the similarity between subjects’ phenotypic features (age, gender, handedness, etc.), as is shown in the right panel in Fig.~\ref{fig:Fig27}.

\paragraph{Individual-based graph methods}

Ktena et al.~\cite{ktena2017distance} proposed a GNN method to learn a similarity (distance) metric between irregular graphs, such as the functional connectivity graphs obtained from the Autism Brain imaging Data Exchange (ABIDE) dataset~\cite{di2014autism}, to classify individuals as autism spectrum disorder (ASD) or healthy controls (HC). 

\begin{figure}[!t]
\centering
\includegraphics[width=1\linewidth]{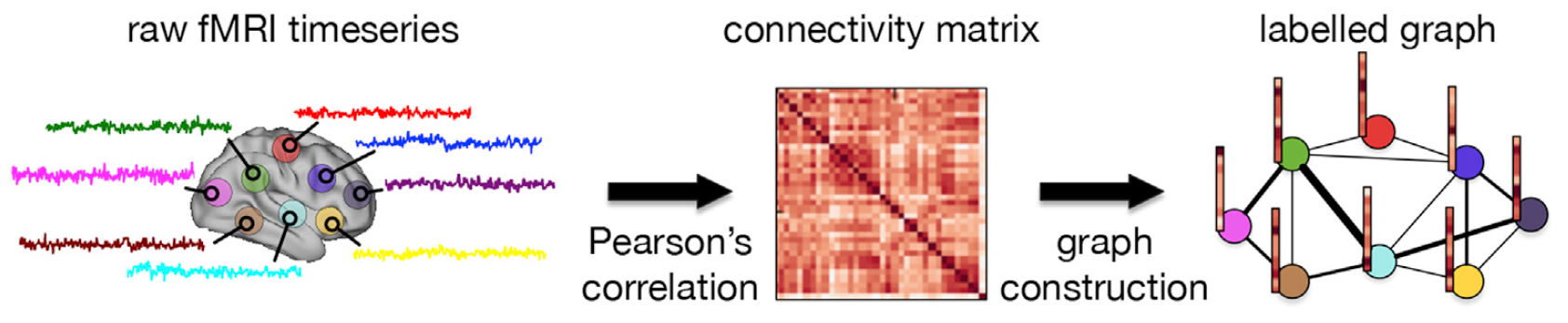}
\vspace{-10pt}
\caption{
Estimation of single subject connectivity matrix and labelled graph representation. Pearson's correlation coefficient is used to obtain a functional connectivity matrix from the raw \textit{f}MRI time series. Image taken from~\cite{ktena2018metric} .
}
\label{fig:Fig8}
\vspace{-6pt}
\end{figure}

The method of Ktena et al.~\cite{ktena2018metric} is based on their previous work~\cite{ktena2017distance} to learn a graph similarity metric in spectral graph domain obtained from brain connectivity networks via supervised learning. They applied their method to individual graphs constructed from the ABIDE database to classify subjects into ASD or HC. The graph construction is illustrated in Fig.~\ref{fig:Fig8}. They showed their spectral graph matching method not only outperforms non-graph matching, but is also superior to individual subject classification and manifold learning methods. 

The graph similarity metric proposed by Ktena et al.~\cite{ktena2018metric} using a specific template for brain region of interest (ROI) parcellation could impose a limitation such as analysis of single spatial scale (\textit{i.e.}, a fixed graph). 
Yao et al.~\cite{yao2019triplet} dealt with this limitation by proposing a multi-scale triplet GCN. They constructed multi-scale functional connectivity patterns for each subject through multi-scale templates for coarse-to-fine ROI parcellation. A triple GCN model was designed to learn multi-scale graph features of brain networks. Their application on \textit{f}MRI data obtained from the ABIDE dataset showed their high performance in ASD and HC classification. 

For GCN methods, all nodes are required to be presented during training which result in low performance on unseen nodes. Li et al.~\cite{li2019graph} proposed a GCN algorithm to discover ASD brain biomarkers from t-\textit{f}MRI. Different from the semi-supervised spectral GCN algorithm~\cite{kipf2017semi} used in~\cite{parisot2017spectral}, this GCN classifier is isomorphism graph-based which can interpret graphs with different nodes and edges. In other words, the GCN is trained on the whole graph and tested on sub-graphs, such that they could determine the importance of sub-graphs and nodes. 
In both works from Li et al.~\cite{li2020braingnn,li2020pooling}, the authors also improved their individual graph level analysis by proposing a BrainGNN and a pooling regularized GNN model to investigate the brain region related to a neurological disorder from t-\textit{f}MRI data for ASD or HC classification.  

In addition, the low signal-to-noise ratio of \textit{f}MRI and its high dimensionality impose another limitation on using \textit{f}MRI for graph level classification and detection of functional differences between ASD and HC groups. 
Li et al.~\cite{li2020graph} dealt with this challenge by modeling the the whole brain \textit{f}MRI as a graph. This allowed them to preserve the geometrical and temporal information and learn a better graph embedding. They implemented their method on a group of 75 ASD children and 43 age- and IQ-matched healthy controls collected at the Yale Child Study Center~\cite{li2019graph}. Their results indicated a more robust classification of ASD or HC.

\paragraph{Population-based graph methods}

Population graphs have been shown to be effective for brain disorder classification.
Parisot et al.~\cite{parisot2017spectral} investigated the performance of GCN for brain analysis in a population where the authors built a population graph using both rs-\textit{f}MRI and non-imaging data (acquisition information). They applied their model on the ABIDE dataset~\cite{di2014autism} to classify subjects as ASD or HC. Their semi-supervised method showed better performance in comparison to a standard linear classifier (which only considered the individual features for classification). 
In an extension of this work, Parisot et al.~\cite{parisot2018disease} proposed a spectral GCN model which takes into account both the pairwise similarity between subjects (phenotypic information) and information obtained from subject-specific imaging features to classify subjects as ASD or HC in a population.

\begin{figure}[!t]
\centering
\includegraphics[width=1\linewidth]{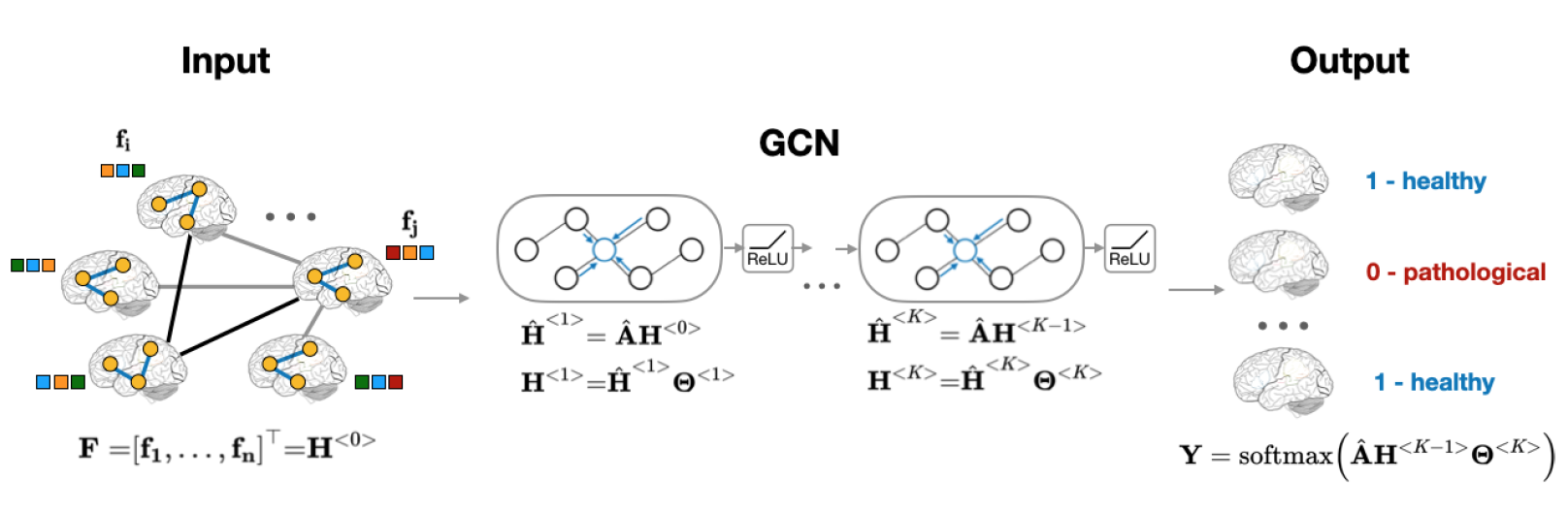}
\vspace{-10pt}
\caption{
Proposed population graph-based approaches for subject classification. Image taken from~\cite{rakhimberdina2019linear}.
}
\label{fig:Fig35}
\vspace{-6pt}
\end{figure}

As illustrated in Fig.~\ref{fig:Fig35}, Rakhimberdina and Murata~\cite{rakhimberdina2019linear} applied a linear simple graph convolution (SGC)~\cite{wu2019simplifying} for brain disorder classification. They construct the population graphs by using the hamming distance between phenotypic features of the subjects as weights of the edges of the graph. Their results on the ABIDE dataset~\cite{di2014autism} showed a high performance and efficiency of the linear SGC over the GCN based model deployed by Parisot et al~\cite{parisot2018disease} on the same dataset. 

As there is no standard method to construct graphs for a GNN, Anirudh et al.~\cite{anirudh2019bootstrapping} proposed a bootstrapped version of GCNs that made models less sensitive to the initialisation of the construction of the population graph. They generated random graphs from the initial population graph (from the ABIDE dataset~\cite{di2014autism}) to train weakly a GCN for ASD and HC classification, and fused their prediction as the final result. 
To avoid the spatial limitation of a single template and learn multi-scale graph features of brain networks, Yao et al.~\cite{yao2019triplet} proposed a multi-scale triplet GCN model. These solutions, however, are problem specific, and choosing a particular graph definition over the other has remained a challenging problem.
Rakhimberdina et al.~\cite{rakhimberdina2020population} proposed a population graph-based multi-model ensemble method to deal with this problem. Their results on the ABIDE dataset~\cite{di2014autism} showed a 2.91\% improvement in comparison to the best result reported for a non-graph solution~\cite{sherkatghanad2020automated}.

The heterogeneity of the graph is challenging. Kazi et al.~\cite{kazi2019inceptiongcn} proposed Inception-GCN as a spectral domain architecture for deep learning on graphs for node-level classification of disease prediction. This inception graph model is capable of capturing intra- and inter-graph structural heterogeneity during convolutions. The Inception-GCN could improve the performance of node classification in comparison to Parisot~\cite{parisot2017spectral} as the baseline GCN using s-\textit{f}MRI data from ABIDE.   

To preserve the the topology information in the population network and their associated individual brain function network, Jiang et al.~\cite{jiang2020hi} proposed a hierarchical GCN framework to map the brain network to a low-dimensional vector while preserving the topology information. Their method leveraged a correlation mechanism in populating the network which could capture more information and result in more accurate brain network representation, and thus better classification of ASD from the ABIDE dataset~\cite{di2014autism} in comparison to Eigenpooling GCN~\cite{ma2019graph} and the other population GCN~\cite{parisot2017spectral} methods.  

Finally, as stated earlier, uncertainties associated with ASD makes it challengings~\cite{mastrovito2018differences}, and thus Huang et al.~\cite{huang2020edge} proposed an Edge-Variational GCN (EV-GCN) model with a learnable adaptive population graph core to incorporate multi-modal data for uncertainty-aware disease detection. Their model was tested on ASD/HC data, collected at the Yale Child Study Center~\cite{li2019graph} and showed the efficacy of the proposed method for embedding ASD and HC brain graphs.

\subsubsection{Schizophrenia}
Automatic classification of schizophrenia (SZ) based on \textit{f}MRI data has also attracted attention. SZ is a devastating mental disease with extraordinary complexity characterized by behavioral symptoms such as hallucinations and disorganized speech. SZ shows local abnormalities in brain activity and in functional connectivity networks which can have unusual or disrupted topological properties.
Rakhimberdina and Murata~\cite{rakhimberdina2019linear} exploited the simple linear graph~\cite{wu2019simplifying} model for SZ detection, achieving an accuracy of 80.55\% for a binary classification task. The use of the linear model within the graph model has a clear impact on decreasing its computational time. However, the edge construction strategy can be further improved by incorporating techniques to learn the edge weights such as self-attention weight features.

\subsubsection{Attention deficit hyperactivity disorder}
Some studies have shown that \textit{f}MRI-based analysis is also effective in helping understand the pathology of brain diseases such as attention-deficit hyperactivity disorder (ADHD). ADHD is a condition that affects people's behaviour and learning, making it difficult for them to concentrate, and impulsive and overactive.
The model proposed by Rakhimberdina and Murata~\cite{rakhimberdina2019linear} based on a population graph was also used to separate adults with ADHD from healthy controls. The graph constructed using gender, handedness and acquisition site features reached an accuracy of 74.35\%. 
Yao et al.~\cite{yao2019triplet} also implemented the multi-scale tripled GCN previously introduced to identify ADHD using the ADHD-200 dataset~\cite{bullmore2012economy}. 
To generate functional connectivity networks under different spatial scales and ROI definitions, the authors first apply a multi-scale templates to each subject. From a specific template, a graph is generated where the ROI represents each node and the connections between a pair of ROIs is defined by the Pearson correlation of their mean time series.

\subsubsection{Major depressive disorder}
Major depressive disorder (MDD) is a mental disease characterised by a depressed mood, diminished interests and impaired cognitive function. Among various neuroimaging techniques, rs-\textit{f}MRI can observe dysfunction in brain connectivity on BOLD signals, and has been used to discriminate between MDD patients and healthy controls.
Yao et al.~\cite{yao2020temporal} exploited time-varying dynamic information with a temporal adaptive GCN on rs-\textit{f}MRI data to learn the periodic brain status changes to detect MDD. The model learns a data-based graph topology and captures dynamic variations of the brain \textit{f}MRI data, and outperforms traditional GCNs~\cite{kipf2017semi} and GATs~\cite{velivckovic2017graph} models.

\subsubsection{Bipolar disorder}
Bipolar disorder (BD), or manic depression, is a mental health condition that causes extreme mood swings. Functional and structural brain studies have identified quantitative differences between BD and healthy controls; thus, combining modalities may uncover hidden relationships.
Yang et al.~\cite{yang2019interpretable} proposed a graph-attention based method that integrates structural MRI and \textit{f}MRI to detect bipolar disorder. 
The main challenges in multimodal data fusion are the dissimilarity of the data types being fused and the interpretation of the results. 
One of the advantages of attention mechanisms is that they allow for the use of variable-sized inputs when focusing on the most important parts of the data to make decisions, which can then be used to interpret the salient input features. The model showed superiority over other machine learning classifiers and alternative GCN formulations.

\subsubsection{Gender classification with brain connectivity}
Locating brain areas with a critical role in human behaviour and mapping functions to brain regions are among the most important goals in the field of neuroscience. To explore the task of brain ROI identification, multiple authors have performed gender classification on functional connectivity networks, based on previous evidence for gender-related differences in brain connectivity~\cite{satterthwaite2015linked}. Although, gender classification is not directly related with detecting or classifying a disease, the outcome of these studies can be used to identify brain regions that are related to a certain disease.

Graph convolutional networks have been applied to brain connectivity data to distinguish between male and female subjects.
Arslan et al.~\cite{arslan2018graph} explored GCNs for the task of brain ROI identification in the gender classification of more than 5000 participants from the UK Biobank dataset~\cite{bycroft2018uk}. The prediction is based on their functional connectivity networks captured at rest. The activations of the feature maps are used for visual attribution of the nodes, each of which is associated with a brain region. However the applicability of the method may be limited by the definition of the number of nodes and signal choice.
The graph similarity metric proposed by Ktena et al.~\cite{ktena2018metric}, was also adopted in the UK Biobank dataset. Individuals of the same sex are represented with matching pair graphs, \textit{i.e.} non-matching pairs include one male and one female subject.

\begin{figure}[!t]
\centering
\includegraphics[width=1\linewidth]{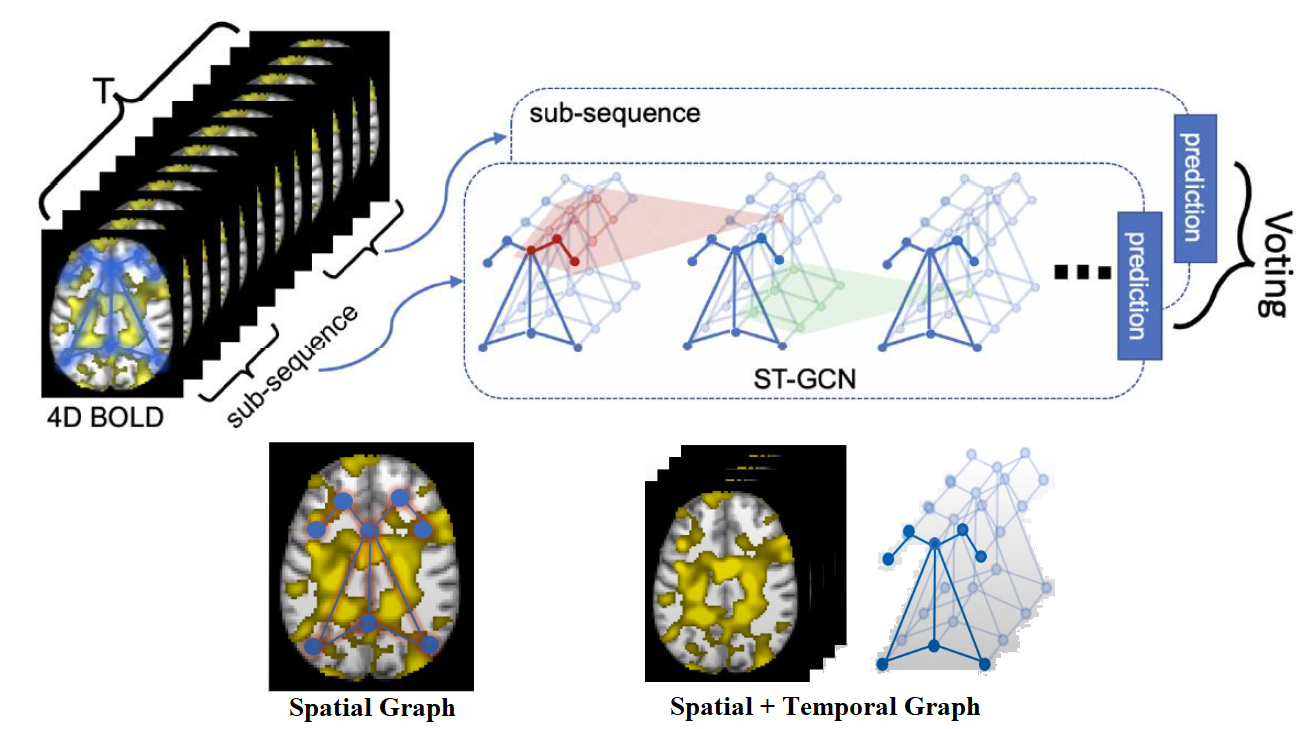}
\vspace{-10pt}
\caption{
Framework for classifying BOLD time series with spatio-temporal GCN. 
A \textit{spatial graph} is constructed by considering each ROI to be a node connected with edges, where the weights of these edges are the functional affinity between ROIs defined by the magnitude of correlation between the concatenated time series. 
In the \textit{Temporal graph} the spatial graph is copied across time and the nodes of the same ROIs are connected across time points $T$. Image adapted from~\cite{gadgil2020spatio2}.
}
\label{fig:Fig6}
\vspace{-6pt}
\end{figure}

Existing deep learning methods used to rs-\textit{f}MRI data either eliminate the information of the temporal dynamics of brain activity or overlook the functional dependency between different brain regions in a network~\cite{arslan2018graph}. 
To address this limitation, Gadgil et al.~\cite{gadgil2020spatio2} proposed a spatio-temporal GCN on blood-oxygen-level-dependent (BOLD) time series data to model the non-stationary nature of functional connectivity. The model is used to predict the age and gender of healthy individuals on the Human Connectome Project (HPC) dataset. The achieved accuracy of 83.7\% outperforms traditional RNN-based methods where the learned edge importance localizes meaningful brain regions and functional connections associated with gender differences. Fig.~\ref{fig:Fig6} illustrates the spatio-temporal GCN framework. 
In the model proposed by Azevedo et al.~\cite{azevedo2020towards}, embeddings are created for each node through 1D convolutional operations, where each node corresponds to a single timeseries sampled in one brain region. Temporal convolutional networks (TCNs) are used on top of normal convolutional networks to capture temporal features. This is followed by a GCN layer to transform each node's features according to information passed from its neighbors and a linear transformation is adopted to generate the final prediction.
Azevedo et al.~\cite{azevedo2020deep} also used a single end-to-end architecture that included temporal convolutions and graph neural networks to leverage both the spatial and temporal information in rs-\textit{f}MRI data and apply this to the HCP dataset~\cite{van2013wu}.
TCNs capture the intra-temporal dynamics of BOLD time series while the GNNs extract the spatial inter-relationships between brain regions, \textit{i.e} intra- and inter-feature learning.
The same authors expanded this work~\cite{azevedo2020deep2} on a larger dataset, UK biobank~\cite{bycroft2018uk}, and included edge features (weights) when leveraging the graph structure in the network. To demonstrate the flexibility to extract human readable knowledge from the model, the authors analysed the clusters created by the graph using the association matrix learnt from the time series. For example, the brain regions were grouped in a manner that mirrors to a certain degree the well-known cytoarchitectural and functional properties of the cerebral cortex.

Filip et al.~\cite{filip2020novel} adapted a GAT architecture and employed an inductive learning strategy and the idea of a master node to create a graph classification architecture for gender on the HPC dataset~\cite{van2013wu}.

To visualise the important brain regions that are related to a certain phenotypic difference, Kim et al.~\cite{kim2020understanding} adopted a graph isomorphism network~\cite{xu2018powerful}, which is a generalized CNN in the graph space. Thus, traditional saliency map visualization techniques for CNNs such as Grad-CAM can be used to visualize important brain regions. This provides more accurate and better interpretability of the sex classification task.

Current brain network methods either ignore the intrinsic graph topology or are designed for a single modality. To address these challenges, Zhang et al.~\cite{zhang2020deeprep} proposed a graph representation to fuse functional (\textit{f}MRI) and structural brain networks (MRI). The cross-modality relationships and encoding is generated by an encoder-decoder process. 
Brain functional networks are more dynamic and fluctuate on the edge connections than brain structural networks. Brain areas that are strongly connected in the brain structural network, for example, are not necessarily strongly connected in the brain functional network.
The authors adopted the idea of the GAT model for a dynamic adjustment of the weights. Here, three aggregation mechanisms are dynamically combined (graph attention weight, the original edge weight, and the binary weight) through a multi-stage graph convolutional kernel.

\subsubsection{Brain responses to stimulus}
Identifying the relationship between brain regions in relation to specific cognitive stimuli has been an important area of neuroimaging research. An emerging approach is to study this brain dynamic using \textit{f}MRI data. To identify these brain states, traditional methods rely on acquisition of brain activity over time to accurately decode a brain state. 

\begin{figure}[!t]
\centering
\includegraphics[width=1\linewidth]{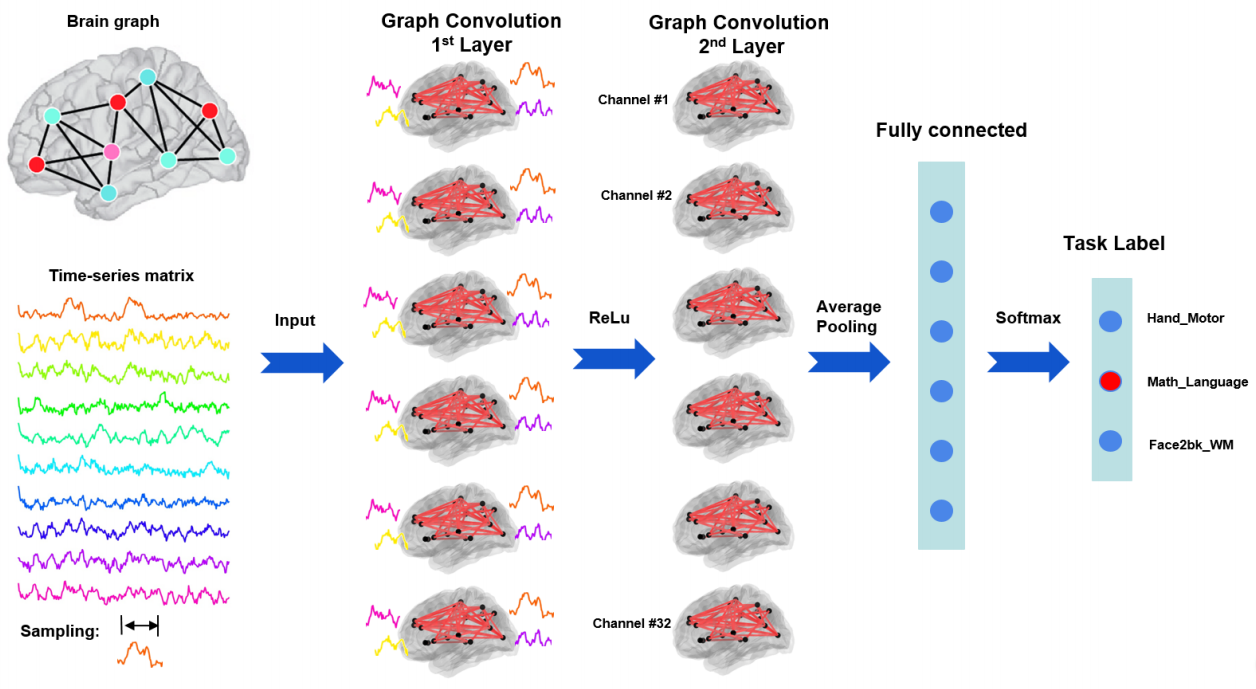}
\vspace{-10pt}
\caption{
Pipeline of functional brain decoding using graph convolutions. Image adapted from~\cite{zhang2019functional}.
}
\label{fig:Fig4}
\vspace{-6pt}
\end{figure}

Zhang et al.~\cite{zhang2019functional} proposed a GCN for classifying human brain activity on 21 cognitive tasks by associating a given window of \textit{f}MRI data with the task used. The GCN takes a short series of \textit{f}MRIs as input (10 seconds), propagates information among inter-connected brain regions, generates a high-level domain-specific graph representation, and predicts the cognitive state as depicted in Fig.~\ref{fig:Fig4}. This model outperforms a multi-class support vector machine classifier in identifying a variety of cognitive states in the HCP dataset~\cite{van2013wu}. However, the model only incorporates spatial graph convolutions, thus potentially losing the fine temporal information present in the BOLD signal~\cite{zhang2019functional}.

Identifying the particular brain regions that relate to a specific neurological disorder or cognitive stimuli is also critical for neuroimaging research. GNNs have been widely applied as a graph analysis method. Nodes in the same brain graph have distinct locations and unique identities. Thus, applying the same kernel over all nodes is problematic.
Li et al.~\cite{li2020braingnn} adopted weighted graphs from \textit{f}MRI and ROI-aware graph convolutional layers to infer which ROIs are important for prediction of cognitive tasks. The model maps regional and cross-regional functional activation patterns for classification of cognitive task decoding in the HCP 900 dataset~\cite{van2013wu}.
The framework is also capable of learning the node grouping and extracts graph features jointly, providing the flexibility to choose between individual-level and group-level explanations.

Deep learning has also been considered a competitive approach for analysing high-dimensional spatio-temporal data such as MEG signals. These signals are captured with 306 sensors (electrodes) distributed across the scalp that record the cortical activation. For reliable analysis it is critical to learn discriminative low-dimensional intrinsic features. 
Guo et al.~\cite{guo2017deep} proposed a spectral GCN model that integrates brain connectivity information to predict visual tasks using MEG data. The authors introduced an autoencoder-based network that integrates graph information to extract meaningful representations in an unsupervised manner, and classify whether a subject visualises a face or an object. This work focused on learning a low-dimensional representation from the input of MEG signals (\textit{i.e.} a dimensionality reduction technique).

\subsubsection{Image super resolution of functional brain connectome}
Ultra-high field MRI captures fine-grained variations in brain function and structure. However, MRI data at sub-millimeter resolutions is very scarce due to the high cost of the ultra-high field scanners.
Some works have proposed CNNs and GANs for image super-resolution to transform a lower resolution brain intensity image to an image of higher resolution. However, super-resolving brain connectomes (brain graphs) has seen limited attention. To generate brain connectomes at different resolutions, image brain atlases (templates) are used to define the parcellation of the brain into different anatomical regions of interest. However, the pre-processing phase of registration and label propagation are prone to variability and bias. 
Thus, given a low-resolution connectome a high-resolution connectome can be generated to prevent the need for manual labelling of anatomical brain regions and costly data collection.
Isallari et al.~\cite{isallari2020gsr} proposed a graph super-resolution network operating on graph-structured data that creates high-resolution brain graphs from low-resolution input graphs. This model introduces a Graph U-Autoencoder (encoder-decoder architecture based on CNNs) block and a super resolution block to generate a high-resolution connectome from the node feature embedding of the low-resolution connectome.

\begin{table*}[h!]
\caption{Summary of GCN approaches adopted to electrical-based analysis and their applications.}
\vspace{-5pt}
\centering
\label{table:electrical}
\resizebox{1\textwidth}{!}{%
\begin{tabular}{
>{\raggedright\arraybackslash}p{2.5cm} 
c l 
>{\raggedright\arraybackslash}p{4.7cm} 
>{\raggedright\arraybackslash}p{8cm}}
\toprule
\textbf{Authors} &
\textbf{Year} &
\textbf{Modality} & 
\textbf{Application} &  
\textbf{Dataset} \\
\midrule
Jang et al.~\cite{jang2019brain} & 2019 & EEG &
Classification: Affective mental states & DEAP~\cite{koelstra2011deap} (40 classes) \\
Jang et al.~\cite{jang2018eeg} & 2018 & EEG &
Classification: Affective mental states & DEAP~\cite{koelstra2011deap} (40 classes) \\
%
%
Yin et al.~\cite{yin2020eeg} $\star$ & 2020 & EEG &
Classification: Emotions & DEAP~\cite{koelstra2011deap} (2 classes) \\
Zhong et al.~\cite{zhong2020eeg} & 2020 & EEG &
Classification: Emotions & SEED~\cite{zheng2015investigating} (3 classes), SEED-IV~\cite{zheng2018emotionmeter} (4 classes) \\
Wang et al.~\cite{wang2020functional} & 2020 & EEG &
Classification: Emotions & DEAP~\cite{koelstra2011deap} (2 classes) \\
Liu et al.~\cite{liu2019sparse} $\star$ $\dagger$ & 2019 & EEG  &
Classification: Emotions & Southeast University (private) (3 classes), MPED~\cite{song2019mped} (7 classes) \\
Wang et al.~\cite{wang2019phase} & 2019 & EEG &
Classification: Emotions & SEED~\cite{zheng2015investigating} (3 classes), DEAP~\cite{koelstra2011deap} \\
Zhang et al.~\cite{zhang2019gcb} & 2019 & EEG &
Classification: Emotions & SEED~\cite{zheng2015investigating} (3 classes), DREAMER~\cite{katsigiannis2017dreamer} (9 classes) \\
Song et al.~\cite{song2018eeg} & 2018 & EEG &
Classification: Emotions & SEED~\cite{zheng2015investigating} (3 classes), DREAMER~\cite{katsigiannis2017dreamer} (9 classes) \\
Wang et al.~\cite{wang2018eeg} & 2018 & EEG &
Classification: Emotions & SEED~\cite{zheng2015investigating} (3 classes) \\
%
%
Mathur et al.~\cite{mathur2020graph} & 2020 & EEG  &
Classification: Seizure detection & University of Bonn~\cite{andrzejak2001indications} (2 classes) \\
Wang et al.~\cite{wang2020sequential} $\star$ & 2020 & EEG &
Classification: Seizure detection & University of Bonn~\cite{andrzejak2001indications} (2 classes), SSW-EEG (private) (2 classes) \\
Covert et al.~\cite{covert2019temporal} $\star$  & 2019 & EEG  &
Classification: Seizure detection & Cleveland Clinic Foundation (private) (2 classes) \\
Lian et al.~\cite{lian2020learning} $\dagger$ & 2020 & iEEG  &
Regression: Seizure prediction (preictal) & Freiburg iEEE (EPILEPSIAE)~\cite{ihle2012epilepsiae} \\
%
%
Wagh et al.~\cite{wagh2020eeg} & 2020 & EEG  &
Classification: Abnormal EEG & TUH EEG corpus~\cite{obeid2016temple}, MPI LEMON~\cite{babayan2019mind} (2 classes)\\
%
%
Wang et al.~\cite{wang2020weighted} $\dagger$ & 2020 & ECG &
Classification: Heart abnormality & HFECGIC~\cite{TIANCHI} (34 classes) \\
Sun et al.~\cite{sun2020graph} & 2020 & EGM &
Classification: Heart abnormality & EGM open-heart surgery~\cite{yaksh2015novel} (2 classes) \\
%
%
Jia et al.~\cite{jia2020graphsleepnet} $\star$ $\dagger$ & 2020 & PSG &
Classification: Sleep staging & MASS-SS3~\cite{o2014montreal} (5 classes) \\
%
%
Lun et al.~\cite{lun2020gcns} & 2020 & EEG &
Classification: Brain motor imagery & EEG PhysioNet~\cite{schalk2004bci2000,goldberger2000physiobank} (4 classes) \\ 
Kwak et al.~\cite{kwak2020graph} & 2020 & EEG &
Classification: Brain motor imagery & EEG PhysioNet~\cite{schalk2004bci2000,goldberger2000physiobank} (4 classes) \\ 
Zhang et al.~\cite{zhang2018brain2object} & 2018 & EEG &
Classification: Brain motor imagery & EEG-L (private) (4 classes) \\ 
Li et al.~\cite{li2019classify} $\star$ & 2019 & EEG &
Classification: Brain motor imagery & EEG PhysioNet~\cite{schalk2004bci2000,goldberger2000physiobank} (2 classes) \\ 
Jia et al.~\cite{jia2020attention} $\dagger$ & 2020 & EEG &
Classification: Brain motor imagery & EEG PhysioNet~\cite{schalk2004bci2000,goldberger2000physiobank} (4 classes) \\ 
\bottomrule
\multicolumn{5}{p{350pt}}
{ 
$\star$ GCN with temporal structures for medical diagnostic analysis. \newline
$\dagger$ GCN with attention structures for medical diagnostic analysis.
}
\end{tabular}}
\vspace{-6pt}
\end{table*}

\vspace{-6pt}
\subsection{Electrical-based analysis}

\subsubsection{Affective mental states}
\begin{figure}[!t]
\centering
\includegraphics[width=1\linewidth]{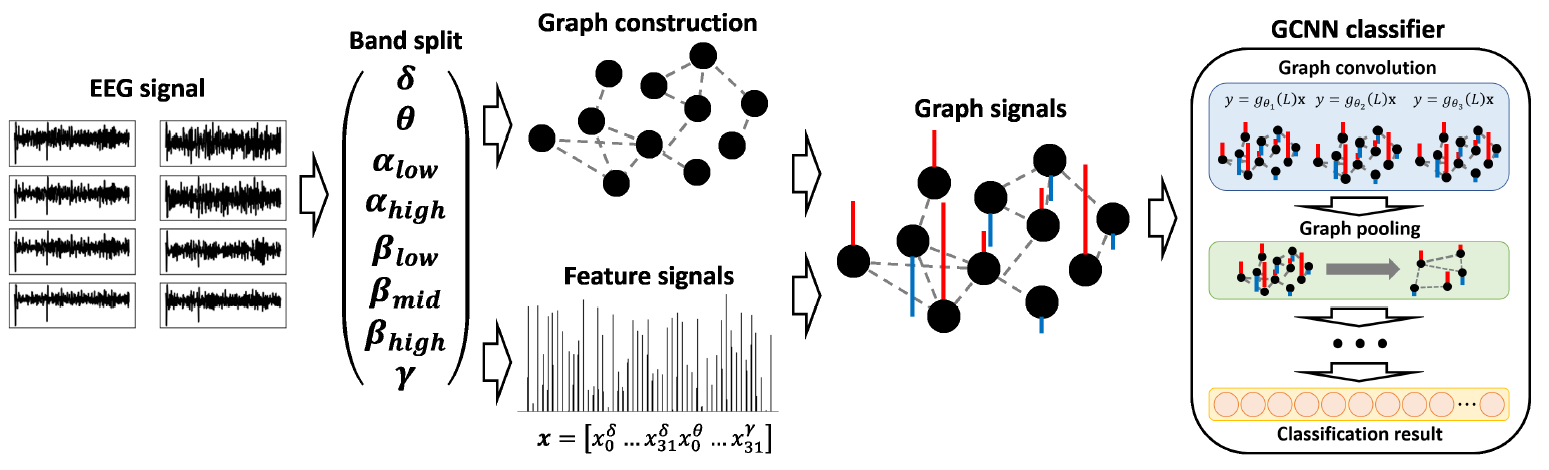}
\vspace{-15pt}
\caption{
Features are extracted from EEG signals to construct a graph-based architecture and classify mental states. Image adapted from~\cite{jang2018eeg}.
}
\label{fig:Fig17}
\vspace{-6pt}
\end{figure}

Brain signals provide comprehensive information regarding the mental state of a human subject. 
Jang et al.~\cite{jang2018eeg} proposed the first method to apply deep learning on graph signals to EEG-based visual stimulus identification. The model converts the EEG into graph signals with appropriate graph structures and signal features as input to GCNs to identify the visual stimulus watched by a human subject. Compared to \textit{f}MRI signals, EEG analysis is limited to observing a smaller number of brain regions (\textit{i.e.} electrodes) which may not allow for a sufficiently rich graph representation. Thus, the authors create a graph containing both intra-band and inter-band connectivity. This proposed approach is illustrated in Fig.~\ref{fig:Fig17}. 
Defining the graph connectivity structure for a given task is an ongoing problem and current models still have the limitation that appropriate graph structures need to be manually designed. 
To address this, Jang et al.~\cite{jang2019brain} proposed an EEG classification model that can determine an appropriate multi-layer graph structure and signal features from a collection of raw EEG signals and classify them. In contrast to approaches that use a pre-defined connectivity structure, this method for learning the graph structure enhances classification accuracy.

\subsubsection{Emotion recognition}
Human emotion is a complex mental process that is closely linked to the brain’s responses to internal or external events. The analysis of the outcomes of emotion recogniton can be used to potentially detect emotion changes which occur when exposed to mental stress or depression, which are common characteristics of post traumatic stress disorders. However, there are not current clinical applications related to graph-based emotion recognition.

\begin{figure}[!t]
\centering
\includegraphics[width=1\linewidth]{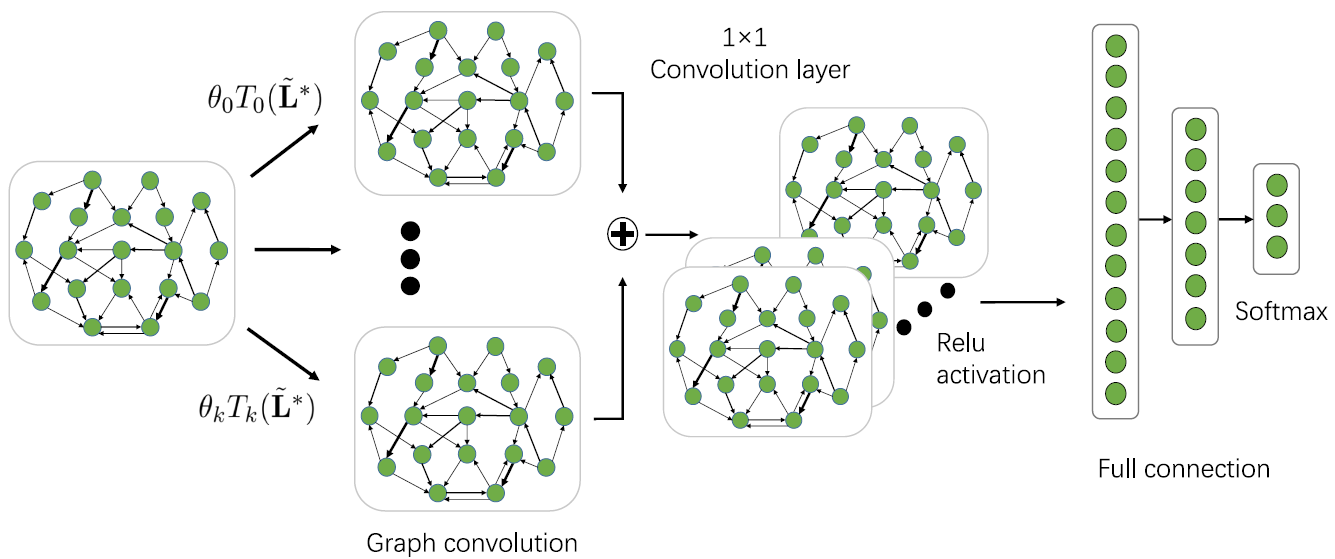}
\vspace{-15pt}
\caption{
A dynamic GCN model is proposed for EEG emotion recognition where the adjacency matrix that characterises the relationship between various vertices is learned. Image adapted from~\cite{song2018eeg}.
}
\label{fig:Fig16}
\vspace{-6pt}
\end{figure}

Song et al.~\cite{song2018eeg} proposed a dynamic GCN which could dynamically learn the intrinsic relationship between different EEG channels (represented by an adjacency matrix) through back propagation, as depicted in Fig.~\ref{fig:Fig16}. This method facilitates more discriminative feature extraction and the experiments conducted in the SJTU emotion EEG dataset (SEED)~\cite{zheng2015investigating} and the DREAMER dataset~\cite{katsigiannis2017dreamer} achieved recognition accuracy rates of 90.4\% and 86.23\% respectively.

While learning the adjacency matrix addresses the challenges of designing this by hand, the learned graph feature space of the EEG may not be the most representative feature space. 
Motivated by the random mapping ability of the broad learning system~\cite{chen2017broad}, Wang et al.~\cite{wang2018eeg} introduced a broad learning system that is combined with a dynamic GCN. This model can randomly generate a learned graph space that maps to a low-dimensional space, then expand it to a broad random space with enhancement nodes to search for suitable features for emotion classification.
GCNs do not benefit from depth the way DCNNs do, and accuracy decreases as the depth of graph convolutional layers increases beyond a few layers.
Hence, Zhang et al.~\cite{zhang2019gcb} proposed a graph convolutional broad network which uses regular convolution to capture higher-level (\textit{i.e.} deeper, more abstract) information. This model stacks a regular CNN after graph convolution to obtain high-level features from the learned graph representation, and preserve more information for searching features in broad spaces through layer concatenation. Broad learning systems can handle and search for more powerful features in both deep and broad spaces~\cite{chen2017broad}. Broad connections can also enhance the stability of models to ensure that performance of the whole network won’t be worse than a single hierarchy.

Because of the way brain regions cooperate and labour is divided between them, the spatial relationships and functional connections between EEG channels are not consistent over time.
Therefore, Wang et al.~\cite{wang2019phase} integrated a phase-locking value (PLV) technique~\cite{piqueira2011network} with GCN for emotion recognition, which determines emotional-related functional connections through the connectivity of the EEG signal. The spatial and functional intrinsic connections in the data are captured by modeling univariate EEG feature as a multivariate feature with the PLV brain network structure.
The same authors also described the functional connection relationship of the brain in a later work~\cite{wang2020functional}. After the brain network based on PLV is constructed, the model combines the functional integration and functional separation perspectives to detect differences in brain connectivity in the process of emotion generation. 

Yin et al.~\cite{yin2020eeg} proposed a fusion of GCNs and LSTMs for classifying emotions into positive (amusement, joy tenderness) and negative emotions (anger, sadness, fear, disgust). First, features such as the differential entropy are extracted from several segments. Then, a GCN layer is used to calculate the relationship between two EEG channels for a period of time, an LSTM layer is used to memorize changes between two EEG channels over a certain period, and a dense layer performs the final emotion recognition. Although the results were promising, the authors only explored the binary classification of emotions.

GCNs have been used to capture inter-channel relationships using an adjacency matrix. However, similar to CNNs and RNNs, GNN approaches only consider relationships between the nearest channels, meaning valuable information between distant channels may be lost.
A regularized graph neural network (RGNN) is applied by Zhong et al.~\cite{zhong2020eeg} for EEG-based emotion recognition, which captures inter-channel relations. Regularizations are techniques used to reduce the error or prevent overfitting by fitting a function appropriately.
Inter-channel relations are modeled via an adjacency matrix and a simple graph convolution network~\cite{wu2019simplifying} is used to learn both local and global spatial information. A node-wise domain adversarial training method and an emotion-aware distribution learning are adopted as regularizers for better generalization in subject-independent classification scenarios. A classification accuracy of 73.84\% is achieved on the SEED dataset~\cite{zheng2015investigating}; however, this method relies on hand-crafted features. The above studies indicate that the regional and asymmetric characteristics of EEGs are helpful to improve the performance of emotion recognition.

\begin{figure}[!t]
\centering
\includegraphics[width=1\linewidth]{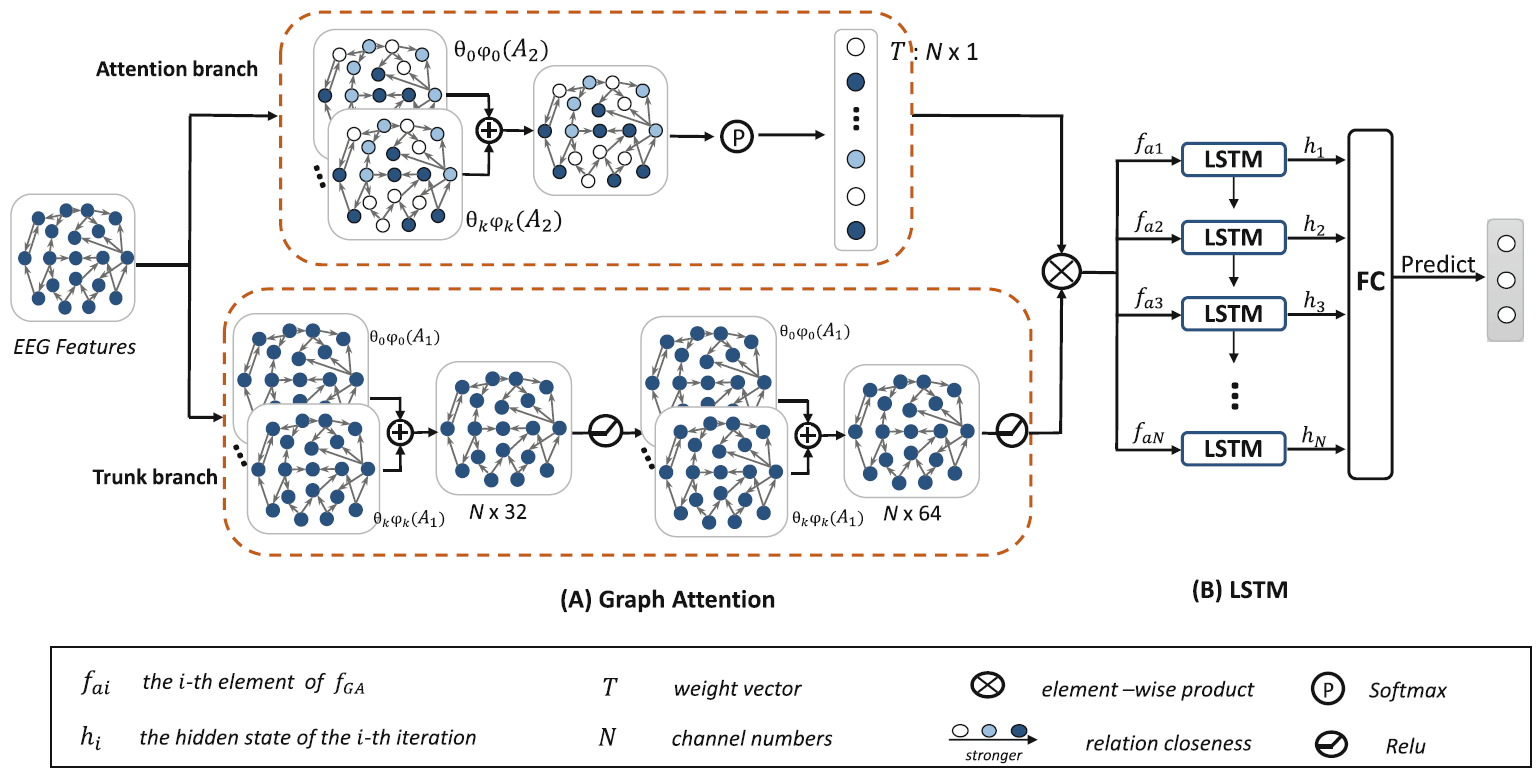}
\vspace{-15pt}
\caption{
A graph attention structure is used to extract discriminative features and an LSTM is adopted for modeling the spatial information in EEG channels. Image adapted from~\cite{liu2019sparse}.
}
\label{fig:Fig11}
\vspace{-8pt}
\end{figure}

To recognize emotions not all EEG channels are helpful. Although there have been algorithms used for channel selection, the relationships between EEG channels are rarely considered due to imperceptible neuromechanisms, which are critical for EEG emotion recognition.
Liu et al.~\cite{liu2019sparse} proposed a graph-based attention structure to select EEG channels for extracting more discriminative features. The framework, which consists of a graph attention structure and an LSTM, is illustrated in Fig.~\ref{fig:Fig11}. The higher recognition accuracies achieved are likely due to the use of an attention mechanism in building the network; however, they often require longer time to train than a simple graph convolutional network~\cite{wu2019simplifying}.

\subsubsection{Epilepsy}
Epilepsy is one of the most prevalent neurological disorders characterised by the disturbance of the brain electrical activity, and recurrent and unpredictable seizures. Machine learning applications have been used for seizure prediction, seizure detection and seizure classification through the analysis of EEG/iEEG signals. CNNs and RNNs have shown success in analysing these signals for Epilepsy related tasks, but they suffer from a loss of neighborhood information. On the other hand, GCNs represent the relationships between electrodes using edges, and can thus preserve rich connection information. 

\textit{Seizure detection} from time-series refers to recognising the ictal activity or that a seizure is occurring (\textit{i.e.} determine the presence or absence of ongoing seizures).
Mathur et al.~\cite{mathur2020graph} presented a method for detecting ictal activity using a visibility graph on the EEG by employing a Gaussian kernel function to assign edge weight. A graph discrete Fourier transform is also applied to obtain features which are used in the classification phase.
Some works have proven the relationship between epilepsy and EEG components on certain frequencies and this frequency-domain representation can generate highly interpretable results. 
Wang et al.~\cite{wang2020sequential} introduced a sequential GCN that preserves the sequential information in 1D signals. The model is based on a complex network that represents a 1D signal as a graph~\cite{zou2019complex}, in which each data point corresponds to a node and each edge is computed by a connection rule. The authors first transform the time-domain signal using a fast Fourier transform to produce a sequence of frequency-domain features that are aligned in the time domain, from which they develop a graph representation. Then, a GCN is adopted to learn features from the input network to improve the classification performance. By combining the frequency-domain network representation with the GCN the model can detect conventional seizures in the Bonn dataset~\cite{andrzejak2001indications}, and a seizure type known as absence epilepsy from a private dataset. However, multi-channel EEG signals were not considered in the experimental setup.
Covert et al.~\cite{covert2019temporal} proposed a temporal graph convolutional network (TGCN) which consists of feature extractors that are localized and shared over both time and space. TGCN is inherently invariant to when and where the patterns occur. The authors investigate the benefits of TGCN's interpretability in terms of assisting clinicians in determining when seizures occur and which areas of the brain are most involved. However, the model is limited to allow varying graph structures.

\textit{Seizure prediction} aims to predict upcoming seizures or the pre-ictal brain state (\textit{i.e.} before a seizure). The underlying relationship in the pre-ictal period can be diverse across patients, making it difficult to build a predefined graph that is effective for a large number of patients. 
To address this, instead of directly using a prior graph, Lian et al.~\cite{lian2020learning} proposed to build a graph based on the influences of relationships.
The authors introduced global-local GCNs that jointly learn the structure and connection weights to optimize the task-related learning of iEEG signals. The connections in nodes are updated with attention and gating mechanisms, but the model requires a large volume of data for training.

\subsubsection{Abnormal EEG in neurological disorders}
The application of machine learning techniques to automatically detecting anomalies in medical data is particularly attractive considering the difficulties in consistency and objectivity identifying anomalies. There exist numerous medical anomaly detection tasks, including identifying abnormal EEG recordings of patients with neurological disorders. An assessment is made when analysing an EEG recording to see whether the recorded signal appears to indicate abnormal or regular brain activity patterns.

Recent GCNs have addressed the challenges of learning the spatio-temporal relationships in EEG data. Wagh et al.~\cite{wagh2020eeg} introduced a GCN that captures both spatial and functional connectivity for multi-channel EEG data to distinguish between ``normal'' EEGs on patients with neurological diseases and the EEGs of healthy individuals. First, a graph-based representation with its corresponding node-level embedding is extracted from 10-second windows of EEG signals fed through a GCN model. Then, a graph-level embedding is computed using an averaging operation, the output of which is input to a fully connected network to obtain the output class. Finally, a maximum likelihood estimation based on the window-level prediction is adopted to determine if the entire EEG recording was recorded from a particular patient (\textit{i.e.} subject prediction). 
Results on two large-scale scalp EEG databases, TUH EEG corpus~\cite{obeid2016temple} and MPI LEMON~\cite{babayan2019mind}, significantly outperform traditional machine learning models. The authors also evaluated the effect of depth on GCNs, and find higher depth offers only a marginal improvement in performance. However, the data from patients and control participates was collected using different systems which may help to distinguish both classes and a feature engineering phase was considered which limits the model's ability to directly discover the optimal features from the data.

\subsubsection{Heart abnormalities}
Electrocardiograms (ECG) are widely used to identify cardiac abnormalities and a variety of methods have been proposed for the classification of ECG signals. However, an ECG record may contain multiple concurrent abnormalities and current deep learning methods may ignore the correlations between classes, and looks at each class independently. This can be addressed via graph-based representations. 

The GAT architecture has matched or surpassed state-of-the-art results across graph learning benchmarks. Still, it is designed to only classify nodes within a single network, and it can only deal with binary graphs. 
Wang et al.~\cite{wang2020weighted} proposed a multi-label weighted graph attention network to classify 34 kinds of electrocardiogram abnormalities. In this model ECG features are extracted from a CNN (1-D ResNet). The features of each class are fed into an improved GAT by integrating a co-occurrence weight with masked attentional weights. The weighted GAT helps capture the relationships within the ECG abnormalities. Then, the features learnt by the CNN and GAT are concatenated to output the probability of each class.

The epicardial electrogram (EGM) is measured on the heart's surface and has been used to analyse atrial fibrillation, a clinical arrhythmia correlated with stroke and sudden death. Conventional signal processing methods are less suitable for joint space time and frequency domain analysis. 
Sun et al.~\cite{sun2020graph} represented the spatial relationships of epicardial electrograms through a graph to formulate a high-level model for atrial activity. The authors evaluated the spatio-temporal variation of EGM data with a graph-time spectral analysis framework and identified spectral differences between normal heart rhythms and atrial fibrillation from EGM signals taken during open heart surgery~\cite{yaksh2015novel}.

\subsubsection{Sleep staging}
Sleep stage classification, the process of segmenting a sleep period into epochs, is essential for clinical assessment of sleep disorders including insomnia, circadian rhythm disorders and sleep-related breathing and movement disorders~\cite{panossian2009review}; which may lead to serious health problems affecting quality of life. Sleep staging analysis is conducted through the analysis of electro-graphic measurements of the brain, eye movement, chin muscles, cardiac and respiratory activity and is collected with a polysomnography (PSG). The manual determination of sleep stages on PSG records is a complex, costly, and problematic process that requires expertise. Although traditional CNN and RNN models can achieve high accuracy for automatic sleep stage classification, the models ignore the connections among brain regions and capturing the transition between sleep stages continues to be challenging. Sleep experts identify one sleep stage according to both EEG patterns and the class label of its neighbors.
To address these challenges, Jia et al.~\cite{jia2020graphsleepnet} adopted an adaptive graph connection representation with attention, ST-GCN~\cite{guo2019attention}, for automatic sleep stage classification and to capture sleep transition rules temporally. First, the pairwise relationship between nodes (EEG channels) is constructed dynamically; then, a ST-GCN model with attention is adopted to extract both spatial and temporal features. Experimental results in classifying 5 sleep stages on the PSG dataset MASS-SS3~\cite{o2014montreal} achieves the best performance compared to SVM, CNN and RNN baselines.

\subsubsection{Brain motor imagery (human motor intentions)}
Brain-computer-interfaces (BCIs) have been used to assist the rehabilitation of patients with brain injuries, stroke and Parkinson’s Disease (PD). In particular, EEG-based motor imagery techniques have been extensively employed to manipulate the peripherals via neural activities~\cite{mahmood2019fully}. These brain signals can interact with external devices such as wheelchairs and intelligent robots.
For example, Zhang et al.~\cite{zhang2018brain2object} combined CNNs and GNNs to discover the latent information from the EEG signal. The proposed system is able to learn an illustration of an object seen by an individual from visually-evoked EEG signals (Brain-2-Object recognition). Thus, there is interest in better understanding the mechanism of cognitive functions and build robust EEG-based BCI systems. 

Some researchers have adopted GCNs to detect human motor intents from raw EEG signals (graph-structured data). This aims to classify several motor imagery tasks including opening and closing the left fist, right fist, both fists and both feet from the EEG Motor Movement/Imagery dataset~\cite{schalk2004bci2000}.
%
Lun et al.~\cite{lun2020gcns} introduced a GCN to detect four-classes of motor imagery intentions correlated with the functional topological relationship of EEG electrodes. A graph Laplacian is built to represent the correlation between electrodes based on the absolute Pearson correlation coefficient. Then, during various types of motor imagery tasks, the GCN learns generalized features to improve the decoding efficiency of raw EEG signals. Results on the PhysioNet dataset~\cite{schalk2004bci2000,goldberger2000physiobank} show that it is necessary to optimize the GNN structure. 
Kwak et al.~\cite{kwak2020graph} improved this structure by introducing multilevel feature fusion to the GNN that alleviates the limitations of sequential convolutional and pooling layers, which results in each node losing their local information. In this model the feature representation is combined with the author's previous 3D-CNN~\cite{kwak2020multilevel} to improve the performance of brain motor imagery classification.

Li et al.~\cite{li2019classify} proposed an edge-aware ST-GCN for EEG classification, where the EEG is represented as frames of a graph. The authors selected the dataset for the task of imagining opening and closing the left or right fist~\cite{schalk2004bci2000,goldberger2000physiobank}. 
The proposed model learns both spatial and temporal patterns from data. The authors applied a learnable mask to automatically learn the graph structure, feeding each graph convolution with the inner product of the adjacency matrix of a complete graph and a learnable weight matrix. During the training process, the weight matrices learn the dynamic latent graph structure.
%
Jia et al.~\cite{jia2020attention} introduced a structure using a GCN and an attention based graph ResNet to achieve precise detection of human motor intentions. Their approach considered the topological relationships between EEG electrodes. This method outperformed traditional RNN-based and CNN-based approaches with attention structures, and was shown to handle inter-subject and inter-trial variations in raw EEG data~\cite{schalk2004bci2000,goldberger2000physiobank}.

\begin{table*}[t!]
\caption{Summary of GCN approaches adopted for anatomical structure analysis and their applications (Group 1).}
\vspace{-5pt}
\centering
\label{table:anatomical1}
\resizebox{1\textwidth}{!}{%
\begin{tabular}{
l c l 
>{\raggedright\arraybackslash}p{5cm} 
>{\raggedright\arraybackslash}p{7.6cm}}
\toprule
\textbf{Authors} &
\textbf{Year} &
\textbf{Modality} & 
\textbf{Application} &  
\textbf{Dataset} \\
\midrule
Ma et al.~\cite{ma2020attention} $\dagger$ & 2020 & MRI &
Classification: Alzheimer's disease & ADNI~\cite{jack2008alzheimer} (2 classes) \\
Huang et al.~\cite{liu2020identification} & 2020 & MRI / \textit{f}MRI &
Classification: Alzheimer's disease & ADNI~\cite{petersen2010alzheimer} (3 classes) \\ 
Huang et al.~\cite{huang2020edge} & 2020 & MRI &
Classification: Alzheimer's disease & ADNI~\cite{petersen2010alzheimer} (3 classes), TADPOLE~\cite{marinescu2018tadpole} (3 classes) \\ 
Yu et al.~\cite{yu2020multi} & 2020 & MRI &
Classification: Alzheimer's disease / MCI & ADNI~\cite{petersen2010alzheimer} (3 classes) \\ 
Gopinath et al.~\cite{gopinath2020learnable} & 2020 & MRI &
Classification: Alzheimer's disease  & ADNI~\cite{petersen2010alzheimer} (2 classes) \\ 
Zhao et al.~\cite{zhao2019graph} & 2019 & MRI &
Classification: Alzheimer's disease / MCI & ADNI~\cite{petersen2010alzheimer} (2 classes) \\ 
Wee et al.~\cite{wee2019cortical} & 2019 & MRI &
Classification: Alzheimer's disease & ADNI~\cite{petersen2010alzheimer} (2 classes), Asian cohort (private) (2 classes) \\
Kazi et al.~\cite{kazi2019inceptiongcn} & 2019 & MRI &
Classification: Alzheimer's disease  & TADPOLE~\cite{marinescu2018tadpole} (3 classes) \\ 
Song et al.~\cite{song2019graph}  & 2019 & MRI &
Classification: Alzheimer's disease  & ADNI~\cite{petersen2010alzheimer} (4 classes) \\ 
Gopinath et al.~\cite{gopinath2019adaptive} & 2019 & MRI &
Classification: Alzheimer's disease  & ADNI~\cite{petersen2010alzheimer} (2 classes) \\ 
Guo et al.~\cite{guo2019predicting} & 2019 & PET &
Classification: Alzheimer's disease  & ADNI$_{2}$~\cite{beckett2015alzheimer} (2/3 classes) \\ 
Parisot et al.~\cite{parisot2018disease} & 2018 & MRI &
Classification: Alzheimer's disease  & ADNI~\cite{petersen2010alzheimer} (3 classes) \\ 
Parisot et al.~\cite{parisot2017spectral} & 2017 & MRI &
Classification: Alzheimer's disease  & ADNI~\cite{petersen2010alzheimer} (3 classes) \\ 
Xing et al.~\cite{xing2019dynamic} $\star$ & 2019 & T1WI / \textit{f}MRI &
Classification: Alzheimer's disease / EMCI & ADNI~\cite{jack2008alzheimer} (2 classes) \\ 
%
%
Zhang et al.~\cite{zhang2018multi} & 2018 & sMRI / DTI  &
Classification: Parkinson's disease & PPMI~\cite{marek2011parkinson} (2 classes) \\
McDaniel and Quinn~\cite{mcdaniel2019developing} $\dagger$ & 2019 & sMRI / dMRI  &
Classification: Parkinson's disease & PPMI~\cite{marek2011parkinson} (2 classes) \\
Zhang et al.~\cite{zhang2020deeprep} $\dagger$ & 2020 & sMRI / dMRI  &
Classification: Parkinson's disease & PPMI~\cite{marek2011parkinson} (2 classes) \\
%
%
Yang et al.~\cite{yang2019classification} & 2019 & MRI & 
Classification: Brain abnormality & Brain MRI images (private) (2 classes) \\
%
%
Gopinath et al.~\cite{gopinath2020learnable} & 2020 & T1WI  &
Classification: Gender & Mindboggle-101~\cite{klein2017mindboggling} (2 classes) \\
%
%
Wang et al.~\cite{wang2020covid} & 2020 & CT &
Classification: COVID-19 detection & Chest CT scans (private) (2 classes) \\
Yu et al.~\cite{yu2020resgnet} & 2020 & CT &
Classification: COVID-19 detection & Hospital of Huai’an City (private) (2 classes) \\
%
%
Wang et al.~\cite{wang2021explainable} & 2021 & CT &
Classification: Tuberculosis & Chest CT scans (private) (2 classes) \\
%
%
Hou et al.~\cite{hou2021multi} $\dagger$ & 2021 &  X-Ray &
Classification: Chest phatologies & IU X-ray~\cite{demner2016preparing} (14 classes), MIMIC-CXR~\cite{johnson2019mimic} (14 classes)  \\ 
Zhang et al.~\cite{zhang2020radiology} $\dagger$ & 2020 &  X-Ray &
Classification: Chest phatologies &  IU-RR~\cite{demner2016preparing} (20 classes)  \\ 
Chen et al.~\cite{chen2020label}  & 2020 &  X-Ray &
Classification: Chest phatologies & ChestX-ray14~\cite{wang2017chestx} (14 classes), CheXpert~\cite{irvin2019chexpert} (14 classes) \\ 
Zhang et al.~\cite{zhang2021improved} & 2021 & X-Ray &
Classification: Breast Cancer & mini-MIAS (mammogram)~\cite{suckling1994mammographic} (6 classes) \\
Du et al.~\cite{du2019zoom} & 2019 & X-Ray & 
Classification: Breast cancer & INbreast (full field digital mammogram)~\cite{moreira2012inbreast} (2 classes)  \\
%
%
Yin et al.~\cite{yin2019multi}  & 2019 & US &
Classification: Kidney disease & Children’s Hospital of Philadelphia (private) (2 classes) \\
%
%
Liu et al.~\cite{liu2020deep} & 2020 & MRI &
Regression: Relative brain age & Preterm MRI (private) \\ 
Gopinath et al.~\cite{gopinath2020learnable} & 2020 & MRI &
Regression: Relative brain age & ADNI~\cite{petersen2010alzheimer} \\ 
Gopinath et al.~\cite{gopinath2019adaptive} & 2019 & MRI &
Regression: Relative brain age & ADNI~\cite{petersen2010alzheimer} \\ 
%
%
Chen et al.~\cite{chen2020estimating} & 2020 & DMRI &
Regression: Brain data & BCP~\cite{howell2019unc} \\
Kim et al.~\cite{kim2019graph} & 2019 & DMRI &
Regression: Brain data & DMRI neonate (private) \\
Hong et al.~\cite{hong2019longitudinal} & 2019 & DMRI &
Regression: Brain data & DMRI infant (private) \\
Hong et al.~\cite{hong2019multifold} & 2019 & DMRI &
Regression: Brain data & HCP~\cite{sotiropoulos2013advances} \\
Hong et al.~\cite{hong2019reconstructing} & 2019 & DMRI &
Regression: Brain data & HCP~\cite{sotiropoulos2013advances} \\
%
%
%
%
Cheng et al.~\cite{cheng2020acceleration} & 2020 & MRF &
Regression: High-resolution 3D MRF & 3D MRF (private) \\
Hu et al.~\cite{hu2020feedback} $\dagger$ & 2020 &  MRI &
Regression: Medical image enhancement & MUSHAC~\cite{tax2019cross}, FLAIR~\cite{kuijf2019standardized}  \\
\bottomrule
\multicolumn{5}{p{350pt}}
{ 
$\star$ GCN with temporal structures for medical diagnostic analysis. \newline
$\dagger$ GCN with attention structures for medical diagnostic analysis.
}
\end{tabular}}
\end{table*}

\vspace{-6pt}
\subsection{Anatomical structure analysis (classification and prediction)}

\subsubsection{Alzheimer disease}

Alzheimer’s disease (AD) is an irreversible brain disorder which destroys memory and cognitive ability. There is as yet no cure for AD and monitoring its progress [Cognitively Normal (CN), Significant Memory Concern (SMC), Mild Cognitive Impairment (MCI) (including early MCI (EMCI) and late MCI (LMCI)) and AD] is essential to adjust the therapy plan for each stage. 

Similar to Autism Spectum Disorder (ASD), GCNs can be used to classify subjects into healthy or AD. Parisot et al.~\cite{parisot2018disease} constructed a population graph by integrating subject-specific imaging (MRI) and pairwise interactions using non imaging (phenotypic) data, then fed the sparse graph to a GCN to perform a semi-supervised node classification. Their experiments on the ADNI dataset for AD classification (conversion from (MCI) to AD) showed a high performance in comparison to a non-graph method~\cite{tong2016novel}. In addition, comparing to their prior work~\cite{parisot2017spectral} they showed a better graph structure (combining APOE4 gene data and eliminating AGE information) that could increase the accuracy of binary classification of AD on the ADNI dataset. 

Huang et al.~\cite{huang2020edge}  applied their edge-variational GCN (EV-GCN) method to the ADNI dataset for AD classification (the data was prepared in the same manner as Parisot~\cite{parisot2017spectral}). In addition, they applied their method on TADPOLE~\cite{marinescu2018tadpole} which is a subset of ADNI for classifying subjects into cognitive normal, MCI, and AD.  For TADPOLE, the authors constructed a graph by using the segmentation  features inferred  from  MRI  and  PET  data,  phenotypic data, APOE and FDG-PET biomarkers. Their results on both datasets showed a high performance in comparison to Parisot~\cite{parisot2017spectral} and Inception GCN~\cite{kazi2019inceptiongcn}. 

Zhao et al.~\cite{zhao2019graph} developed a GCN based method to predict MCI (EMCI vc NC, LMCI vs NC and LMCI vs EMCI) from rs-\textit{f}MRI. They constructed the MCI-graph using both imaging data extracted from rs\textit{f}MRI and non-imaging data including gender and collection device information. They classified the nodes in the generated MCI-graph using GCN and Cheby-GCN and compared the results with a Ridge, a random forest classifier and a multilayer perceptron, and demonstrated a high performance for Cheby-GCN over those methods. 

Xing et al.~\cite{xing2019dynamic} proposed a model consisting of dynamic spectral graph convolution networks (DS-GCNs) to predict early mild cognitive impairment (EMCI), and two assistive networks for gender and age to provide guidance for the final EMCI prediction. They constructed graphs using T1-weighted and \textit{f}MRI images from the ADNI~\cite{jack2008alzheimer} dataset. Apart from predicting age and gender for EMCI prediction, their model used an LSTM which could extract temporal information related to the EMCI prediction. 

Yu et al.~\cite{yu2020multi} used a multi-scale enhanced GCN (MSE-GCN) and applied it to a population graph which was built by combining imaging data(rs-\textit{f}MRI and diffusion tensor imaging (DTI)) and demographic relationships (\textit{e.g.} gender and age) to predict EMCI. This resulted in better performance in comparison to the prior methods of Zhao et al.~\cite{zhao2019graph} and Xing et al.~\cite{xing2019dynamic}. 
Huang et al.~\cite{liu2020identification} processed multi-modal data, MRI and rs-\textit{f}MRI, to identify EMCI. First, feature representation and multi-task feature selection are applied to each input. Then, a graph was developed using imaging and non-imaging (phenotypic measures of each subject) data. Finally, a GCN was used to perform the EMCI identification task from the ADNI dataset~\cite{petersen2010alzheimer}.

Song et al.~\cite{song2019graph} built a structural connectivity graph from DTI data from the ADNI imaging dataset and implemented a multi-class GCN classifier for the four class classification of subjects on the AD spectrum. The receiver operating characteristic (ROC) curve was compared between GCN and SVM classifiers for each class and demonstrated the capability of GCN over SVM (which relies on a predefined set of input features) for AD classification.

For the subject-specific aggregation of cortical features (MRI images), Gopinath et al.~\cite{gopinath2019adaptive,gopinath2020learnable} proposed an end-to-end learnable pooling strategy.
This method is a two-stream network, one for calculating latent features for each node of the graph, and another for predicting node clusters for each input graph. The learnable pooling approach can handle graphs with a varying number of nodes and connectivity. The results of their binary classification on the ADNI dataset~\cite{jack2008alzheimer} for NC vs AD, MCI vs AD and NC vs MC, showed the value of leveraging geometrical information in the GCN. 

Guo et al.~\cite{guo2019predicting} constructed a graph from the ROI of each subject's PET images from the ADNI$_{2}$ dataset~\cite{beckett2015alzheimer}, and proposed a PETNet model based on GCNs for EMCI, LMCI or NC prediction. The proposed method is computationally inexpensive and more flexible in comparison to voxel-level modeling. 

Ma et al.~\cite{ma2020attention} proposed an Attention-Guided Deep Graph Neural (AGDGN) network model to derive both structural and temporal graph features from the ADNI dataset~\cite{jack2008alzheimer}. This dataset contains four classes, however due to a shortage of data to train this model, they combined CN and SMC to form the CN group, and MCI and AD to form the AD group. This resulted in a two-class classification problem. They used an attention-guided random walk (AGRW) process to extract noise-robust graph embeddings. Their results indicated that the identified AD characteristics detected by the proposed model aligned with those reported by clinical studies.

To reduce the burden of creating a reliable population-specific classifier from scratch, generalization of classifiers to other datasets or populations, especially those with a limited sample size, is critical.
Wee et al.~\cite{wee2019cortical} employed a spectral graph CNN that incorporates the cortical thickness and geometry from MRI scans to identify AD. To demonstrate the generalisation and the feasibility to transfer classifiers learned from one population to another, the authors trained on a sizable caucasian dataset from the ADNI cohort~\cite{petersen2010alzheimer}, and evaluate how well the classifier can predict the diagnosis of an Asian population. 
To transfer the spectral graph-CNN model, the model that worked best on the ADNI cohort's testing set was fine-tuned on the training set of the Asian population. The performance of the fine-tuned model was then assessed using the testing set of the Asian cohort.

\subsubsection{Parkinson's disease}
Parkinson’s Disease (PD) is a neurological disorder characterized by motor and non-motor impairments. Motor deficits include bradykinesia, rigidity, postural instability, tremor, and dysarthria; and non-motor deficits include depression, anxiety, sleep disorders, and slowing of thought. 
Neuroimaging research using structural, functional and molecular modalities have also shed light on the underlying mechanism of Parkinson's disease. Many imaging based biomarkers have been demonstrated to be closely related to the progression of PD. 
Zhang et al.~\cite{zhang2018multi} developed a framework for analyzing neuroimages using GCNs to learn similarity metrics between subjects with PD and HC using data from the PPMI dataset~\cite{marek2011parkinson}. Structural brain MRIs are divided into a set of ROIs where each region is treated as a node on an undirected and weighted brain geometry graph. The authors showed the effectiveness of GCNs to learn features from similar regions and proposed a multi-view structure to fuse different MRI acquisitions. However, in this approach temporal dependency is not considered.

McDaniel and Quinn~\cite{mcdaniel2019developing} addressed the issue of analyzing multi-modal MRI data together by implementing a GAT layer to perform whole-graph classification. 
Instead of making predictions based on pairwise examples, GCNs predict the class of neuroimage data directly.

The features on each vertex must be pooled to generate a single feature vector for each input in order to convert the task from classifying each node to classifying the entire graph. The self-attention mechanism in GAT is used to compute the importance of graph vertices in a neighborhood, allowing for a weighted sum of the vertices' features during pooling. The results of combining diffusion and anatomical data from the PPMI dataset~\cite{marek2011parkinson} with the proposed model outperforms baseline algorithms on the features constructed from the diffusion data alone. 
The GAT attention layer also enables the possibility to interpret the magnitude of each node’s attention weight as the relative importance of a brain area for discriminating PD participants.

Zhang et al.~\cite{zhang2020deeprep} also adopted the cross-modality network embedding through the encoder-decoder network introduced above for gender classification with brain activity for PD detection. This model achieves the best prediction performance compared to CNN-based and graph-based approaches.
The model is capable of localizing 10 key regions associated with PD classification via a saliency map (\textit{e.g.} the bilateral hippocampus and basal ganglia which are structures conventionally conceived as PD biomarkers).

\subsubsection{Brain abnormality}
The ability to correctly recognize anomalous data is a deciding and crucial factor, so a highly accurate abnormality detection model is needed.
Yang et al.~\cite{yang2019classification} proposed a synergic graph-based model for a normal/abnormal classification of brain MRI images. The synergic deep learning method~\cite{zhang2017classification} can address the challenges faced by a GCN in distinguishing intra-classs variation and inter-class similarity. To improve the efficiency, the authors first extract the ROI of the image and use segmentation models as input to the model. The network consists of a dual GCN component (a pair of GCN models of identical construction) and a synergic training component. The synergic training component is used to predict whether a pair of images in the input layer belong to the same class and gives feedback if there is a synergic error.

\subsubsection{Gender based on brain structure}
As mentioned above in the Subsection of AD analysis, Gopinath et al.~\cite{gopinath2020learnable} also used the strategy that enables pooling operations on arbitrary graph structures for subject gender classification with T1-weighted MRI data~\cite{klein2017mindboggling}. 
Diversity in terms of brain regions is represented by the activation maps and clusters in the network. Several of these learned clusters highlight brain regions such as the hippocampus and amygdala, which are associated with gender-related differences in the literature. Further, the application of this method can be used to support the analysis of brain regions for disease diagnosis.

\subsubsection{Coronavirus 2 (SARS-CoV-2 or COVID-19)}
Early diagnosis of coronavirus is significant for both infected patients and doctors providing treatments. Viral nucleic acid tests and CT screening are the most widely used techniques to detect pneumonia which is caused by the virus, and thus to make a diagnosis.
Although CNNs have demonstrated a powerful capability to extract and combine spatial features from CT images, they are hindered because the underlying relationships between each element are ignored. Thus, GCNs are receiving attention in the analysis of COVID-19 patient CT images.
Yu et al.~\cite{yu2020resgnet} develop a graph framework that combined a graph representation with a CNN suitable for COVID-19 detection. A CNN model is used for feature extraction and graphs of the extracted features are constructed. Each feature is taken as one node of the graph while the edges between nodes are built according to the top \textit{k} neighbors with the highest similarity. The distance between nodes is measured by the Euclidean distance, while edges are quantified by the adjacency matrix. Classification performance into healthy and infected classes shows promising results, but the search domain of the size of batch and the number of neighbors needs further exploration.

Wang et al.~\cite{wang2020covid} also proposed an improved CNN that is combined with a GCN for higher classification accuracy. CNNs yield an individual image-level representation and the GCN focuses on a relation-aware representation. These representations are fused at the feature-level for COVID-19 detection from CT images. Although the model outperforms traditional CNN architectures, the method is limited in handling other modalities such as chest X-rays which are widely used to assist COVID-19 detection due to its availability, quick response, and cost-effective nature.

\subsubsection{Tuberculosis}
Tuberculosis(TB) is an infectious decease that can affect different organs such as abdomen, nervous system, but normally infects the lungs and known as pulmonary TB (PTB). Two main categories of PTB are primary pulmonary tuberculosis (PPT) and secondary pulmonary tuberculosis (SPT).  
Wang et al.~\cite{wang2021explainable} investigated GCN model to recognize the SPT as many PTB cases are turned to be SPT type. They proposed a rank-based pooling neural network (RAPNN) by which  individual image-level features can be extracted, then integrated the GCN to RAPNN and build a new model called GRAPNN to identify the SPT. The explanability of the proposed model analyzed using Grad-ACM, and their results outperformed SOTA including CNN models. 

\subsubsection{Chest pathologies}
Chest X-Ray imaging has been used to assist clinical diagnosis and treatment of several thoracic diseases where an individual image might be associated with multiple abnormalities, necessitating a multi-label image classification task. Several approaches have transformed a multi-label classification problem into multiple disjoint binary classification problems without acknowledging any label correlations. Abnormalities may be closely linked and label co-occurrence and interdependencies between these abnormal patterns (\textit{i.e.}, strong correlations among pathologies) are important for diagnosis.

\begin{figure}[!t]
\centering
\includegraphics[width=1\linewidth]{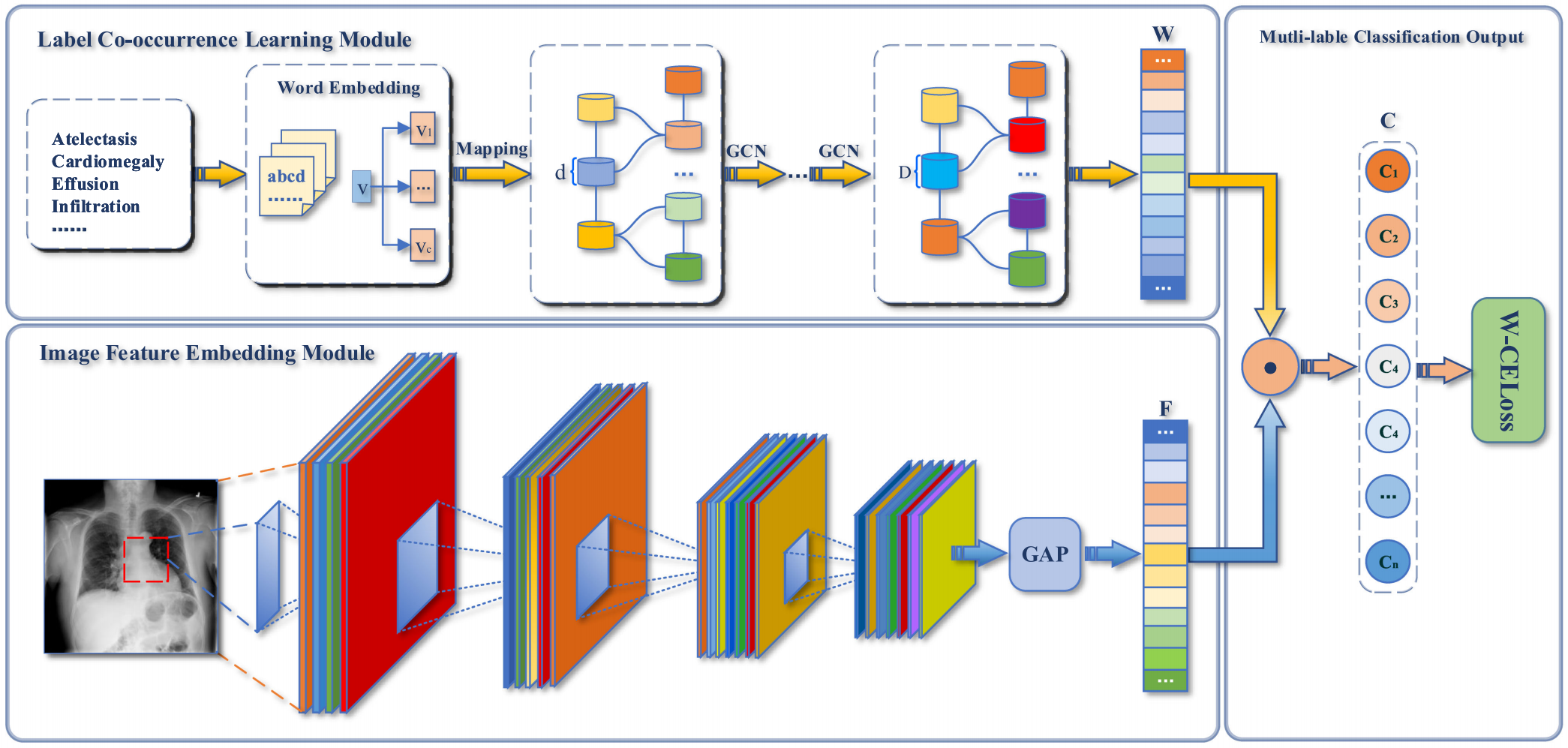}
\vspace{-10pt}
\caption{
GCN-based label co-occurrence learning framework to explore potential abnormalities with the guidance of semantic information, including the pathology co-occurrence and interdependency. Image adapted from~\cite{chen2020label}.
}
\label{fig:Fig26}
\vspace{-8pt}
\end{figure}

To address the limitations of current models that lack a robust ability to model label co-occurrences and capture interdependencies between labels and regions, 
Chen et al.~\cite{chen2020label} introduced a label co-occurrence learning framework based on GCNs to find dependencies between pathologies from chest X-ray imaging. This framework consists of two modules, an image feature embedding module that learns high-level features from images and a label co-occurrence learning module that classifies different pathology categories. 
In the framework which is illustrated in Fig.~\ref{fig:Fig26}, each pathology is illustrated with semantic vectors via a word embedding, and the graph representation is learned from the co-occurrence matrix of training data. 
The classifiers are combined with image-level features to adaptively revise prediction beliefs for each pathology in two large-scale chest X-Ray datasets, ChestX-ray14~\cite{wang2017chestx} and CheXpert~\cite{irvin2019chexpert}. Although this approach model the correlations among disease labels, the utilization of medical reports paired with radiology images was not covered.

Zhang et al.~\cite{zhang2020radiology} adapted attention mechanisms and GCNs to learn graph embedded features to improve classification and report generation. In this approach, a CNN feature extractor and attention mechanism are used to compute initial node features. Then, a graph is developed with prior knowledge on chest findings to learn discriminatory features and the relationship between them for classifying disease findings. Each node in the graph corresponds to a finding category. Once the classification network is trained, a two-level decoder with recurrent units (LSTMs) is trained to generate reports. The decoder learns to attend to different findings on the graph, and focuses on one concept in each sentence. The performance demonstrated with the IU-RR dataset~\cite{demner2016preparing} indicates that graphs with prior knowledge help to generate more accurate reports.
Hou et al.~\cite{hou2021multi} employed a transformer encoder as the feature-fusion model of both visual features and label embeddings (semantic features pre-trained on large free-text medical reports). These features are fed to a GCN model which is built as the knowledge graph to model the correlations among different thoracic diseases. The graph is constructed by a data-driven method from medical reports, with primary and auxiliary nodes that correspond to disease labels and other medical labels, respectively. However its extension to handle other domains is limited because the graph is not built automatically.

\subsubsection{Breast cancer}
\begin{figure}[!t]
\centering
\includegraphics[width=1\linewidth]{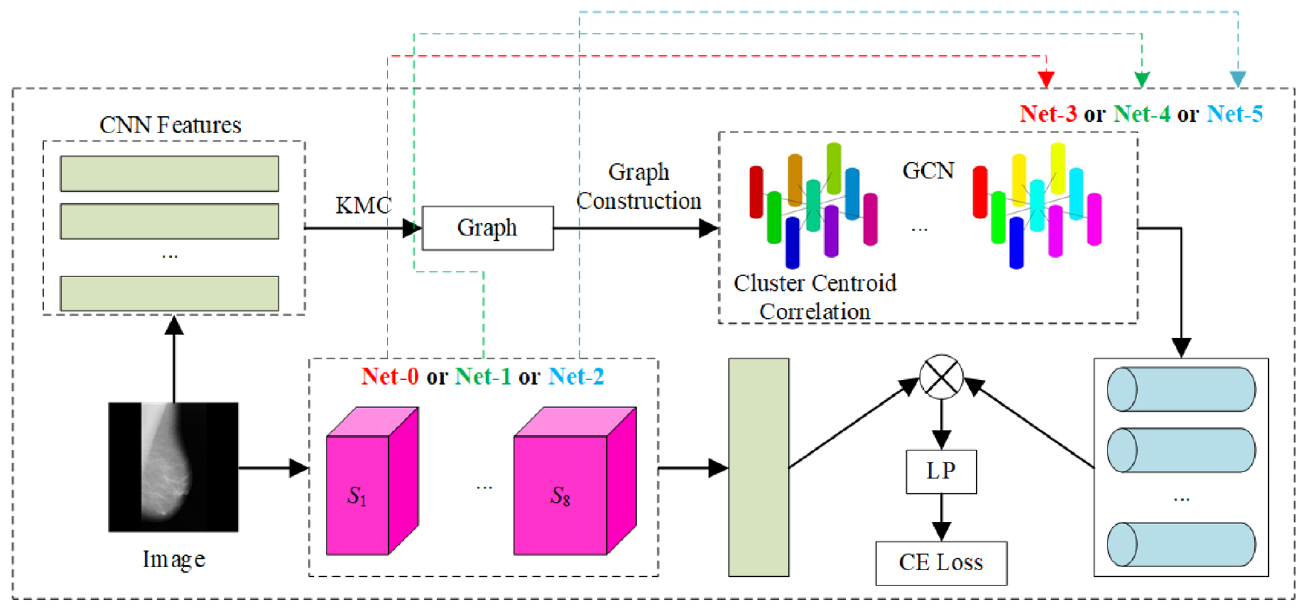}
\vspace{-15pt}
\caption{
Illustration of the framework that combines CNN and GCN features. Bottom row shows the CNN pipeline to extract image-based features while the top row illustrates the GCN pipeline to learn the interactions. Image adapted from~\cite{zhang2021improved}.
}
\label{fig:Fig38}
\vspace{-6pt}
\end{figure}

For abnormal breast tissue detection, the aim is to not only learn the image-level representation automatically, but also the relation-aware representation to more accurately detect abnormal masses using mammography. 
Zhang et al.~\cite{zhang2021improved} fused a CNN pipeline with a GCN pipeline to attain superior performance in classifying six abnormal types in the mini-MIAS dataset~\cite{suckling1994mammographic}. First, a CNN extracts individual image-level features; then, a GCN estimates a relation-aware representation. These features are combined via a dot product and a linear projection with trainable weights. This framework is illustrated in Fig.~\ref{fig:Fig38}.
Although the proposed model achieves high accuracy when analysing mammographic data, further optimization on larger datasets was not considered and other combination mechanisms of GCN and CNN should be assessed.

In clinical practice, experts review medical images by zooming into ROIs for a close-up examination. Thus, Du et al.~\cite{du2019zoom} model the zoom-in mechanism of radiologists' operation with a hierarchical graph-based model to detect abnormal lesions with full field digital mammogram (FFDM) images from the INbreast dataset~\cite{moreira2012inbreast}. 
A pre-trained CNN trained on lesion patches is used to extract features and a GAT model classifies nodes to predict whether to zoom or not into the next level to predict a benign or malignant mammogram. By adding the zoom-in  mechanism, model interpretability is improved. However, the INbreast dataset is relatively small making this method difficult to assess, and a new loss is required to supervise the zoom-in mechanism.

\subsubsection{Kidney disease}
In nephrology, ultrasound (US) data is widely used for diagnostic studies of the kidneys and urinary tract and the anatomic measurement of the renal parenhymal area is correlated with kidney function. Machine learning studies have shown promising performance for the tasks of segmentation and classification of US data; however, kidney disease diagnosis is still a challenging task due to the heterogeneous appearance of multiple 2D US scans of the same kidney from different views.
Multiple instance learning has been used to estimate instance-level classification probabilities and fuse them to generate a bag-level classification probability, but correlation between instances has not been well explored.
To improve these methods, Yin et al.~\cite{yin2019multi} introduced a graph-based methodology to detect children with congenital anomalies of the kidneys and urinary tract in 2D US images. A CNN is used to learn informative US image features at the instance level and a GCN is used as a permutation-invariant operator to further optimize the instance-level CNN features by exploring potential correlations among different instances of the same bag. The authors also adopted attention-based multiple instance learning pooling to learn a bag-level classifier using and instance-level supervision to enhance the learning of instance features and the bag-level classification.

\subsubsection{Relative brain age}
Predicted brain age is a meaningful index that characterises the current status of brain development which may be associated with functional brain abilities in the future. Measurements of morphological changes, including sulcal depth and cortical thickness, can be key features for brain age prediction. Traditional approaches applied to surface morphological features have not taken into account the topology of surfaces, which is defined with meshes. Therefore, CNN-based methods may not be appropriate for the analysis of cortical surface data.
A relative brain age has been used as a metric computed as the predicted age minus the true age of the subject. 
Liu et al.~\cite{liu2020deep} exploited the brain mesh topology as a sparse graph to predict brain age from MRIs for preterm neonates using vertex-wise cortical thicknesses and sulcal depth as input to a GCN. This model enables the convolutional filtering of input features through the surface topology in the context of spectral graph theory. 
The GCN predicted the ages of preterm neonates better than machine learning and deep learning methods that did not use surface topological knowledge.
The authors also generated cortical sub-meshes that represent brain regions to predict which region estimates the age more accurately, and if they are associated with brain functional abilities in the future.
As discussed previously in the Subsection of AD analysis, Gopinath et al.~\cite{gopinath2019adaptive, gopinath2020learnable} also demonstrated their adaptive graph convolution pooling in a regression problem where the brain age is estimated using the geometry of the brains with point-wise surface-based measurements. 
The model is trained using data labeled as normal cognition from the ADNI dataset~\cite{petersen2010alzheimer}, and the graph model uses cortical thickness, sulcal depth and spectral information to predict the brain age.

\subsubsection{Brain data prediction}
Diffusion MRI (DMRI) provides unique insights into the developing brain, owing to its sensitivity examination of brain tissue microstructures and white matter properties which are useful for diagnosis of brain disorders.
However, DMRI suffers from long acquisition times and is more susceptible to low signal-to-noise ratio, motion artifacts, and partial volume effects. Missing data is also a common problem in longitudinal studies due to unsuccessful scans and subject dropouts, and the high variability in diffusion wave-vector sampling (q-space) makes the longitudinal prediction of DMRI data a challenging task. Therefore, some methods have been developed for DMRI reconstruction. 

\begin{figure}[!t]
\centering
\includegraphics[width=1\linewidth]{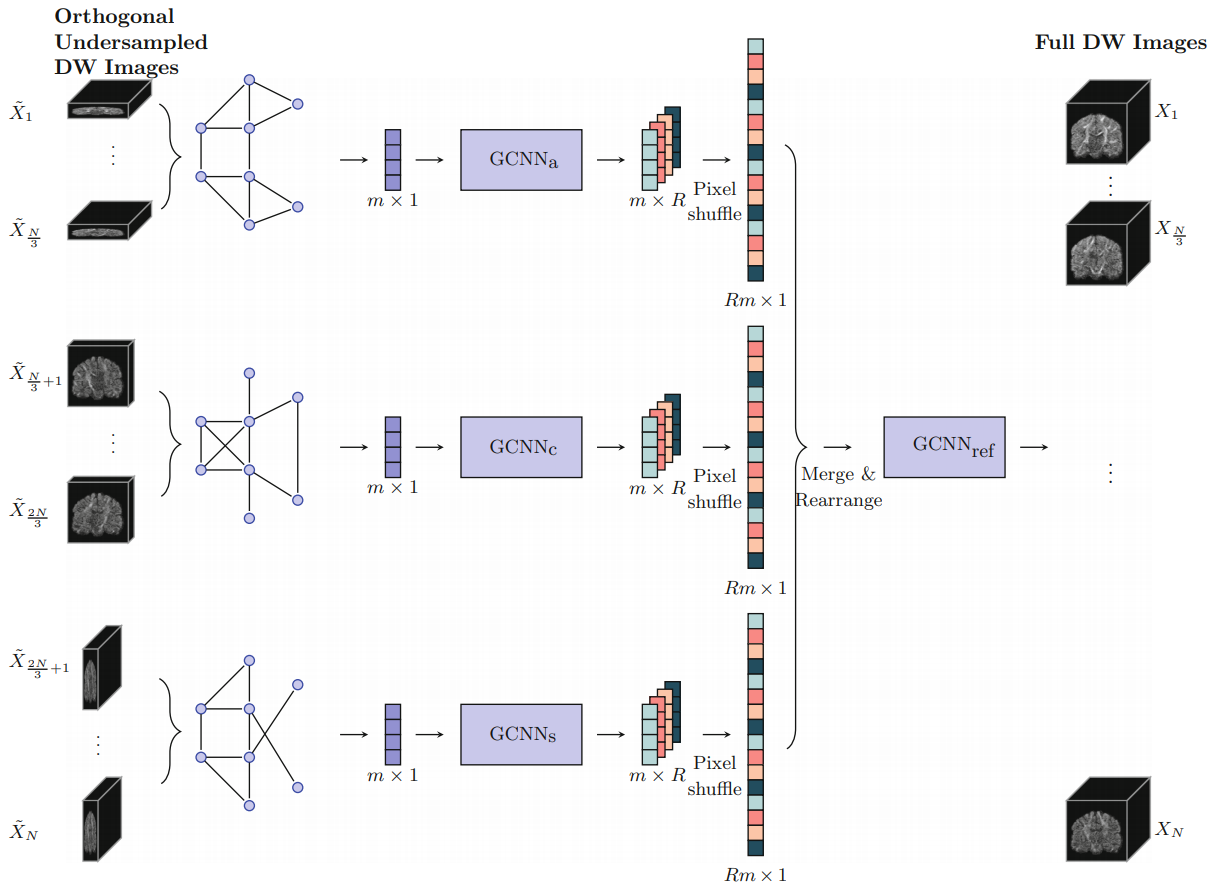}
\vspace{-15pt}
\caption{
Individual GCNs model the axial, coronal and sagitall scan direction. A refinement GCN is used to generate the proposed super-resolution reconstruction. Image adapted from~\cite{hong2019reconstructing}.
}
\label{fig:Fig31}
\vspace{-8pt}
\end{figure}

To improve acquisition speed, Hong et al.~\cite{hong2019multifold} introduced a method for DMRI reconstruction from under-sampled slide data, where only a sub-sample of equally-spaced slices are used to acquire a full diffusion-weighted (DW) image volume. A GCN learns the non-linear mapping from the sub-sampled to full DW image, and spatio-angular relationships are considered when constructing the graph. To improve perceptual quality, the GCN is employed as the generator in a generative adversarial network.
The same authors~\cite{hong2019reconstructing} proposed a super-resolution reconstruction framework based on an orthogonal under-sampling scheme to increase complementary information within the under-sampled DW volume. The set of wave-vectors is divided into three subsets of scan directions (axial, coronal, or sagittal) and they are fitted to individual GCNs. A refinement GCN is used to generate the final DW volume by considering the correlation across scan directions as illustrated in Fig.~\ref{fig:Fig31}. These graph-based methods outperform traditional interpolation methods and 3D U-Net based reconstruction methods.

Kim et al.~\cite{kim2019graph} introduced a graph-based model for longitudinal prediction of DMRI data by considering the relationship between sampling points in the spatial and angular domains, \textit{i.e.}, a graph-based representation of the spatio-angular space. Then, the authors implemented a residual learning architecture with graph convolutions to capture brain longitudinal changes to predict missing DMRI data over time in a patch-wise manner. The proposed model showed improved performance in predicting missing DMRI data from neonate images so that longitudinal analysis can be performed.
Hong et al.~\cite{hong2019longitudinal} also proposed a GCN-based method for predicting missing infant brain DMRI data. This model exploits information from the spatial domain and diffusion wave-vector domain jointly for effective prediction. Generative adversarial networks (GANs) are also adopted to better model the non-linear prediction mapping and performance improvement. Here, the generator estimates the source image and the discriminator distinguishes the source image from the estimated one, where the generator is the GCN and the discriminator is developed via consecutive graph convolutional layers. However, the model cannot predict missing DMRI data for arbitrary time points.

Although DMRI is a powerful tool for the characterization of tissue microstructures, several microstructure models need DMRI data densely sampled in \textit{q}-space that is defined by the number of acquired diffusion-weighted images. Traditional deep learning models learn the relationship between sparsely sampled \textit{q}-space data and high-quality microstructure indices estimated from densely sampled \textit{q}-space, but these models do not consider the \textit{q}-space data structure.
Chen et al.~\cite{chen2020estimating} adopted GCNs to estimate tissue microstructure from DMRI data represented as graphs. The graph encodes the geometric structure of the \textit{q}-space sampling points which harnesses information from angular neighbors to improve estimation accuracy. Results on the baby connectome project dataset~\cite{howell2019unc} demonstrated high-quality intra-cellular volume fraction maps that are close to the gold standard.

\subsubsection{Other applications} \hfill

\paragraph{High-resolution of 3D MR fingerprinting}
Most quantitative MRI methods are comparatively slow and provide a single tissue property at a time, which limit their adoption in routine clinical settings. 
Magnetic resonance fingerprinting (MRF) is a rapid and efficient quantitative imaging method that has been used for simultaneous quantification of multiple tissue properties in a single acquisition~\cite{ma2013magnetic}. 2D MRF has been extended to 3D using stack-of-spirals acquisitions, but the high spatial resolution and volumetric coverage prolongs the acquisition time. 
Cheng et al.~\cite{cheng2020acceleration} adopted a GCN to accelerate high-resolution 3D MRF acquisition by interpolating the under-sampled data along the slice-encoding direction. A network is further applied to generate tissue property maps. For efficient tissue quantification, a U-net is used along the temporal domain.

\paragraph{Medical image enhancement}
Considering the limited MRI resources, the thick slices and low scan time, MRI images have to be utilized to get a desired signal-to-noise ratio. Consequently, the use of image enhancement techniques is an established field of research in medical image computing and imaging physics, for example, to prevent blurring and information loss when co-aligning different image volumes in a multi-parametric sequence.
Hu et al.~\cite{hu2020feedback} introduced a feedback graph attention convolutional network to enhance the visual quality and remove common distortions such as artifacts by considering self-similarity and correlation across image sub-regions. A feedback mechanism is employed to recover texture details by refining low-level representations with high-level information across a time-series. Experiments on a cross-protocol super resolution of a diffusion MRI dataset (MUSHAC~\cite{tax2019cross}) and an artifact removal dataset (FLAIR~\cite{kuijf2019standardized}) demonstrated the capability of the system to remove artifacts and to generate high-resolution MRI.

\begin{table*}[t!]
\caption{Summary of GCN approaches adopted for anatomical structure analysis and their applications (Group 2).}
\vspace{-5pt}
\centering
\label{table:anatomical2}
\resizebox{1\textwidth}{!}{%
\begin{tabular}{
l c l 
>{\raggedright\arraybackslash}p{4.8cm} 
>{\raggedright\arraybackslash}p{9cm}}
\toprule
\textbf{Authors} &
\textbf{Year} &
\textbf{Modality} & 
\textbf{Application} &  
\textbf{Dataset} \\
\midrule
%
%
Wolterink et al.~\cite{wolterink2019graph} & 2019 & CTA  &
Segmentation: Coronary artery & Coronary Artery Stenoses Detection~\cite{kiricsli2013standardized} \\
Zhai et al.~\cite{zhai2019linking} & 2019 & CT  &
Segmentation: Pulmonary artery-vein & Sun Yat-sen University Hospital (private) \\
Noh et al.~\cite{noh2020combining} & 2020 & FA / Fundus &
Segmentation: Retinal vessels & Fundus and FA (private), RITE A/V~\cite{hu2013automated} \\
Shin et al.~\cite{shin2019deep} & 2019 & RGB/FA/XRA  &
Segmentation: Retinal vessels & DRIVE~\cite{staal2004ridge}, STARE~\cite{hoover2000locating}, CHASE\textunderscore DB1~\cite{fraz2012ensemble}, HRF~\cite{budai2013robust} \\ 
Chen et al.~\cite{chen2020automated} & 2020 & MRA &
Segmentation: Intracranial arteries & MRA~\cite{chen2019quantitative}, UNC~\cite{bullitt2005vessel} \\
Yao et al.~\cite{yao2020graph} & 2020 & CTA  &
Segmentation: Head and neck vessels & Head and neck CTA (private) \\
%
%
Lyu et al.~\cite{lyu2021labeling} & 2021 &  MRI  &
Segmentation: Cerebral cortex & NORA-pediatric~\cite{wendelken2017frontoparietal}, HCP-adult~\cite{van2012human}   \\
Gopinath et al.~\cite{gopinath2020graph} & 2020 &  MRI &
Segmentation: Cerebral cortex & MindBoggle~\cite{klein2017mindboggling} \\
Gopinath et al.~\cite{gopinath2020learnable} & 2020 &  MRI &
Segmentation: Cerebral cortex & MindBoggle~\cite{klein2017mindboggling} \\
Hao et al.~\cite{hao2020automatic} & 2020 &  T1WI  &
Segmentation: Cerebral cortex & University of California Berkeley Brain Imaging Center (private) \\
He et al.~\cite{he2020spectral} $\dagger$ & 2020 &  MRI &
Segmentation: Cerebral cortex & MindBoggle~\cite{klein2017mindboggling} \\
Gopinath et al.~\cite{gopinath2019graph} & 2019 & MRI  &
Segmentation: Cerebral cortex & MindBoggle~\cite{klein2017mindboggling} \\
Wu et al.~\cite{wu2019intrinsic} & 2019 & MRI  &
Segmentation: Cerebral cortex & Neonatal brain surfaces (private) \\
Parvathaneni et al.~\cite{parvathaneni2019cortical} & 2019 &  T1WI  &
Segmentation: Cerebral cortex & Cortical surface (private) \\
Zhao et al.~\cite{zhao2019spherical} & 2019 &  MRI  &
Segmentation: Cerebral cortex & Infant brain MRI (private) \\
Cucurull et al.~\cite{cucurull2018convolutional} $\dagger$ & 2018 &  MRI &
Segmentation: Cerebral cortex & HPC mesh~\cite{van2013wu,jakobsen2016subdivision} \\
Selvan et al.~\cite{selvan2020graph} & 2020 & CT  &
Segmentation: Pulmonary airway & Danish Lung Cancer Screening trial~\cite{pedersen2009danish}  \\
Juarez et al.~\cite{juarez2019joint} & 2019 & CT  &
Segmentation: Pulmonary airway & Danish Lung Cancer Screening trial~\cite{pedersen2009danish}  \\
Selvan et al.~\cite{selvan2018extraction} & 2018 & CT  &
Segmentation: Pulmonary airway & Danish Lung Cancer Screening trial~\cite{pedersen2009danish}  \\
Yan et al.~\cite{yan2019brain} & 2019 & MRI  &
Segmentation: Brain tissue & BrainWeb18~\cite{kwan1999mri}, IBSR18~\cite{cocosco1997brainweb}\\
Meng et al.~\cite{meng2020regression,meng2020cnn} $\dagger$ & 2020 & FA &
Segmentation: Optic disc/cup & Refuge~\cite{orlando2020refuge}, Drishti-GS~\cite{sivaswamy2014drishti}, ORIGA~\cite{zhang2010origa}, RIGA~\cite{almazroa2018retinal}, RIM-ONE~\cite{fumero2011rim} \\
Meng et al.~\cite{meng2020regression,meng2020cnn} $\dagger$  & 2020 & US  &
Segmentation: Fetal head & HC18-challenge~\cite{van2018automated}  \\
Soberanis-Mukul et al.~\cite{soberanis2020uncertainty,soberanis2020uncertaintyini} & 2020 & CT  &
Segmentation: Pancreas / Spleen  & NIH pancreas~\cite{clark2013cancer}, MSD-spleen~\cite{simpson2019large} \\
Tian et al.~\cite{tian2020graph} & 2020 & MRI  &
Segmentation: Prostate cancer & PROMISE12~\cite{litjens2014evaluation}, ISBI2013~\cite{bloch2015nci}, in-house (private) \\
Chao et al.~\cite{chao2020lymph} & 2020 & CT/PET  &
Segmentation: Lymph node gross tumor & Esophageal radiotherapy (private) \\
\bottomrule
\multicolumn{5}{p{350pt}}
{ 
$\star$ GCN with temporal structures for medical diagnostic analysis. \newline
$\dagger$ GCN with attention structures for medical diagnostic analysis.
}
\end{tabular}}
\end{table*}

\vspace{-6pt}
\subsection{Anatomical structure analysis (segmentation)}

Among different medical image segmentation and labeling methods, graph-based methods are showing promising results in clinical applications. Graph-based segmentation approaches play an important role in medical image segmentation. A graph maps pixels or regions in the original image to nodes in the graph. Then, the segmentation problem can be transformed into a labeling problem which requires assigning the correct label to each node according to its properties~\cite{chen2018survey}. GCNs can propagate and exchange the local short-range information through the whole image to learn the semantic relationships between objects. We cover only application with evidence of graph representation learning in anatomical structures including the vasculature system and organs.  

\subsubsection{Vasculature segmentation} \hfill

\paragraph{Coronary arteries}
Quantitative examination of coronary arteries is an important step for the diagnosis of cardiovascular diseases, stenosis grading, blood flow modeling and surgical planning.
Coronary CT angiography (CTA) images are used to determine the anatomical or functional severity of coronary artery stenosis (\textit{i.e.} a narrowing in the artery). Methods for coronary artery segmentation are related to lumen (\textit{i.e.}, vessel wall) segmentation. 
Deep learning-based segmentation predicts dense segmentation probability maps (voxel-based segmentation methods), or incorporates a shape prior to exploit the fact that vessel segment has a roughly tubular shape. Thus, the segmentation can be obtained by deforming the wall of this tube to match the visible lumen in the CTA image.

Graph convolutional networks have also been investigated by Wolterink et al.~\cite{wolterink2019graph} for coronary artery segmentation in CTA. 
The authors proposed to use GCNs to directly optimize the position of the tubular surface mesh vertices. The locations of these tubular surface mesh vertices were directly optimized using vertices on the coronary lumen surface mesh as graph nodes. Predictions for vertices rely on both local features and representations of adjacent vertices on the surface. The authors demonstrated that by considering the information from neighboring vertices, the GCN generates smooth surface meshes without post-processing.

\paragraph{Pulmonary arteries and veins}
Separation of pulmonary arteries and veins is challenging due to their similarity in morphology and the complexity of their anatomical structures. Using chest CT, vasculopathy or disease affecting blood vessels can be quantified automatically by detecting pulmonary vessels.
Zhai et al.~\cite{zhai2019linking} proposed a method that links CNNs with GCNs and can be trained in an end-to-end manner. The model includes both local image and graph connectivity features for pulmonary artery-vein separation. Instead of using entire graphs, the authors proposed a batch-based technique for CNN-GCN training and validation. In this approach, the size of the adjacency matrix can be reduced as the nodes in the GCN are from sub-sampled pixel or voxel grids. 

\paragraph{Retinal vessels}
Assessment of retinal vessels is needed to diagnose various retinal diseases including hypertension and cerebral disorders. 
Fluorescein angiography (FA) and fundus images have been used for artery and vein classification and segmentation techniques because arteries and veins are highlighted separately at different times due to the flow of the fluorescent dye through the vessels. 
Noh et al.~\cite{noh2020combining} combined both the fundus image sequence and FA image as input for artery and vein classification. The proposed method comprises a feature extractor CNN for the input images and a hierarchical connectivity GNN based on Graph U-Nets~\cite{gao2019graph} to incorporate higher order connectivity into classification.
%
Shin et al.~\cite{shin2019deep} also incorporated a GCN into a unified CNN architecture for 2D vessel segmentation on retinal image datasets. A CNN was trained for feature extraction of local appearance and vessel probabilities and a GCN was trained to predict the presence of a vessel based on global connectivity of vessel structures. The vessel segmentation is generated by using the relationship between the neighborhood of vessels pixels. This is based on the local appearance of vessels instead of vessel structure. The method achieved competitive results, but the classifier cannot be trained end-to-end.

\paragraph{Intracranial arteries}
Characterization of intracranial arteries (ICA), including labeling each artery segment with its anatomical name, is beneficial for clinical evaluation and research. Many natural and disease related (\textit{e.g.} stenosis) variations in ICA are challenging for automated labeling.
Chen et al.~\cite{chen2020automated} proposed a GNN model with hierarchical refinement to label arteries in magnetic resonance angiography (MRA) data by classifying types of nodes and edges in an attributed relational graph. GNNs based on the message passing framework~\cite{battaglia2018relational} take a graph with edge and node features as input and return a graph with other features for node and edge types.

\paragraph{Head and neck vessels}
Vessel segmentation and anatomical labeling are important for vascular disease analysis. The direct use of CNNs for segmentation of vessels in 3D images encounters great challenges. Specifically, head and neck vessels have long and tortuous tubular-like vascular structures with different sizes and shapes. Therefore, it is challenging to automatically and accurately segment and label vessels to expedite vessel quantification. Point cloud representations of head and neck vessels enables quantification of spatial relationships among vascular points.
Yao et al.~\cite{yao2020graph} proposed a GCN-based point cloud learning framework to 
label head and neck vessels and improve CNN-based vessel segmentation on CTA images.
To refine vessel segmentation a point cloud network is first incorporated to the points formed by initial vessel voxels. Then, a GCN is adopted on the point cloud to leverage the anatomical shapes and vascular structures to label the vessel into 13 major segments. 
 
\subsubsection{Organ segmentation} \hfill

\paragraph{Cerebral cortex}
The cerebral cortex is the outermost layer of the brain, and is the most prominent visible feature of the human brain. 
Different regions of the cortex are involved in complex cognitive processes. Reconstructions of the cortical surface captured with sMRI are used to analyse healthy brain organization as well as abnormalities in neurological and neuropsychiatric conditions. Separating the cerebral cortex into anatomically distinct regions based on structure or function is known as parcellation. Traditional CNN approaches have dealt with the mesh segmentation problem by using irregular data represented using graph or mesh structures.

Cucurull et al.~\cite{cucurull2018convolutional} investigated the usefulness of graph networks in which contextual information can be exploited for cortical mesh segmentation using the Human Connectome Project data~\cite{van2013wu,jakobsen2016subdivision} (\textit{i.e.} functional and structural features from cortical surface patches are used for segmentation).
The model receives a mesh as input and produces one output label for each node of the mesh, and parcellates the cerebral cortex into three parcels using a graph attention-based model (GAT)~\cite{velivckovic2017graph}. However, brain meshes are constrained within a particular graph structure, ignoring the complex geometry of the surface and hinder all meshes to use the same mesh geometry. Furthermore, the authors conducted cortical parcellation on only selected regions due to memory capacity.

\begin{figure}[!t]
\centering
\includegraphics[width=1\linewidth]{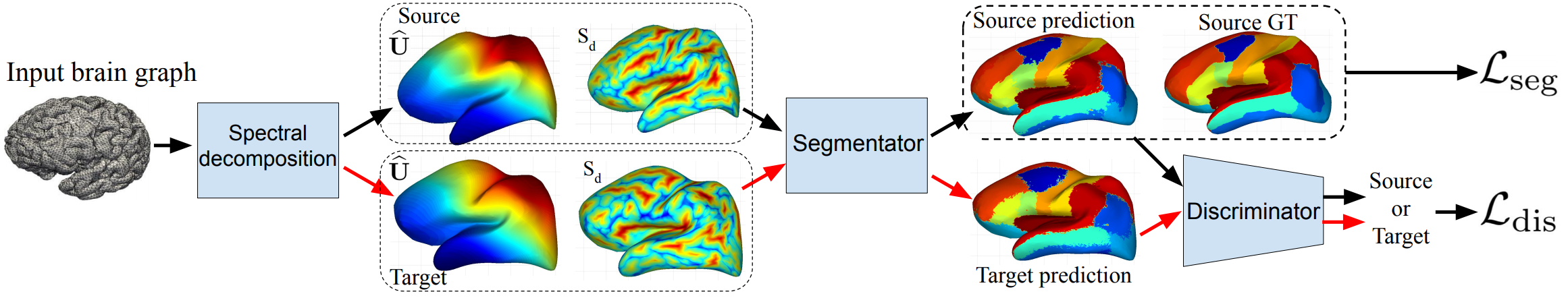}
\vspace{-10pt}
\caption{
Adversarial graph domain adaptation for segmentation. A cortical brain graph is mapped to a spectral domain. The source and target domain are aligned to a reference template. A GCN segmentator learns to predict a generic cortical parcel label for each domain. Finally, the discriminator classifies the segmentator predictions. Image adapted from~\cite{gopinath2020graph}.
}
\label{fig:Fig30}
\vspace{-6pt}
\end{figure}

Gopinath et al.~\cite{gopinath2019graph} leveraged recent advances in spectral graph matching to transfer surface data across aligned spectral domains, and to learn a node-wise prediction. Authors proposed better capabilities for full cortical parcellation on adult brains with GCNs on the MindBoggle dataset~\cite{klein2017mindboggling}.
The authors also extend this previous work and proposed a method that learns an intrinsic aggregation of graph nodes based on graph spectral embeddings for cortical region size regression~\cite{gopinath2020learnable}.

Despite offering more flexibility to analyse unordered data, GCNs are also domain-dependent and are limited to generalize to new domains (datasets) without explicit re-training. Spectral GCNs cannot be used to compare multiple graphs directly and need an explicit alignment of graph eigenbases as an additional pre-processing step. 
Thus, Gopinath et al.~\cite{gopinath2020graph} proposed an adversarial graph domain adaptation method for surface segmentation. This approach focused on generalizing parcellation across multiple brain surface domains by eliminating the dependency on domain-specific alignment. In this approach, two networks are trained in an adversarial manner, a fully-convolutional GCN segmentator and a GCN domain discriminator. These networks operate on the spectral components of surface graphs as illustrated in Fig.~\ref{fig:Fig30}. The authors also demonstrate that the model could be useful for semi-supervised surface segmentation, by that alleviating the need for large numbers of labeled surfaces.

\begin{figure}[!t]
\centering
\includegraphics[width=1\linewidth]{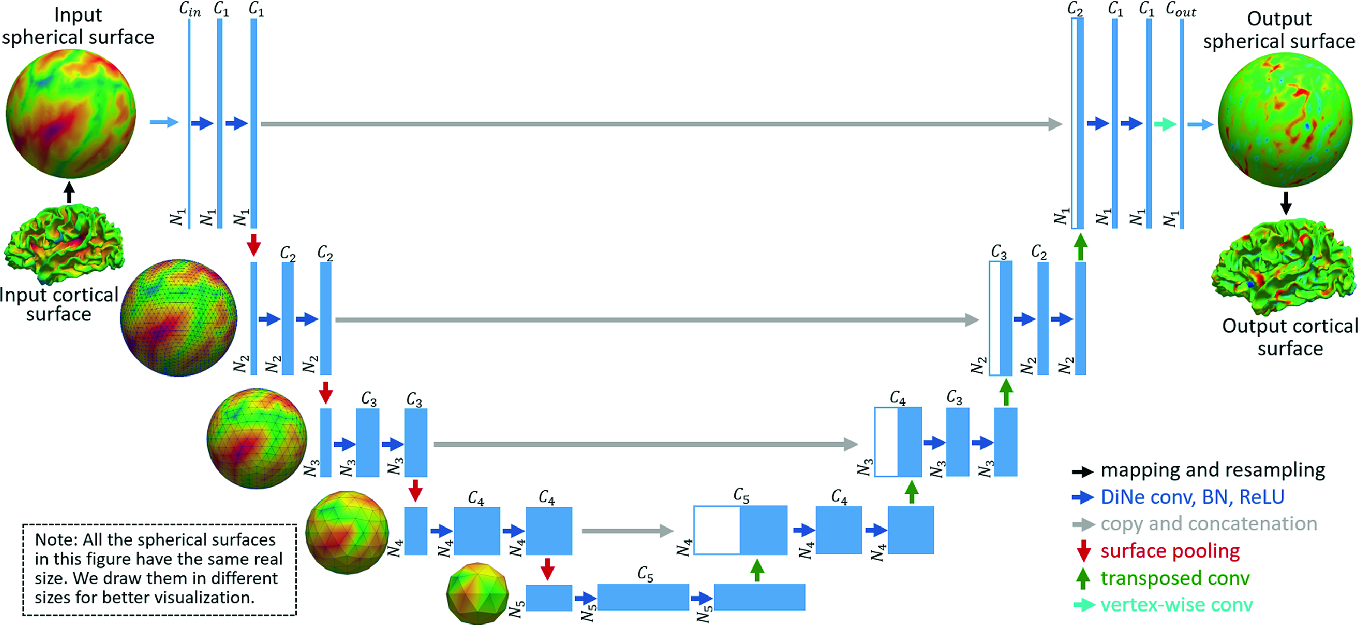}
\vspace{-10pt}
\caption{
Spherical U-Net architecture. The output surface is a cortical parcellation map or a cortical attribute map, and the blue boxes reflect feature maps in spherical space. Image adapted from~\cite{zhao2019spherical}.
}
\label{fig:Fig25}
\vspace{-6pt}
\end{figure}

Zhao et al.~\cite{zhao2019spherical} suggested a convolution filter on a sphere, termed Direct Neighbor, which is used to develop surface convolution, pooling and transposed convolution in spherical space. The authors extend the U-Net architecture to spherical surface domains as illustrated in Fig.~\ref{fig:Fig25}. 
The spherical U-Net is efficient in learning useful features to predict cortical surface parcellation and cortical attribute map development.
Although the method does not rely on spherical registration, it still needs to map cortical surfaces onto a sphere. Spherical mapping is susceptible to topological noise and cortical surfaces are required to be topologically correct before mapping.
Therefore, Wu et al.~\cite{wu2019intrinsic} proposed to parcellate the cerebral cortex on the original cortical surface manifold without the need for spherical mapping by taking advantage of the high learning potential of GCNs (\textit{i.e.} the model is free of spherical mapping and registration).
The GCN receives intrinsic patches from the original cortical surface manifold that are mapped using the intrinsic local coordinate system. The extracted intrinsic patches are then combined with the trained models to predict parcellation labels.

Spectral graph matching has been used to transfer surface data across aligned spectral domains, enabling the learning of spectral GCNs across multiple surface data. However, this involves an explicit computation of a transformation map for each brain towards one reference template.
He et al.~\cite{he2020spectral} introduced a spectral graph transformer (SGT) network to learn this transformation function across multiple brain surfaces directly in the spectral domain, mapping input spectral coordinates to a reference set. The spectral decomposition of a brain graph is randomly sub-sampled as an input point cloud to a SGT network. The SGT learns the transformation parameters aligning the eigenvectors of multiple brains. The learnt transformation matrix is multiplied by the original spectral coordinates and fed to the GCN for parcellation.

While Laplacian-based graph convolutions are more efficient than spherical convolutions, they are not exactly equivariant. Graph-based spherical CNNs strike an interesting balance, with a controllable trade-off between cost and equivariance (which is linked to performance)~\cite{defferrard2020deepsphere}.
Parvathaneni et al.~\cite{parvathaneni2019cortical} adopted a deep spherical U-Net~\cite{jiang2019spherical} to encode a relatively large surface mesh. Using a spherical surface registration process, the authors computed deformation fields to produce deformed geometric features that best match ground-truth parcel boundaries.
The same authors also implement a spherical U-Net for cortical sulci labeling from relatively few samples in a developmental cohort~\cite{hao2020automatic}. To enhance the capability of the spherical U-Net with limited samples, the authors augmented the geometric features from the training data with their deformed features guided by the intermediate deformation fields.
In another work~\cite{lyu2021labeling}, the authors proposed a context-aware training and co-registered every possible pair of training samples for the automated labeling of sulci in the lateral prefrontal cortex in pediatric and adult cohorts.

\begin{figure}[!t]
\centering
\includegraphics[width=1\linewidth]{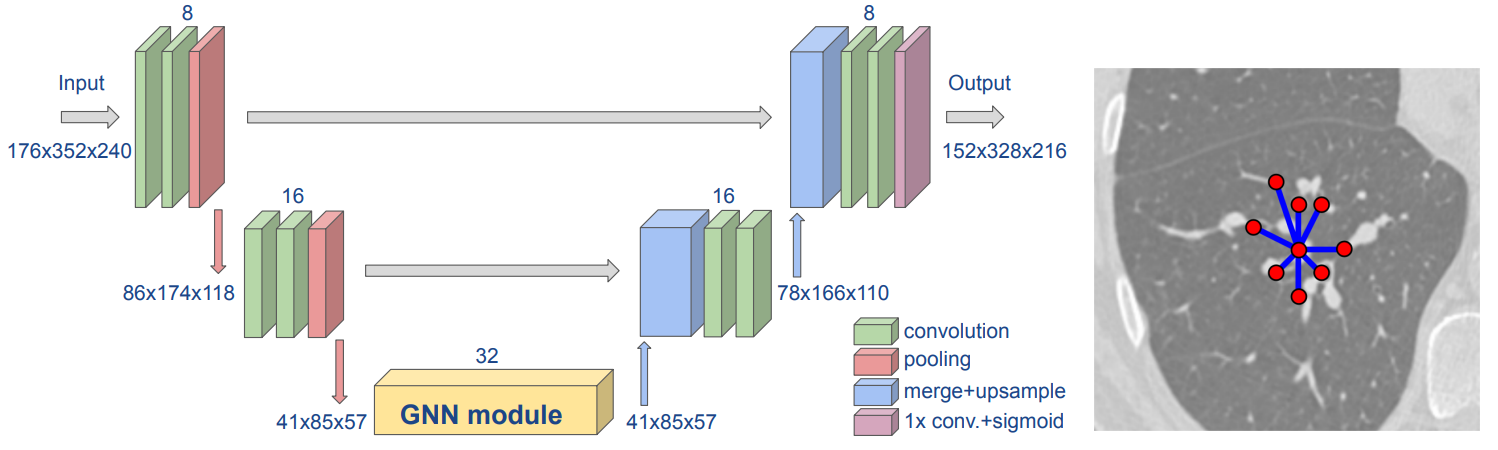}
\vspace{-10pt}
\caption{
Schematic of a UNET-GNN and illustration of irregular node connectivity for a given voxel in the initial graph. Image adapted from~\cite{juarez2019joint}.
}
\label{fig:Fig24}
\vspace{-6pt}
\end{figure}

\paragraph{Pulmonary airway}
The segmentation of tree-like structures such as the airways from chest CT images is a complex task, with branches of varying sizes and different orientations.
Quantifying morphological changes in the chest can indicate the presence and stage of related diseases (\textit{e.g.} bronchial stenosis). Unlike spheroid-like organs such as liver and kidney, tree-like airways are divergent, thin and tenuous.
GNNs were investigated as a way to integrate neighborhood information in feature utilization for mapping airwaves in the lungs~\cite{juarez2019joint,selvan2020graph}.
Juarez et al.~\cite{juarez2019joint} explored the application of GCNs to improve the segmentation of tubular structures like airways. The authors designed a UNet-GNN architecture by replacing the convolutional layers at the deepest level of the 3D-UNet with a GCN module. The GNN module uses a graph structure obtained from the dense feature maps resulting from the contracting path of the UNet.
The GNN learns variations of the input feature maps based on the graph topology, and then outputs a new graph with the same nodes as the input graph, as well as a vector of learnt features for each node. 
These output feature maps are fed to the up-sampling path of the UNet as illustrated in Fig.~\ref{fig:Fig24}. By introducing a GCN, the method is able to learn and combine information from a larger region of CT chest scans, and is evaluated on the Danish Lung Cancer Screening trial dataset~\cite{pedersen2009danish}.

Selvan et al.~\cite{selvan2020graph} also used  this volumetric dataset to explore the extraction of tree-structures with a focus on airway extraction, formulated as a graph refinement task, extending the authors own prior work~\cite{selvan2018extraction}. 
The input image data is first processed to create a graph-like representation, which consists of nodes containing information derived from local image neighborhoods. Then, a GCN predicts the refined subgraph that corresponds to the structure of interest in a supervised setting, where edge probabilities are predicted from learnt edge embeddings. However, the proposed work treats graph structure learning to be an expensive approximation of a combinatorial optimization problem.

\begin{figure}[!t]
\centering
\includegraphics[width=1\linewidth]{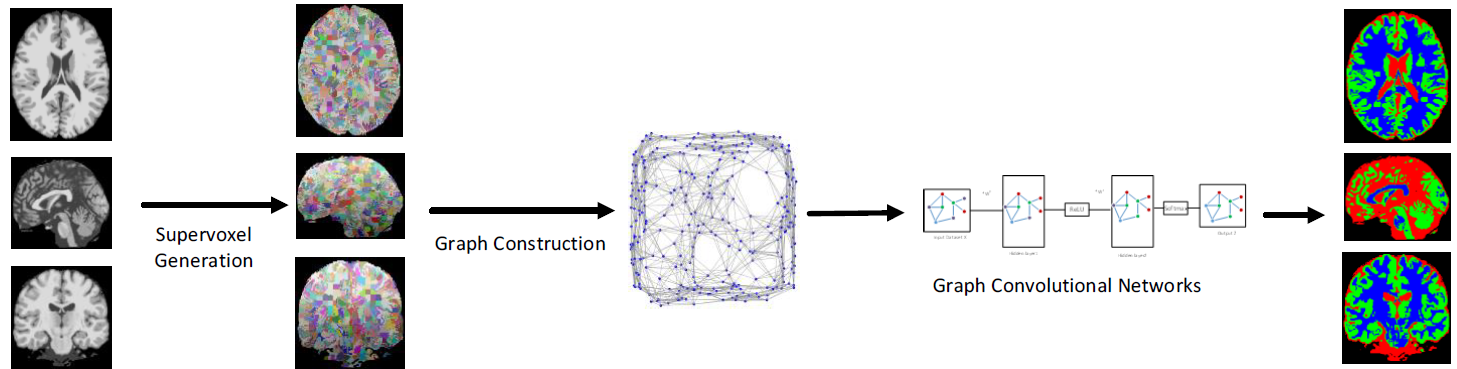}
\vspace{-10pt}
\caption{
Supervoxels are generated from the brain MRI volume. A graph is constructed from these supervoxels with KNNs. A GCN is employed to classify supervoxels into different types of tissue. Image adapted from~\cite{yan2019brain}.
}
\label{fig:Fig21}
\vspace{-6pt}
\end{figure}

\paragraph{Brain tissues}
In brain MRI analysis, image segmentation is used for analyzing brain changes, for measuring a brain's anatomical structures, for delineating pathological regions, and for surgical planning and image-guided interventions. In MRIs of low contrast and resolution, volume effects appear where individual voxels contain different tissues which makes brain tissue segmentation challenging. From several methods in the literature, voxel-wise MRI image segmentation approaches neglects the spatial information within data. As brain MRIs consist of approximately piecewise constant regions, they are well suited to supervoxel generation which has been increasingly used for high dimensional 3D brain MRI volumes.
Yan et al.~\cite{yan2019brain} proposed a segmentation model based on GCNs. First, supervoxels from the brain MRI volume are generated, then a graph is developed from these supervoxels with the \textit{k}-nearest neighbor algorithm used to identify the nodes. Finally, a GCN is adopted to classify supervoxels into different types of tissue such as cerebrospinal fluid, grey matter and white matter. This framework is illustrated in Fig.~\ref{fig:Fig21}.

\begin{figure}[!t]
\centering
\includegraphics[width=0.95\linewidth]{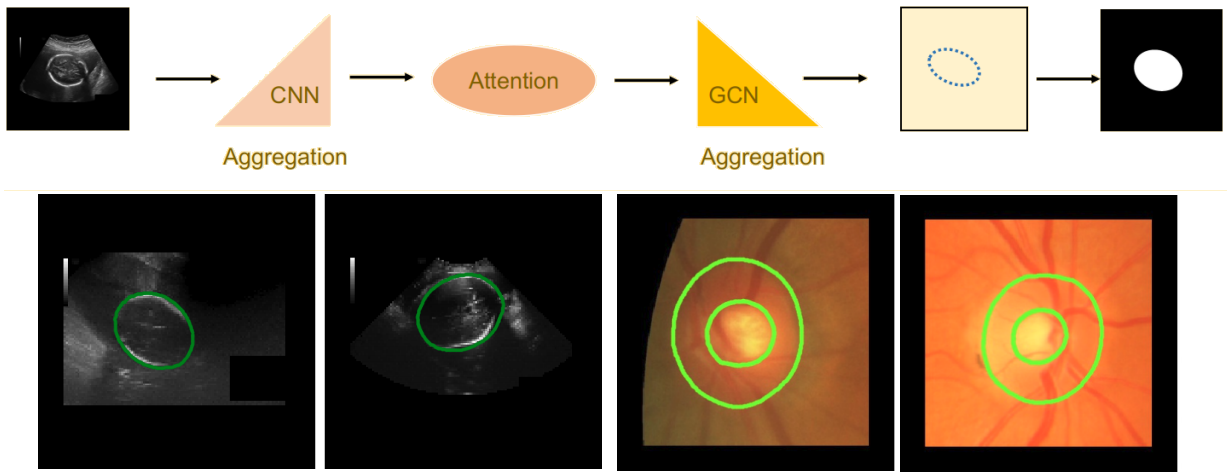}
\caption{
Segmentation paradigm. Method that regresses the locations of object boundaries by information aggregation through a CNN and GCN enhanced by attention modules. Qualitative results of the segmentation of the fetal head and optic disc/cup. Image adapted from~\cite{meng2020regression}.
}
\label{fig:Fig23}
\vspace{-6pt}
\end{figure}

\paragraph{Optic disc/cup and fetal head}
The size of the optic disc and optic cup in color fundus images is also of great importance for the diagnosis of glaucoma, an irreversible eye disease. 
Meng et al.~\cite{meng2020regression} developed a multi-level aggregation network to regress the coordinates of the boundary of instances instead of using a pixel-wise dense prediction. This model combines a CNN with an attention refine module and a GCN. The attention module works as a filter between the CNN encoder and the GCN decoder to extract more effective semantic and spatial features, as illustrated in Fig.~\ref{fig:Fig23}. Compared to a previous work from the same authors~\cite{meng2020cnn}, this model also extracts feature correlations among different layers in the GCN.
Meng et al.~\cite{meng2020regression} also demonstrated the effectiveness of the network in the segmentation of the fetal head in ultrasound images. Fetal head circumference in ultrasound images is a critical indicator for prenatal diagnosis and can be used to estimate the gestational age and to monitor the growth of the fetus~\cite{van2018automated}. In this application the feature map and vertex map size will be different because of different input sizes and the number of contours of instances in the HC18-Challenge dataset~\cite{van2018automated}.

\paragraph{Pancreas and spleen}
Organ segmentation in CT volumes is an important pre-processing phase for assisted intervention and diagnosis. However, the limitations of expert example annotation and the inter-patient variability of anatomical structures may lead to potential errors in the model prediction. The incorporation of a post-processing refinement phase is a traditional approach to improve the segmentation results. This additional knowledge about the accuracy of the prediction may be helpful in the process. Related to this idea, CNN uncertainty estimation has been used as an attention mechanism for finding potentially misclassified regions for segmentation refinement~\cite{dias2018semantic}.
Soberanis-Mukul et al.~\cite{soberanis2020uncertainty,soberanis2020uncertaintyini} proposed a semi-supervised graph learning problem operating over CT data for pancreas and spleen segmentation refinement task. 
First, a Monte Carlo dropout process is applied to a CNN (2D U-Net) to extract the model's expectation and uncertainty which are used to divide the CNN output into high confidence points (\textit{i.e.} to find incorrectly estimated elements). This process is represented by a binary mask indicating voxels with high uncertainty. Then, these confidence predictions are used to train a GCN in a semi-supervised way with partially-labeled nodes to refine (reclassify) the output of the CNN.
The authors investigated various connectivity and weighting mechanisms to construct the graph. A sparse representation is established that takes into account local and long-range relations between high and low uncertainty elements. Gaussian kernels are used to define the weights for the edges considering the intensity and the 3D position associated with the node. Results on pancreas~\cite{clark2013cancer} and MSD-spleen~\cite{simpson2019large} datasets shows better performance over traditional CNN prediction with conditional random field refinement.

\paragraph{Prostate cancer}
MRI is being increasingly used for prostate cancer diagnosis and treatment planning. Accurate segmentation of the prostate has several applications in the management of this disease. However, it is difficult to develop a fully automatic prostate segmentation method that can address various issues, such as variations of shape and appearance patterns in basal and apical regions.
Tian et al.~\cite{tian2020graph} proposed an interactive GCN-based prostate segmentation method for MRI. The method is similar to Curve-GCN~\cite{ling2019fast} and adopts a GCN to obtain the coordinates of the contour vertices by regression. The graph module takes the output feature from the CNN encoder applied to the cropped image as its input. Then, the coordinates of a fixed number of vertices from the initial contour are adjusted to fit the target. The interactive GCN model improves the accuracy by correcting the points on the prostate contour with user interactions. Finally, neighboring points/vertices are connected with spline curves to form the prostate contour. The model outperforms several state-of-the-art segmentation methods on the PROMISE12 datasets~\cite{litjens2014evaluation}.

\begin{figure}[!t]
\centering
\includegraphics[width=1\linewidth]{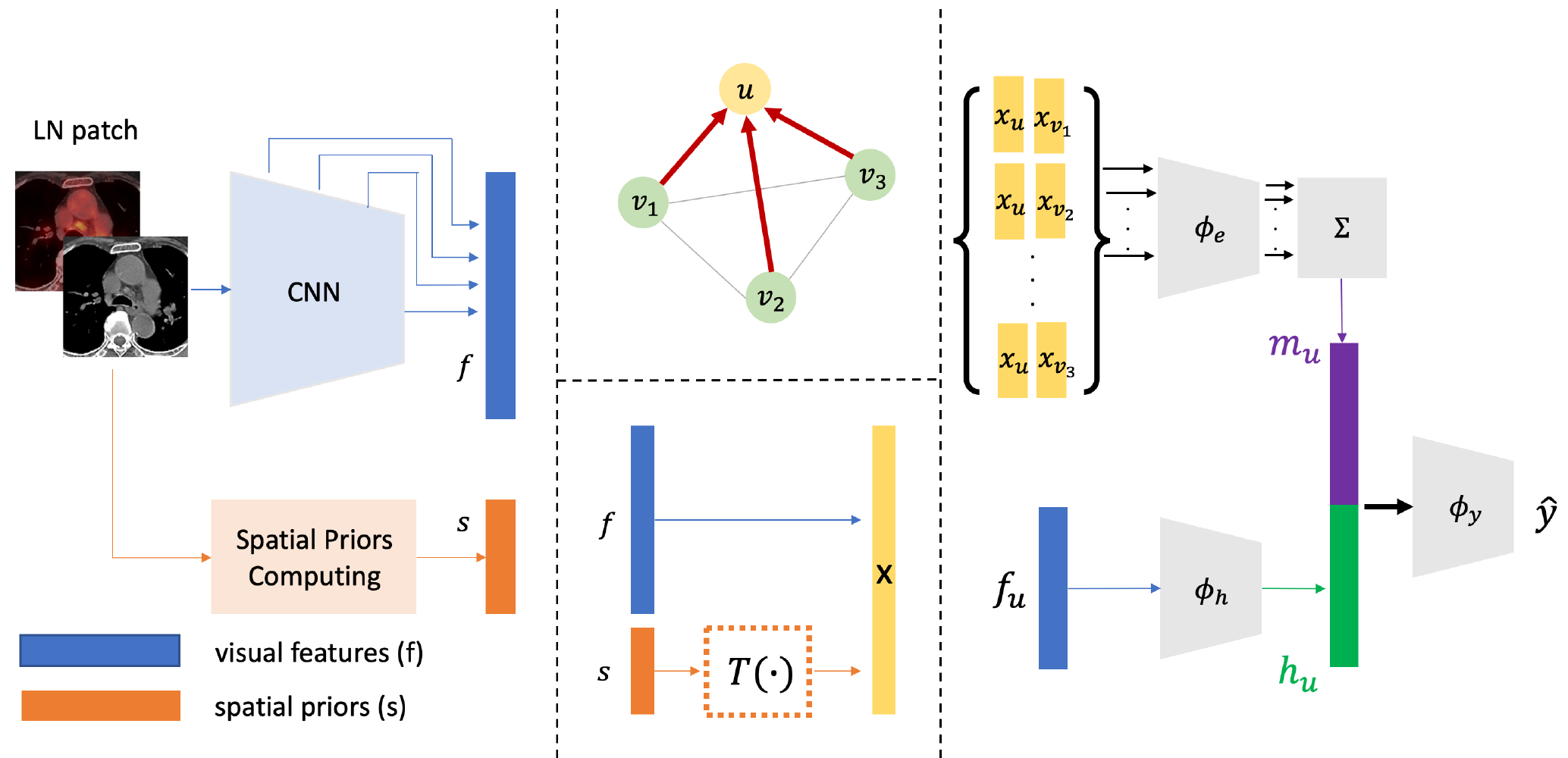}
\vspace{-15pt}
\caption{
To construct the node representation, the model extracts CNN appearance features and spatial priors for each candidate. Each GTV candidate corresponds to a node in the graph and the GCN is used to exchange information. Image adapted from~\cite{chao2020lymph}.
}
\label{fig:Fig14}
\vspace{-6pt}
\end{figure}

\paragraph{Lymph node gross tumor}
Gross tumor volume (GTV) delineations are a critical step in cancer radiotheraphy planning. In cancer treatment all metastasis-suspicious lymph nodes (LN) are also required to be treated, which is referred to as lymph node gross tumor volume. The identification of the small and scattered metastasis LNs is especially challenging in non-contrast RTCT.
Chao et al.~\cite{chao2020lymph} combined two networks, a 3D-CNN and a GNN, to model instance-wise appearance and the inter-lymph node relationships, respectively. The GNN also models the partial priors computed as the 3D distances and angles for each GTV with respect to the primary tumor. PET imaging is included as an additional input in the CNN to provide complementary information. Fig.~\ref{fig:Fig14} depicts this framework. The model delineates the location of esophageal cancer on an esophageal radiotherapy dataset, outperforming traditional CNN models.

\section{Research challenges and future directions}
\label{sec:sec4}%

\subsection{Overview}

Traditional techniques such as CNN and RNN-based models have been shown to be successful in supporting the analysis of diseases and clinical applications for tasks such as classification, detection, segmentation and reconstruction, however they are inefficient when dealing with data which are presented in an irregular grid or more generally in a non-Euclidean space. Graph embeddings provide a research avenue to deal with this form of data by constructing a vectorized feature space using graphs. Using the constructed graph representation, many machine learning problems on graphs such as node classification and graph classification can be solved using standard frameworks including GCNs and their variants which incorporate temporal dependencies and attention. One of the main goals of graph modeling is to represent a variety of irregular domains and their relations. Understanding the links between various pieces of information is crucial to providing knowledge in the best way possible, and capturing connections between nodes is as important as the data itself. This gives GNNs the ability to learn a low dimensional representation of graph-like data by an ``iterative process of transferring, transforming, and aggregating the representations from topological neighbors''~\cite{Khasahmadi2020Memory-Based}.

Graph-based deep learning models have achieved promising success in the field of medical data analysis, but there are challenges associated with their adoption in this domain that merit further discussion. 
In Subsection~\ref{sec:sec4b}, we summarise these problems and introduce some works listed in this survey to address them. We also introduce graph models that deserve attention for their potential to be adopted for medical diagnostic analysis.
Finally, in Subsection~\ref{sec:sec4c}, we suggest a group of applications for which GCNs have not yet been considered, but have great potential to improve healthcare outcomes.

\vspace{-6pt}
\subsection{Challenges in adapting graph-based deep learning methods for medical diagnostic analysis}
\label{sec:sec4b}%



We consider challenges in graph deep learning that have been highlighted in the majority of reviewed works, largely because of the numerous technical difficulties they present.
We identify \textit{seven major challenges} for graph-based deep learning adoption: 
\begin{enumerate}
\item Graph representation and estimation;
\item Dynamicity and temporal graphs;
\item Complexity of graph models and training efficiency;
\item Explainability and interpretability;
\item Generalization of graph models;
\item Data annotation efficiency and training paradigms; and
\item Uncertainty quantification.
\end{enumerate}

For each problem we discuss the theoretical and practical issues as well as discuss relevant state-of-the-art research. Advances on these challenges would permit GNNs to be extended to a broader variety of domains and applications in circumstances where traditional 2D grid representations are limited.

\subsubsection{Graph representation and estimation} 
Graph neural networks have been used to directly model graph representations of physiological signals and anatomical structures including brain signals, and organs. However, in multiple proposals discussed in this survey, graph structures are designed manually ~\cite{kazi2019inceptiongcn,parisot2017spectral}. There is a lack of structural knowledge and there are also situations where part or all of the graph structure is unknown.
Defining an appropriate graph representation where the vertices correspond to the entities or ROIs specific to the problem (\textit{e.g.} brain regions), and edges represent the connectivity of these entities, is highly relevant.
A graph itself can be complex, with several different forms and properties. Graphs can be heterogeneous or homogeneous, weighted or unweighted, and directed or undirected, to name a few examples.
Graphs also perform a wide range of tasks, ranging from node-focused problems such as node classification and link prediction to graph-focused problems such as graph classification and graph generation. Different model architectures are needed for these diverse properties and tasks.
The aim of graph structure estimation therefore is to find a suitable graph to represent the data as input to the graph model in each research domain. Several requirements are needed to improve this process.

Models that infer graph structure from data would be particularly useful when several possible graph nodes and connectivities can be chosen to represent brain signals. For ASD analysis, Rakhimberdina et al.~\cite{rakhimberdina2020population} used a method to analyse different sets of configurations to build a set of graphs and selected the best performing graph. Thus, it is expected that the whole graph, vertices and edges should be generated at the same time. 
Jang et al.~\cite{jang2019brain} proposed an EEG classification model that can automatically extract a multi-layer graph structure and signal features directly from raw EEG signals and classify them. This approach for learning the graph structure improves classification performance in comparison to approaches where a defined connectivity structure is used.
An appropriate automated graph construction method includes at least one of the following requirements:

\paragraph{Weight matrix and node connectivity}
The entries in the weight matrix of a graph are optimized with the learning objective of the model. Some authors have applied a learnable mask to automatically learn the graph structure (a learnable weight matrix)~\cite{li2019classify}. During the training process of a model, the weight matrices learn the dynamic latent graph structure. A function also learns the interactions between the nodes in the graph.

\paragraph{Dynamic weights}
The adjacency matrix is dynamically learned instead of being predetermined.
Song et al.~\cite{song2018eeg} proposed a model which could dynamically learn the intrinsic relationship between nodes through back propagation.
    
\paragraph{Edge attributes}
Edge embeddings in a graph is a poorly studied field. Edge features are included when leveraging the graph structure in the network. Learning is principally conducted on the vertices, where the edge attributes/signals supplement the learning as auxiliary information.
Edge-weighted models that incorporate edge features in the graph have been studied for ASD~\cite{li2020pooling} and BD~\cite{yang2019interpretable}.

\paragraph{Adaptive graphs}
Adaptive graph models can adapt to arbitrary topological structures and scales. They can also learn the embedding features between nodes more effectively. 
Yao et al.~\cite{yao2020temporal} introduced an adaptive graph convolutional layer that learns a data-based graph topology and captures dynamic variations of the brain for depressive disorder detection.
Gopinath et al.~\cite{gopinath2019adaptive} proposed an adaptive graph convolution pooling approach which predicts optimal node clusters for each input graph, and can handle graphs with varying numbers of nodes or varied connectivity.
To avoid the spatial limitations of a single template and learn multi-scale graph features of brain networks, Yao et al.~\cite{yao2019triplet} proposed a muli-scale triplet GCN model.
Anirudh et al.~\cite{anirudh2019bootstrapping} proposed a bootstrapped version of GCNs that made models less sensitive to the initial step of the construction of a population graph.
Isomorphism graph-based models~\cite{xu2018powerful} are designed to interpret graphs with different nodes and edges for gender~\cite{kim2020understanding} and ASD~\cite{li2019graph} analysis. 

\paragraph{Embedding knowledge} 
Incorporating domain knowledge into the model when constructing a graph or choosing an architecture has become a promising approach to improve  performance~\cite{xie2020survey}. Medical domain knowledge can be exploited to solve specific problems by creating networks that seek to mimic the way medical doctors analyse samples.
Graph based mapping with label representations (word embeddings) that guide the information propagation among nodes has been explored~\cite{chen2019multi}.
In~\cite{zhang2020radiology} a graph embedding module developed with prior knowledge on chest findings is used to learn relationships between chest pathologies and assist the generation of reports.
Hou et al.~\cite{hou2021multi} incorporate disease label embeddings as a knowledge graph for visual-semantic learning to model the correlations among different thoracic diseases.

Building graph generation models using neural networks has attracted increasing attention. These models have greater capacity to learn structural information from data and can model graphs with complicated topologies and constrained structural properties. By modeling graph generation as a sequential process, the model can compute complex dependencies between generated edges. Some graph generation approaches have adopted RNNs to exploit the graph structure during the generation process. 
Several of these scalable auto-regressive frameworks that deserve exploration within the medical domain are GraphRNN~\cite{you2018graphrnn} and GRANs~\cite{liao2019efficient}. Using GNNs with attention, GRANs, for example, capture the auto-regressive conditioning between the already-generated and to-be-generated sections of the graph.

GNNs such as GCN and GAT recursively update node embeddings by passing information from topological neighbors, allowing GNNs to capture the local structure of nodes. Then, the learned embeddings can be used for node or graph classification. However, these models cannot learn hierarchical representations which are essential for several scenarios. For example, to predict the presence of a neurological disorder, it would be desirable to infer the sub-parts which are important for brain regions hierarchically.
Various graph pooling methods have been proposed to learn the coarse-grained graph structure by adaptively retaining the most informative nodes in a hierarchical manner (SAGPool)~\cite{lee2019self}, or by finding strongly connected entities (MEMPool)~\cite{Khasahmadi2020Memory-Based}. In particular, the latter proposed a memory layer for joint graph representation learning and graph coarsening that relies on global information rather than local topology. This improved both efficiency and performance.

The above discussion has exemplified the challenges of estimating a graph structure from data with the desired characteristics. While there is work in this field, it is ripe for further exploration. Automated graph generation where a graph model infers the structural content from data is also less explored in the clinical domain.

\subsubsection{Dynamicity and temporal graphs} 
Many real-world medical application are dynamic in nature. In a graph context, this means their nodes, edges, and features can change over time. Thus, static embeddings work poorly in temporal scenarios. Several methods that analyse rs-\textit{f}MRI or EEG data discard the temporal dynamics of brain activity. 
By relying on static functional connectivity networks, these studies implicitly assume that the brain functional connectivity network is temporally stationary during the whole scanning period.
Other works have addressed this limitation by adopting generic temporal graph frameworks (RNN-based or CNN-based approaches). Spatio-temporal GCNs that exploit time-varying dynamic information have been demonstrated to outperform traditional GCNs or their variants with attention mechanisms for AD~\cite{xing2019dynamic}, MDD~\cite{yao2020temporal} and gender~\cite{gadgil2020spatio2} classification.

Although the dynamicity of graphs can be partly addressed by ST-GCNs, few clinical applications consider how to perform graph convolutions when dynamic spatial relations are present (\textit{i.e.} nodes, connections or attributes are altered). The majority of spatio-temporal methods in this survey use a predefined graph structure which assumes the graph reflects fixed relationships among nodes. Thus, generating adaptive STGNNs would play a key role in predicting how networks evolve.

Graph WaveNet~\cite{wu2019graph} proposes a self-adaptive adjacency matrix to perform graph convolutions. Graph WaveNet performs well without being given an adjacency matrix by using a complex CNN-based spatio–temporal neural network. Learning latent dynamic spatial dependencies may further improve model precision. 
Via a CNN-based approach, ASTGCN~\cite{guo2019attention} includes a spatial attention function and a temporal attention function to learn latent dynamic spatial and temporal dependencies.
Jia et al.~\cite{jia2020graphsleepnet} introduced the first attempt in using an adaptive graph model that directly models the spatio-temporal correlations without introducing elaborately constructed mechanisms for sleep staging. This model is able to learn and adjust the sleep connection structure that best serves the ST-GCN for the classification task.

In this challenge we consider temporal graphs and the need for dynamic graphs which can be applied to data where dependencies and the underlying structure changes over time. This challenge is tightly coupled to the previous challenge of learning graph representations, but adds the additional complication of seeking to learn how connections change over time. Apart from the work proposed for sleep stage classification~\cite{jia2020graphsleepnet}, the research on adaptive graphs for spatio-temporal analysis is limited.

\subsubsection{Complexity of graph models and training efficiency}
It is noted that GCNs share considerable complexity from their deep learning lineage, which can be demanding and unnecessary for less challenging applications. The simple graph convolution network proposed by Wu et al.~\cite{wu2019simplifying} reduces the complexity of GCNs by collapsing multiple weight matrices into a single linear transformation and eliminating nonlinearities between GCN layers. This model was adopted for the evaluation of ASD and ADHD~\cite{rakhimberdina2019linear}, and emotion recognition~\cite{zhong2020eeg}.

Significant efforts have recently been dedicated to coping with the problem of depth in graph neural networks, in the hope of achieving better performance. However, some experiments have shown that GCN performance drops dramatically with an increase in the number of graph convolutional layers~\cite{li2018deeper}. This raises the question of whether going deeper is still a good strategy for learning from graph data~\cite{li2019deepgcns}. Therefore, it is crucial to design deep graph models such that high-order information can be aggregated in an effective way for better predictions.
Motivated by the random mapping ability of the broad learning system~\cite{chen2017broad}, Wang et al.~\cite{wang2018eeg} and Zhang et al.~\cite{zhang2019gcb} proposed broad learning systems that are combined with a dynamic GCN for emotion analysis. These models capture deep and high-level features from the learned graph representation. GCNs are used to extract features from graph-structured input and stacks of multiple regular convolutional layers to extract abstract features. The final concatenation utilized the broad concept, which preserves the outputs of all hierarchical layers, allowing the model to search features in broad space.

Training a GCN usually requires saving the whole graph and the intermediate states of all nodes in memory. The full-batch training algorithm for GCN suffers significantly from memory overflow issues. Thus, the adoption of an efficient training approach is uncommon in the applications surveyed. 
GraphSage~\cite{hamilton2017inductive} proposes a batch-training algorithm for GCNs to save memory. PinSage~\cite{ying2018graph} extends importance sampling based on random walks. 
Improvements in speed and optimization methods for training GCNs were also suggested in~\cite{chen2018fastgcn}. Some approaches addressing these issues include:
\begin{itemize}
\item Fast learning with GCN (FastGCN)~\cite{chen2018fastgcn} takes a fixed number of nodes for each graph convolutional layer and interprets graph convolutions as integral transforms of node embedding functions under probability measures to facilitate training.
\item Stochastic GCNs (StoGCN)~\cite{chen2017stochastic} reduces the receptive field size of a graph convolution with control variate based algorithms. 
\item Cluster-GCN~\cite{chiang2019cluster} uses a graph clustering algorithm to sample a subgraph and then performs graph convolutions on nodes within the sampled subgraph. Cluster-GCN can handle larger graphs and use deeper architectures at the same time, in less time and with less memory, since the neighborhood search is also limited within the sampled subgraph.
\item Layer-wise GCN (L-GCN)~\cite{you2020l2} separates feature aggregation and feature transformation during training and greatly reduces complexity. $L^2$-GCN is also introduced where an RNN controller learns a stopping criteria for each layer trained within the L-GCN. 
\end{itemize}

There has been a growing interest in the literature in graph embedding problems, but these approaches do not usually scale to real-world graphs. Recent advances in distributed and batch training for graph neural networks looks promising but they require hours of CPU training, even for small and medium scale graphs. The use of accelerators such as GPU-based tools to deal with graphs is largely under-explored~\cite{zhu2019graphvite}. To address this, Akyildiz et al.~\cite{akyildiz2020gosh} introduced GOSH which utilizes a graph partitioning and coarsening approach, a process in which a graph is compressed into smaller graphs, to compress the graph and provide fast embedding computation on a single GPU with minimal constraints. 

How to effectively compute GNNs in order to realise their full potential will be a key research topic in the coming years.
Several hardware accelerators have been developed to cope with GNNs' high density and alternating computing requirements, but there is not a clear proposal applicable to multiple GNN variants~\cite{abadal2020computing}. On the software side, current deep learning frameworks including extensions of popular libraries such as TensorFlow and PyTorch have limitations when implementing dynamic computation graphs along with specialized tensor operations~\cite{abadal2020computing}. Thus, there is a need to further develop libraries such as DGL~\cite{wang2019deep} which may handle the sparsity of GNN operations efficiently, as well as complex tensor operations in CUDA with GPU computation acceleration.

\subsubsection{Explainability and interpretability}
Lack of transparency is identified as one of the main barriers for AI adoption in clinical practice. Clinical experts should be confident that AI models can be trusted. A step towards trustworthy AI is the development of explainable AI. Explainable AI seeks to create insights into how and why AI models produce predictions~\cite{markus2020role}, and the ability to translate computer studies into clinical applications requires interpretability.
Physicians are reluctant to trust a machine learning model's prediction because of a lack of evidence and interpretation, particularly in disease diagnosis.
Interpretability is also the source of new knowledge. A natural question that arises is if the decision making process in deep learning models can be interpretable.
Interpretability for graph-based deep learning is even more challenging than for CNN or RNN-based models because graph nodes and edges are often heavily interconnected.

Model-based and post-hoc interpretability are the two most common types of interpretation approaches. The former constrains the model so that it readily provides useful details about the uncovered relationships (such as sparsity, modularity, and so on). The latter attempts to extract information about the model's learned relationships.

Several prominent explainability methods for CNN-based models have been introduced including contrastive gradient-based saliency maps, class activation mapping~\cite{zhou2016learning}, guided backpropagation~\cite{springenberg2014striving} as well as variants such as gradient-weighted CAM (Grad-CAM). While these methods have been explored in a number of contexts, they were not proposed to address interpretability in the clinical context which brings additional requirements such as the incorporation of the physician's interpretation (\textit{i.e.} to satisfy a meaningful explanation the model should ``explain'' their outputs in such a way that physicians can understand).
These existing explainability methods are being redesigned and applied to GNNs. The main challenge is that these methods fail to incorporate relational information, which is at the core of graph data~\cite{ying2019gnnexplainer}. CNN explanation methods consider edges (connections between pixels) and focus on the nodes (pixel values); however graph data contains critical information in edges.

A pioneering work on explanation techniques for GNNs was published in 2015~\cite{duvenaud2015convolutional}. In the time since, several explanation methods have been presented including 
layer-wise relevance propagation (LRP)~\cite{baldassarre2019explainability}, 
excitation backpropagation~\cite{pope2019explainability},
graph pruning (GNNExplainer)~\cite{ying2019gnnexplainer}, 
gradient-based saliency (GRAPHGRAD-CAM)~\cite{selvaraju2017grad}, GRAPHGRAD-CAM++~\cite{chattopadhay2018grad}), and layerwise relevance propagation (GRAPHLRP)~\cite{schwarzenberg2019layerwise}.
Attention mechanisms adopted in several medical applications discussed in our survey have also been used as another explanation technique where the attention weights for edges can be used to measure edge importance; however, it is noted that they can only explain GAT models without explaining node features, unlike, for example, GNNExplainer.

Interpretable GNN models, in which internal model information such as weights or structural information can be accessed and used to infer group-level patterns in training instances, have been considered in some works~\cite{kim2020understanding,yang2019interpretable}.
Post-hoc interpretation approaches have been used in other studies to explain GNNs~\cite{pope2019explainability,baldassarre2019explainability}.
These post-hoc methods are typically used to analyze individual feature input and output pairs, limiting their explainability to an individual-level only.

While there is much interesting research within this field, it is immature and there are only a few papers that explore GNN explanation methods in the medical domain.
Post-hoc explanation techniques have been used to visualize attention weights or to recognize relevant subgraphs for classification. 

Other highlighted preliminary experiments considering the interpretability of model outcomes across the surveyed papers are:
\begin{itemize}
    \item Song et al.~\cite{song2018eeg}. The adjacency matrix provides a potential way to find the most important EEG channels for EEG emotion recognition, which is beneficial for further improving EEG emotion recognition performance.
    \item Wang et al.~\cite{wang2020sequential}. The interpretability comes from the learnable weight vector where each weight corresponds to a specific frequency and can be seen as the importance of that frequency for epilepsy detection.
    \item Covert et al.~\cite{covert2019temporal}. This work illustrates the advantages of TGCN for helping clinicians determine precisely when seizures occur, and the parts of the brain that are most involved.
    \item Zhang et al.~\cite{zhang2020deeprep}. A brain saliency map is derived to rank the top key brain regions associated with structures conventionally conceived as the biomarkers of PD.
    \item Yang et al.~\cite{yang2019interpretable}. The activation map and gradient sensitivity of GAT models are used to interpret the salient input features at both the group and individual levels for the analysis of BD.    
    \item Gadgil et al.~\cite{gadgil2020spatio2}. The model learns edge importance to localize meaningful brain regions (ROI) and selective functional connections significantly contributing to gender prediction. For example, the most important ROI identified by the model was the inferior temporal lobe and the frontal-posterior-cingulate (PCC) connection which reflects findings from other neuroimaging analyses.
    \item Azevedo et al.~\cite{azevedo2020deep2}. The model contains elements of explainability by analysing how the graph hierarchical pooling mechanism assembles the brain regions to optimise gender prediction. These clusters are important in explaining the behavioural differences in terms of cognitive, motor and emotional skills between males and females.
    \item Li et al.~\cite{li2020braingnn,li2020pooling,li2019graph}. A BrainExplainer is introduced which uses an ROI-selection pooling layer that highlights the importance of brain region relationships for ASD prediction.
\end{itemize}

Both individual-level and group-level explanations are critical in medical research. Individual-level biomarkers are desirable for planning targeted care in precision medicine, while group-level biomarkers are essential for understanding disease-specific characteristic patterns.
Li et al.~\cite{li2020braingnn} introduced a tuneable regularization term for the graph pooling function to address this challenge where a single framework with built-in interpretability is used for individual and group-level analysis.

Further studies are required that consider clinical workflow integration, and investigate how a clinical expert could refine a model decision via a human-in-the-loop process.
One such example is Tian et al.~\cite{tian2020graph} who introduce an interactive GCN-based prostate segmentation method based on the Curve-GCN~\cite{ling2019fast} model. An annotator can choose any wrong control points and correct these via user interactions.

The implementation of deep-learning based models raises complex clinical and ethical challenges due to difficulties in understanding the logic involved in these models. Interpretability is essential as it can help informed decision-making during diagnosis and treatment planning. Many methodological advances have been made for medical tasks such as dealing with graph learning, graph heterogeneity and multiple graph scenarios. However, GCNs are complex models and interpreting the model's outcome remains a challenging task. 
Interpretability techniques are gaining importance in recent years. However, aside from the initial study in pathology images~\cite{jaume2020towards,jaume2020quantifying} which is outside the scope of this review (but will be explored in our future study), the interpretability of graph neural networks in a clinical context has not been addressed sufficiently. Considering the spread of graph-based processing for various medical applications, graph explainability and the quantitative evaluation with a focus on usability by clinicians is crucial.

\subsubsection{Generalization of graph models} 

Several graph methods suffer from challenges posed by inter-site heterogeneity caused by different scanning parameters and protocols, and from subject populations at different sites. It is difficult to build accurate and robust learning models with heterogeneous data. Due to patient privacy and clinical data management requirements, truly centralized open source medical big data corpora for deep learning are rare.
Medical applications are hindered by non-generalizability that limits deployment to specific institutions. To alleviate the heterogeneity, simultaneously learning adaptive classifiers and transferable features across multiple sites and subjects offers a promising direction.

Transfer learning provides a potential solution by transferring well-trained networks on large scale datasets (related to the to-be-analyzed disease) to a small sample dataset.
Wee et al.~\cite{wee2019cortical} adopted a spectral graph CNN to demonstrate generalization across datasets by detecting AD with a classifier trained on a Caucasian population, and testing it with a model fine-tuned on an Asian population. This demonstrated that a GCN is capable of capturing essential dementia-associated patterns from different datasets.

Domain adaptation is a form of transfer learning in which the source and target domains have the same feature space but different distributions. This also aims to deal with multiple domains and even multiple heterogeneous tasks.
The generalizability of trained classifiers in subject-independent classification settings is hampered by the considerable variation in physiological data across different subjects. Several studies have attempted to tackle this challenge through domain adaptation techniques~\cite{dissanayake2020domain}. Li et al.~\cite{li2018bi} introduced domain adversarial training to lower the influence of individual subjects on EEG data, without exploiting a graph-based input signal.

Graph adversarial methods adopt adversarial training techniques to enhance the generalization ability of graph-based models.
Zhong et al.~\cite{zhong2020eeg} introduced a node-wise domain adversarial training method and an emotion-aware distribution learning approach as regularizers for better generalization in subject-independent classification scenarios. 
Gopinath et al.~\cite{gopinath2020graph} proposed an adversarial graph domain adaptation method for surface segmentation. This method offers better generalization on target-domain datasets where surface data is aligned differently, without requiring manual annotations or explicit alignment of these surfaces. Adversarial networks are also used for brain data prediction~\cite{hong2019multifold,hong2019longitudinal}

Meta-learning, a sub-field of transfer learning, has been used in areas including task-generalization problems such as few-shot learning. Meta-learning is the ability to learn, also known as "learning to learn". 
Few-shot learning (FSL) aims to automatically and efficiently solve new tasks with few labeled samples based on knowledge obtained from previous experiences. These models are emerging in the medical domain~\cite{mahajan2020meta} for decoding brain signals~\cite{bontonou2020few} and a few approaches have explored GNNs for few-shot learning~\cite{kim2019edge}.

Another interesting variant of transfer learning is zero-shot learning (ZSL) which aims to predict the correct class without being exposed to any instances belonging to that class in the training dataset. Although zero-shot learning is flourishing in the field of computer vision~\cite{caceres2017feature}, it is seldomly used for biomedical signal analysis, though zero-shot learning has recently been used to recognize unknown EEG signals~\cite{duan2020zero}.
Recently, graph convolutional networks have shown a lot of promise for zero-shot learning. When encountered with a lack of data, these models are highly sample efficient because related concepts in the graph structure share statistical strength, allowing generalization to new classes~\cite{kampffmeyer2019rethinking}. Knowledge graphs can also be used to guide zero-shot recognition classification as extra information~\cite{kampffmeyer2019rethinking}, making these models a notable future research prospect.

We analyze the challenges that affect the ability of GCNs to generalize, including unknown tasks and domains, and the potential research directions this offers. A number of interesting paths exist, including the development of meta-models that address the problem of knowledge generalization to enable the more rapid deployment of applications.

\subsubsection{Data annotation efficiency and training paradigms}
Graph-based deep learning has achieved great success on various tasks; however, as deep learning exploits hierarchical feature representations which are highly data-driven, there are several critical challenges for medical applications including scarce annotation, complexity and weak annotations, and variability or label sparsity.

To address these challenges several training paradigms have been proposed in the literature including weakly or partially supervised learning, semi-supervised learning, and self-supervised learning~\cite{cheplygina2019not}.

To learn rich representations, GNNs typically require task-dependent labels. However, compared to other modalities such as video, image, text, and audio, annotating graphs is more complex.
Although the issue of missing labels is a general problem not specific to the graph domain, only a few works have adopted the training paradigms previously discussed, with semi-supervised examples being the segmentation of cerebral cortex~\cite{gopinath2020graph} and organs~\cite{soberanis2020uncertainty}.

Although self-supervision in graphs~\cite{you2020does} and graph convolutional adversarial networks for unsupervised domain adaptation~\cite{ma2019gcan} have been investigated to improve the performance in the presence of partially labeled data, their adoption has not been considered in the works surveyed.

Weakly supervised algorithms have been widely explored in pathology imaging such as prostate cancer~\cite{wang2020weakly}, skin cancer~\cite{wu2019weakly}, chest pathologies~\cite{wang2017chestx}, yet further research is required to investigate such methods on anatomical datasets.

When no class labels are available in graphs, we can learn the graph embedding using an end-to-end method in an entirely unsupervised manner. These algorithms exploit edge-level information. One method is to use an autoencoder framework, in which the encoder uses graph convolutional layers to embed the graph in a latent representation, which is then reconstructed using a decoder~\cite{wu2020comprehensive}.
Recent works on contrastive learning by optimizing mutual information (MI) between node and graph representations have achieved state-of-the-art results on both node classification~\cite{velickovic2019deep} and graph classification tasks~\cite{sun2019infograph}. Nonetheless, these methods require specialized encoders to learn graph or node level representations.
Velickovic et al.~\cite{velickovic2019deep} introduce a deep graph infomax learning algorithm for graph‐structured inputs. This unsupervised objective enables every local component of the graph to be aware of the graph’s global structural properties in a seamless manner. As a result, the model can generate node embeddings that are comparable to those generated by similar encoders trained with a supervised objective.
Hassani et al.~\cite{hassani2020contrastive} introduced a self-supervised approach to train graph encoders by maximizing MI between representations encoded from different structural views of graphs. The model outperformed self-supervised models without requiring a specialized architecture.

One aspect of graph-based deep learning not yet discussed for medical applications is reinforcement learning (RL), which learns from experiences by interacting with the environment.
RL can address the limitation of supervised learning with robust and intuitive algorithms trainable on small datasets~\cite{coronato2020reinforcement}. 
RL is derived from the concept that an agent learns the correct behavior through interactions in a dynamic environment. Reinforcement learning, like supervised learning, starts with a classifier constructed from labeled data. However, when the system is then given unlabeled data, it attempts to enhance the classification by better characterizing the data, similar to how unsupervised learning works.
There are opportunities to employ RL in multi-task and multi-agent learning paradigms where graph convolutions adapt to the dynamics of the underlying graph of the multi-agent environment~\cite{jiang2020graph}, or to apply RL for graph classification using structural attention to actively select informative regions in the graph~\cite{lee2018graph}.

Another potential approach that has been not considered in the surveyed manuscripts is Federated learning (FL). FL works on the concept of remote execution and enables collaborative learning among multiple clients, which helps to alleviate label scarcity while also protect data privacy and data access rights. Thus, these approaches have arguably become the most widely used next-generation privacy preservation technique in medical AI applications~\cite{rieke2020future}.
However, existing FL methods 1) perform poorly when data is non-independent identically distributed through clients, 2) cannot manage data with new label domains, and 3) cannot exploit unlabeled data; all of these issues are present in real-world graph-based problems.
Wang et al.~\cite{wang2020graphfl} proposed a FL framework to perform graph-based semi-supervised node classification to address these challenges.
Although a few works on FL in medical imaging have been reported for brain tumor segmentation~\cite{li2019privacy} and ASD detection~\cite{li2020multi}, there are few applications of graph networks and FL in the medical domain.

\subsubsection{Uncertainty quantification}
In medical applications, uncertainty can be decomposed into Aleatoric and Epistemic uncertainty:
(i) \textit{Aleatoric uncertainty} (also known as statistical uncertainty) results from noise in the data. This is common in clinical data due to the complexity of the data and the large variability among experts. One example of noise in medical labels is presented in~\cite{dgani2018training} and they proposed a model to take into account label uncertainty in the network architecture and model training.
(ii) \textit{Epistemic uncertainty} can occur due to the incompleteness of a model. For instance, the lack of well-defined nodes results in uncertainty of the graph topology, and therefore makes it hard to adopt GNNs for disease prediction~\cite{huang2020edge}. This could be partially dealt with by a method that can learn how to construct graphs rather than hand-designing them. Soberanis-Mukul et al.~\cite{soberanis2020uncertainty} also determined the uncertainty along with model expectation to perform a semi-supervised pancreas and spleen segmentation refinement task.

Ryu et al.~\cite{ryu2019uncertainty} investigated Bayesian neural networks to quantify uncertainties in molecular property predictions. From their experiments (using BNNs which is based on an augmented GCN and Bayesian neural network), it can be concluded that when the noise in data increases it causes an increase in the aleatoric uncertainty, whereas the epistemic uncertainty is determined by the quality of the data. In addition, they showed that the uncertainty can be regarded as the confidence of prediction.
Zhao et al.~\cite{zhao2020uncertainty} also mentioned the lack of attention given to uncertainty estimation in methods using GCNs, which could result in increasing the risk of misclassification under uncertainty in real data. Therefore, the authors proposed a multi-source uncertainty framework while using a GNN to determine possible types of predictive uncertainties for node classification predictions. 

Considering the importance of quantifying uncertainty which can be rooted in data noise or the algorithm itself, still more works are required to deal with this challenge in GNNs as they are adopted to medical tasks.

\vspace{-6pt}
\subsection{Future prospects of graph neural networks for patient behavioural analysis}
\label{sec:sec4c}%


Medical applications have benefited from rapid progress in the field of computer vision. Up to this point, the majority of studies have concerned themselves with analysing data that results from diagnostic procedures, and using this to predict the presence of a disease. As a result of this focus, the areas of patient behaviour monitoring, and motor and mental disorder assessment, have received less attention. While several in-clinic systems using CNNs and RNNs-based models have been introduced enabling comprehensive data analysis through accurate and granular quantification of a patient's movements, these methods are not yet sufficiently accurate for widespread clinical use, yet we argue that graph neural networks have great potential in these application areas.

From a clinical standpoint, the main benefits of such behaviour monitoring, and motor and mental disorder assessment tools are: i) Complementary, objective, and quantitative information provided to clinicians; ii) The ability to detect and quantify events that are difficult to observe (\textit{e.g.} a fall during the night); iii) A reduction in time and effort involved in documenting useful information for diagnosis; and iv) Assessment in locations and clinics where human expertise may not be available. 
Existing vision-based systems (using CNNs and RNNs) have attracted great attention due to their non-invasive nature and have shown promising results in analysing in-bed patient-specific pose~\cite{liu2019seeing,chen2018patient} and patient behaviours (facial and body motions) in multiple clinical contexts including breathing disorders~\cite{martinez2019vision}, seizure disorders~\cite{ahmedt2019understanding,ahmedt2019vision}, infant motions~\cite{hesse2019learning}, and pain management~\cite{rodriguez2017deep}.

Image classification, regression and segmentation has been addressed with CNN models by excelling at modeling local relations. However, GCNs can take into account different neighboring relations (global relation) by going beyond the local pixel neighborhoods used by convolutions. To extract interaction information between objects, a CNN model needs to obtain sufficient depth by stacking multiple convolutional layers which is very inefficient.
As can be seen from our survey, GCNs and their variants have shown impressive results in analysing images for the purpose of anatomical structure analysis.
Graph embeddings have also appeared in other computer vision tasks where relations between objects can be efficiently described by graphs, or for the purpose of graph-structured image data analysis. Interesting results have been obtained for object detection, semantic segmentation, skeleton-based action recognition, image classification and human-object interaction tasks~\cite{zhang2020deep,wu2020comprehensive}. However, the adoption of graph-based deep learning models for the analysis of human behaviour in a clinical context has not been sufficiently explored. 
Developing graph based models to support the highly relevant research domain of behaviour analysis is of great interest for clinical applications, and below we outline several potential application domains.


\begin{figure}[!t]
\centering
\includegraphics[width=1\linewidth]{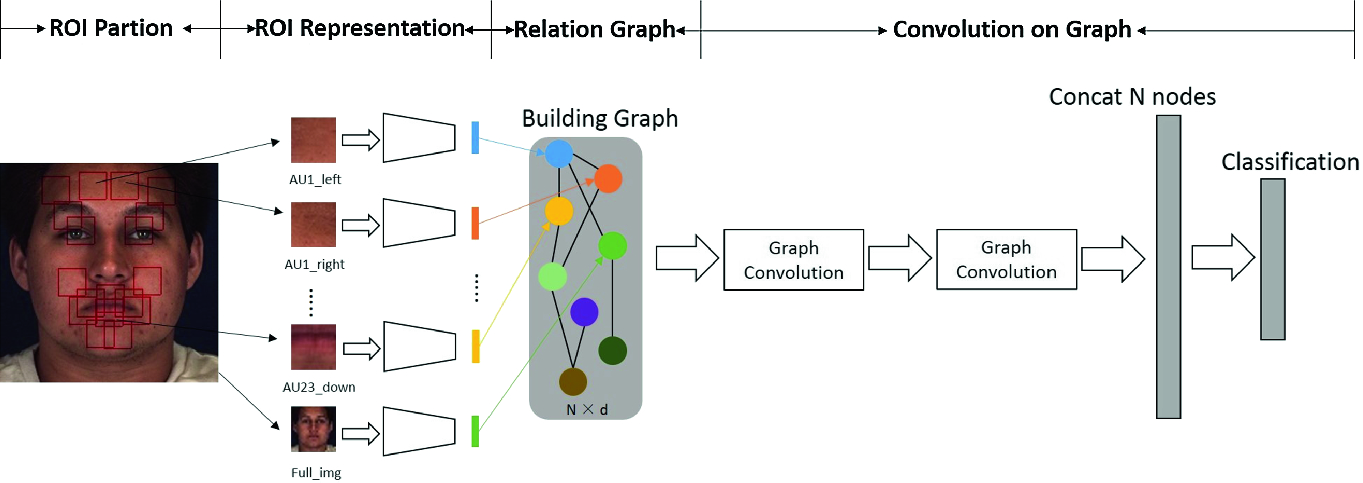}
\caption{
AU detection by considering AU relation modeling through GCN. Image adapted from~\cite{liu2020relation}.
}
\label{fig:Fig33}
\vspace{-6pt}
\end{figure}

\subsubsection{Face analysis}

Clinical experts rely on certain facial modifications and symptoms for assistive medical diagnosis, and computer vision has been introduced to offer an automatic and objective assessment of facial features~\cite{thevenot2017survey}. 
Generic CNNs and spatial methods have been effective for feature extraction for facial expression and emotion recognition using facial landmarks or facial action unit (AUs) detection. However, CNNs used to learn spatio-temporal relationships only pay attention to crucial facial parts and do not consider hidden inter-relations among facial movements, which can be captured with GCNs. With graph-based models, both facial changes and the relationships between changes are captured as important cues. In a general pipeline, latent representations of ROIs or AU regions are extracted with CNNs, and GCNs are adopted to model relations. The relational reasoning and the spatio-temporal pattern learning of graph-based methods have been explored for facial expression recognition~\cite{zhou2020facial}, action unit detection~\cite{liu2020relation} and micro-expression recognition~\cite{lo2020mer}. An example of regions related to AUs is illustrated in Fig.~\ref{fig:Fig33}.
Attempts using graph-based methods to perform 3D face reconstruction have been also explored. Since the structure of a 3D face mesh is naturally a graph structure, the adoption of a graph representation can provide a robust, efficient and detailed 3D facial mesh~\cite{cheng2020faster}. 
Lin et al.~\cite{lin2020towards} used single-view images in-the-wild to recreate 3D facial shapes with high-fidelity textures without the need for a large-scale face texture database. 
Therefore, creating a complementary graph representation and relational reasoning approach using GNNs in the clinical context is yet to be explored.

\textit{Potential applications}: 
Postoperative pain management, monitoring vascular pulse, facial paralysis assessment, and several neurological and psychiatric disorders including seizure semiology, ADHD, autism, bipolarity and schizophrenia.

\subsubsection{Human pose localization}

\begin{figure}[!t]
\centering
\includegraphics[width=0.7\linewidth]{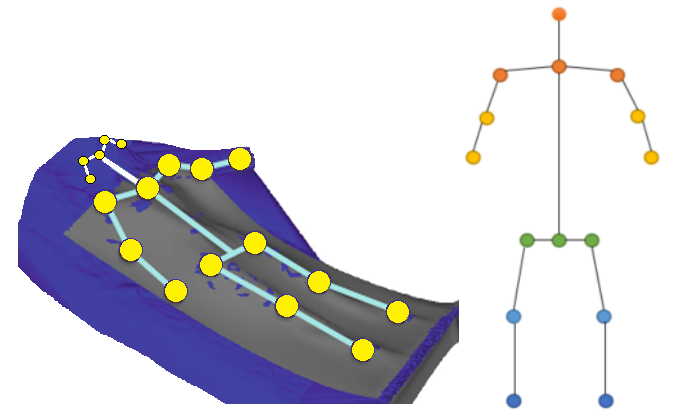}
\caption{
Human body modeling, kinematic model for graph-based representation.
}
\label{fig:Fig32}
\vspace{-6pt}
\end{figure}

Human posture captures important health-related metrics with potential value in the assessment of medical conditions such as epilepsy, sleep monitoring and surgery recovery. The poses that patients occupy carry important information about their physical and mental health~\cite{ahmedt2017automated}. Although camera-based methods for human pose estimation have been studied extensively, in-bed pose analysis comes with specific challenges including joints being occluded, blur and low-light conditions. 
Traditional pose estimation models are roughly divided in two groups: a human bounding box is first detected and then keypoints (which indicate joint locations) for each person are estimated (top-down methods); or all keypoints are detected in the first stage and they are assigned to each person in a second stage (bottom-up methods or box-free human detection).
Since the human skeleton is inherently organised as a graph rather than a grid, RNNs and CNN-based methods find it difficult to fully exploit the structural details embedded in the human skeleton.
These approaches do not take into account relationships between keypoints which is important when judging location,  especially in the case of occlusion. The location can be inferred only from the location of other related keypoints. For example, the position of the elbow joint depends on the location of the upper arm bone, which simultaneously constrains the location of the forearm bone. 
The kinematic model, also termed a skeleton-based model or kinematic chain model, as shown in Fig.~\ref{fig:Fig32}, includes a set of joint positions and limb orientations to represent the human body structure. A Kinematic model has the advantage of a flexible graph-representation which consists of nodes (keypoints) and edges (relations between keypoints). 

Since human pose estimation is related to graph structure, it is important to design appropriate models to estimate joints that are ambiguous or occluded. GCNs can process skeleton data in a flexible way and add various incremental modules to improve the skeleton structure's expressive power. In a refinement phase, considering the relationship between keypoints may help to avoid and correct errors.
Zhang et al.~\cite{zhang2019human} adopted a pose-GNN to exploit spatial contextual information among different joints for precise pose prediction. This model collects information from a set of neighboring nodes and updates all states of the node simultaneously.
Wang et al.~\cite{wang2020graph} proposed a two-stage method where the first stage proposes potential keypoints (via classic heatmap regression) and the second stage is an extended GCN, termed a graph pose refinement (GPR) module, to get refined localised keypoint representations.
%
Graph convolutional networks have also been applied to 3D human pose estimation where for example a GNN captures local and global node relationships for 3D pose estimation in images~\cite{zhao2019semantic}, and an undirected graph is used to model spatio-temporal dependencies between different joints for 3D pose from video data~\cite{cai2019exploiting}.

Multi-person pose estimation is challenging because it must estimate keypoints for an unknown number of persons simultaneously. Several top-down methods assume that every detected human bounding box only contains joints for one target person. However, this is impractical, especially in crowded scenes. This problem manifests when patients, clinicians and family members are visible in the same region of interest, causing confusion in joint estimation and assignment. Thus, a joint-to-joint relation modeling approach is needed.
Jin et al.~\cite{jin2020differentiable} proposed a differentiable hierarchical graph grouping method to learn human part grouping. The nodes of a graph denote the keypoint proposals, and edges denote whether the two keypoints belong to the same person. The graph structure is adaptive to different input images instead of constructing a static graph, so it is able to dynamically group various numbers of keypoints into various numbers of human instances.
Qiu et al.~\cite{qiu2020dgcn} introduced a dynamic GCN which can model rich relations for bottom-up pose estimation. The relations between human keypoints dynamically change according to the variations in viewpoints, occlusion, and truncation.

\textit{Potential applications}: 
In-bed pose estimation to track pressure injuries from surgery and illness recovery and other sleep disorders such as apnea, pressure ulcers, and carpal tunnel syndrome.

\begin{figure}[!t]
\centering
\includegraphics[width=0.8\linewidth]{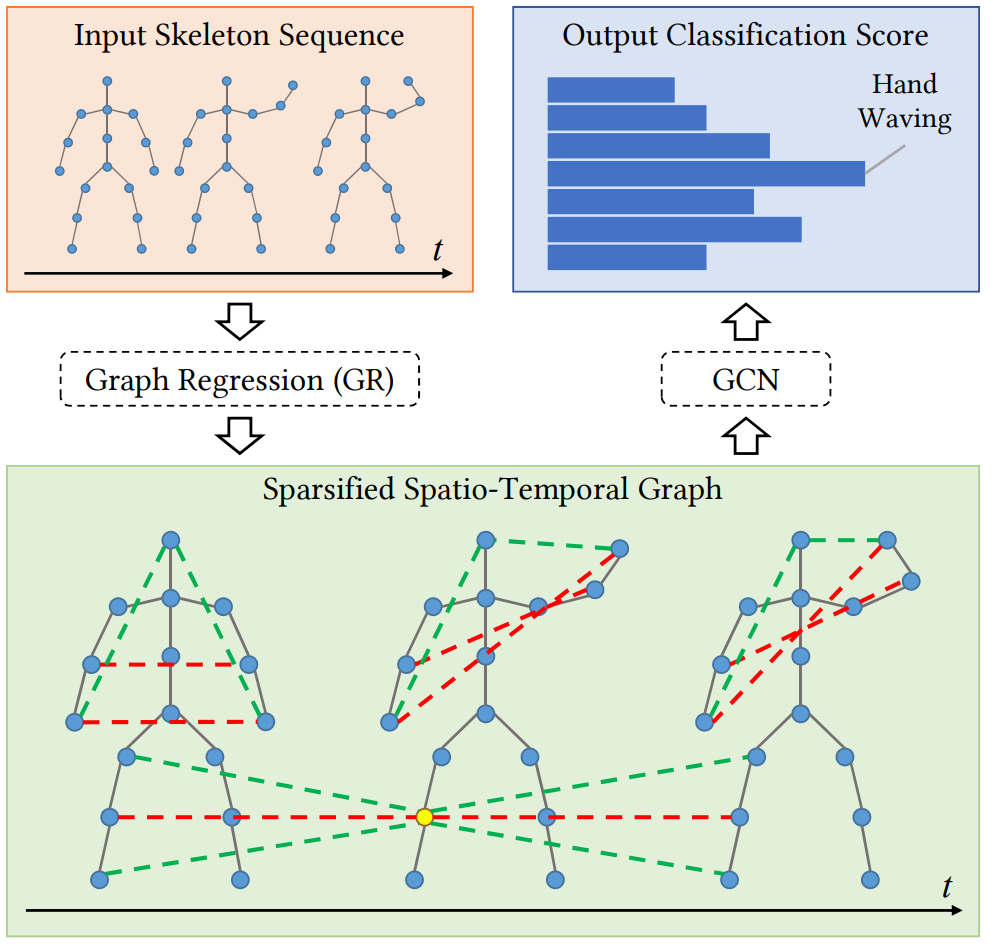}
\vspace{-2pt}
\caption{
Graph regression is adopted to learn a spatio-temporal graph for variation modeling as input to a GCN for action feature learning. Image adapted from~\cite{gao2019optimized}.
}
\label{fig:Fig36}
\vspace{-6pt}
\end{figure}

\subsubsection{Pose-based action recognition and behaviour analysis}

Movement assessment and monitoring is a powerful tool during clinical observations where uncontrolled motions can aggravate wounds and injuries, or aid the diagnosis of motor and mental disorders. These motions are represented as continuous time-series of the kinematics of the head, limbs and trunk movements. 
Pose-based action recognition is often fast since the human pose representation is very compact. Methods also perform reasonably well in recognizing actions that have less relation with the environment and are mostly human-related. The reason for this is that the model should only infer the action from the human motion and should not consider the background environment, or the objects that the subject may interact with. Therefore, these approaches require human poses to be extracted first~\cite{yan2018spatial,liang2019three}. Skeleton-based action recognition has been studied by focusing on two aspects: the intra-frame representation for mutual co-occurrences, and the inter-frame representation to map the temporal evolution of a skeleton~\cite{li2018co}.

Deep neural networks have been used to capture patterns in the spatial configuration of the joints as well as their temporal dynamics. However, as mentioned before, skeletons themselves are in the form of graphs.
There are notable characteristics for graph-based human skeleton sequences: i) joint and bone information are complementary and combining them can lead to further improvements for skeleton-based action recognition; ii) temporal continuity exists not only across joints, but also in the body structure; iii) there is a co-occurrence relationship between spatial and temporal domains; and iv) the temporal dynamics of a skeleton sequence also contain significant information for the recognition task~\cite{li2020temporal}.

\begin{figure}[!t]
\centering
\includegraphics[width=0.65\linewidth]{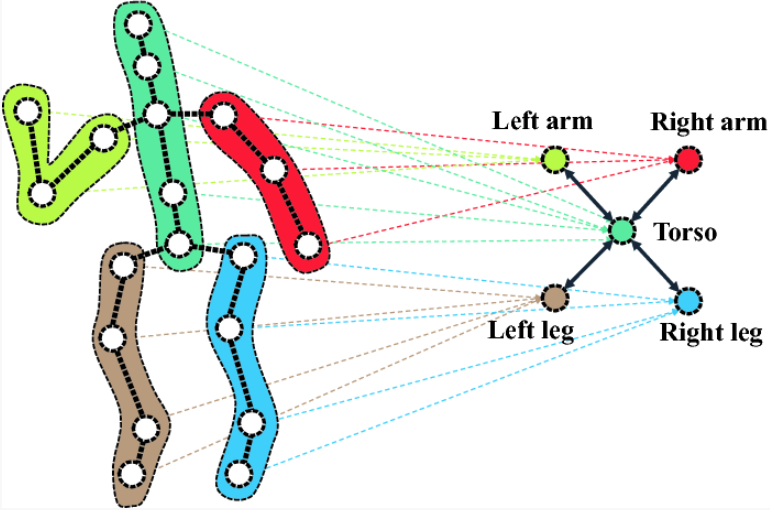}
\caption{
The physical relation between body parts is used to construct the adjacency matrix for graph convolution. Image adapted from~\cite{huang2019hierarchical}
}
\label{fig:Fig37}
\vspace{-6pt}
\end{figure}

Given a time series of human joint locations, GCNs have been widely used to estimate human action patterns. The proposed ST-GCN model~\cite{yan2018spatial} was the first work utilizing graph networks for action recognition which improved the state-of-the-art by a large margin. 
Si et al.~\cite{si2018skeleton} utilizes a graph-based model and LSTM to represent spatial reasoning and temporal stack learning for skeleton-based action recognition, respectively. The same authors proposed an attention enhanced graph convolutional LSTM~\cite{si2019attention} which captures both spatio-temporal features and the co-occurrence relationship between spatial and temporal domains.
A two-stream adaptive GCN with self attention was introduced in~\cite{shi2019two} for action recognition.
Li et al.~\cite{li2019actional} introduced an action-structural graph which captures both action links and structural links over time. To capture rich dependencies between joints over time, the authors introduce an encoder-decoder structure, termed an A-link inference module, to capture action-specific latent dependencies, \textit{i.e.} action links, directly from actions. 
Since learning the graph structure from data is critical for classification, a graph regression based a GCN was proposed to learn a sparse spatio-temporal graph for effective action feature learning~\cite{gao2019optimized} as illustrated in Fig.~\ref{fig:Fig36}.
Knowing that other approaches focus on the spatio-temporal patterns of body joints, Huang et al.~\cite{huang2019hierarchical} proposed a hierarchical GCN to model the hierarchical information of human actions, \textit{i.e.}, the movement of human body parts in action recognition. Fig.~\ref{fig:Fig37} depicts an example of the physical relation between the body parts.
Other graph models that have drawn attention to action recognition include a two-stream GCN for zero-shot action recognition~\cite{gao2019know}, and shift-GCN~\cite{cheng2020skeleton} which introduces a shift graph operation for reduced computational complexity.

Graph representations for skeleton-based action recognition are gaining importance in the last couple of years. Apart from the initial study to flag abnormal behaviour in dementia~\cite{arifoglu2020detecting}, assessment of parkinsonian leg agility and gait~\cite{guo2020sparse,guo2021multi}, and human emotion based on skeleton detection~\cite{tsai2021spatial}, graph neural networks in the context of in-bed pose estimation and patient behaviour estimation are poorly investigated compared to other computer science fields. 

\textit{Potential applications}: 

\begin{itemize}
    \item
    \textit{Motor disorders:} Epilepsy, Parkinson, Alzheimer, stroke, tremor, Huntington and neurodevelopmental disorders. 
    
    \item 
    \textit{Mental disorders:} Dementia, schizophrenia, major depressive, bipolar and autism spectrum.
    
    \item 
    \textit{Other situations:} Breathing disorder, inpatient falls prediction, health conditions such as agitation, depression, delirium, unusual activity or to evaluate human interaction in a hospital environment~\cite{torres2018healthcare}, 
    
\end{itemize}

\section{Conclusion}

Functional, anatomical, electrical and histology data provide essential information on many diseases' etiology, onset, and progression, as well as treatment efficacy. Our survey provides a comprehensive review of research on graph neural networks and their application to medical domains and applications including functional connectivity, electrical, and anatomical analysis. Digital pathology has not been the main focus of this survey, and we have sparsely mentioned the applications of GCNs to this domain.  However, considering the comprehensive application of deep learning to digital pathology (WSI), future work will include a survey to thoroughly cover the potential applications of GCN to WSI. 
As we have shown in this review, the growing mass of literature in this space and the rapid development and search for new tools and methods suggests we are at the verge of a paradigm shift. Further, considering the remarkable ability of GCN in dealing with unordered and irregular data such as brain signals, and their simplicity and scalability, graph-based deep learning will progressively take a more prominent role and complement traditional machine learning approaches.

Recent advances in the adoption of graph-based deep learning models for classification, regression and segmentation of medical data shows great promise. However, we have outlined several challenges related to their adoption, including the graph representation and estimation, graph complexity, dynamicity, interpretability and generalization of graphs. These and many other challenges lead to a vast amount of open research directions, solutions to which will benefit the field and lead to many applications in the medical domain. 
This constitutes a clear challenge to the neuroengineering scientific community and it is hoped the community will increase their efforts to address these emerging challenges. Although one will never replace the power of individual clinical expertise, by providing more quantitative evidence and appropriate decision support, one can definitely improve medical decisions and ultimately the standard of care provided to patients.


%





\ifCLASSOPTIONcaptionsoff
  \newpage
\fi



%

{\small
        \bibliographystyle{IEEEtran}
        \bibliography{ref}
}


%








\end{document}